\documentclass[conference,10pt,letterpaper]{IEEEtran}
\makeatletter
\def\input@path{{./}{arxiv_source/}}
\makeatother
\usepackage[T1]{fontenc}
\usepackage{cite}
\usepackage{comment}
\usepackage{pgfplotstable}
\pgfplotsset{compat=1.18}
\usepackage{bm,adjustbox}
\usepackage{multirow}
\usepackage{amsmath,amssymb}
\usepackage{graphicx}
\usepackage{tikz}
\usetikzlibrary{arrows.meta}
\usepackage{booktabs,array,tabularx}
\usepackage{xcolor}
\usepackage{colortbl}
\usepackage{tikz}
\usepackage{algorithm}
\usepackage{algpseudocode}
\usetikzlibrary{arrows.meta,positioning,fit,calc}
\usepackage[hidelinks]{hyperref}

\newcommand{\areaunit}{\ensuremath{\mu\mathrm{m}^{2}}}
\newcommand{\parhead}[1]{\par\noindent\textbf{#1.}\ \ignorespaces}

\title{CASCoDe: A Closed-Loop Agentic Framework for Analog Sizing--Layout Co-Design}
\author{\IEEEauthorblockN{Ziming Wei, Bingyang Liu, and David Z. Pan}
\IEEEauthorblockA{The University of Texas at Austin, Austin, TX, USA}}
\begin{document}
\maketitle
\begin{abstract}
Expert analog layouts differ from auto-generated ones in density and regularity, not in legality: experts make a succession of discrete, hierarchical physical decisions and revisit each after extracted electrical results, while existing automated flows act on a physical search space fixed before the controller runs. We present CASCoDe, a closed-loop LLM agent framework for analog netlist-to-layout design that formalizes those expert decisions as a hierarchical discrete action space---sizing candidate pool, structural floorplan, intra-row refinement, routing-attempt selection---on which the agent runs feedback-driven tree search from a sizing pool to a verified GDS. The same loop drives all 20 evaluated circuits, with distinct physical bottlenecks, unchanged; the agent authors the circuit-specific floorplans and helper scripts, and a thin execution layer of interface adaptations over unmodified tools makes every action land at exact coordinates and be judged by DRC, LVS, and extracted-RC simulation. On 20 SKY130 amplifiers spanning single-, two-, and three-stage topologies, CASCoDe is the only flow that delivers a DRC- and LVS-clean routed layout for every circuit; it reduces area by 59--85\% (mean 71\%) against a conventional P\&R flow on identical PCells, raises mean device density from 0.39 to 0.74, and attains a higher mean post-layout FoM than APPlace, MAGICAL, and ALIGN, which are DRC-clean on 15, 0, and 1 circuits. A layout that beats every baseline in area and legality appears after 11 minutes of agent time on average, and ablations show that the structured action space and the feedback-guided tree search each reduce area and search cost.
\end{abstract}
\begin{IEEEkeywords}
Analog layout automation, large language models, LLM agents, closed-loop design, sizing--layout co-optimization, physical design.
\end{IEEEkeywords}
\section{Introduction}
\label{sec:intro}

Analog design automation ultimately aims to produce compact physical layouts that satisfy circuit performance requirements. Sizers~\cite{mace,autockt} and placement-and-routing tools~\cite{magical,align,diffsp} have improved the quality and efficiency of individual design stages. However, these improvements do not necessarily translate into better final layouts, because layout compactness and post-layout performance depend on coordinated decisions across stages. 
Although several frameworks integrate multiple stages~\cite{postbo,panda,analogmaster}, further refinement is still required to achieve the quality of carefully engineered layouts. 
These remaining limitations can manifest as excessive layout area and inefficient device organization.

\begin{figure}[t]
  \centering
  \includegraphics[width=\columnwidth]{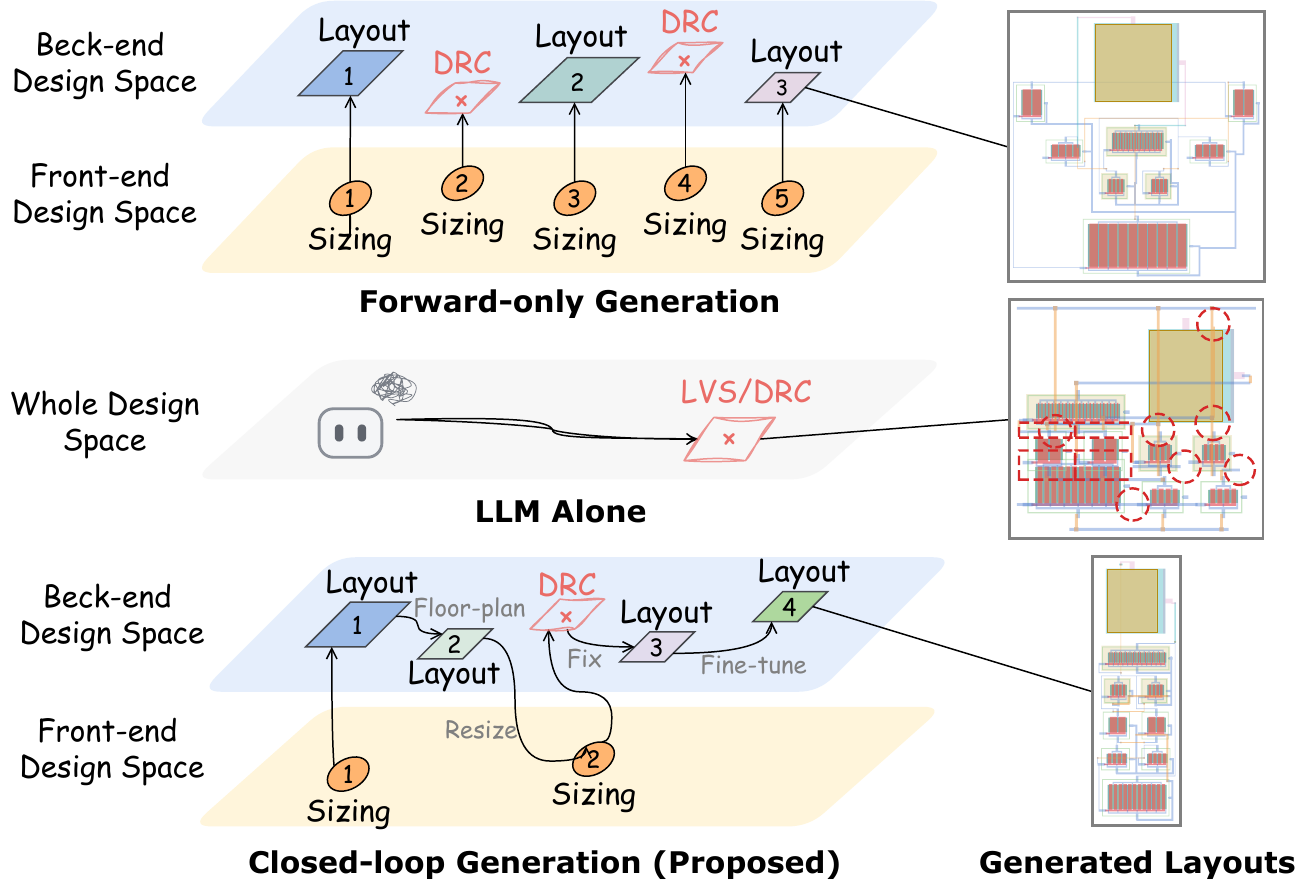}
  \caption{Forward-only generation (top) maps each sizing point to a
  layout independently and discards DRC-failing results. Generation with
  feedback (bottom) revisits floorplan, sizing, and local geometry from
  post-layout observations, yielding a more compact final layout.}
  \label{fig:intro_demo}
\end{figure}

The remaining gap lies in coordinating refinement across design stages.
MAGICAL and ALIGN~\cite{magical,align} automate placement and routing,
while DiffSP~\cite{diffsp} further improves placement optimization.
Within a forward flow, however, these advances leave upstream
sizing and device realization unchanged after layout evaluation
(Fig.~\ref{fig:intro_demo}, top).
Row-based synthesis~\cite{rowbased} couples sizing with quantized
device geometry, and post-layout-driven sizing~\cite{postbo} brings
extracted performance into parameter optimization.
Their exploration nevertheless remains bounded by the prescribed
row-height fabric or the physical choices of the layout generator.
Recent LLM-based frameworks extend stage coordination~\cite{analogmaster}
and introduce simulation-guided parameter refinement~\cite{icpi}.
Yet orchestration and parameter tuning alone do not provide direct
control over every structural decision.
When adjustments are mediated by generator parameters, a localized
physical or electrical diagnosis may not translate into a targeted
layout change.

Closing this gap requires a framework that exposes structural
changes as executable actions and revisits them using post-layout
feedback. We argue that what an optimizer can observe and decide
is as important as how effectively it searches.
Building such a framework raises two challenges:
(1) How much freedom should the optimizer have?
A narrow action space excludes useful structural alternatives,
while unrestricted changes cause combinatorial growth beyond
practical evaluation budgets.
(2) How can the framework close the optimization loop beyond
merely calling tools?
Aggregate scores cannot explain degradation.
Physical and electrical feedback must attribute limitations to
specific devices, nets, or geometric structures and guide
targeted cross-stage revisions.

We present CASCoDe, an closed-loop agentic framework for
analog sizing--layout co-design.
CASCoDe organizes revisable design decisions into a structured
action space explored through parallel tree search.
Physical and electrical feedback guide targeted revisions
(Fig.~\ref{fig:intro_demo}, bottom), while an agent-oriented execution environment supports precise
tool operations and verification.
Together, these components form a domain-specific agent harness
that connects observation, decision, and execution.
Our contributions are summarized as follows:

\begin{itemize}
\item A hierarchical action space spanning sizing, device
realization, floorplan restructuring, geometric refinement, and
routing configuration, explored by parallel tree search that combines
inexpensive geometric screening with selective post-layout evaluation.

\item Physical and electrical feedback that attributes area overhead
and performance degradation to specific structures, devices, and
nets, guiding local corrections and cross-stage revisions such as
renewed candidate selection and resizing.

\item An execution environment with precise tool interfaces,
structured trial records, and artifact consistency checks, which the
agent extends through task-specific bridging tools under fixed
acceptance criteria.

\item On 20 SKY130 amplifiers, CASCoDe is the only flow among APPlace,
MAGICAL, and ALIGN that is DRC- and LVS-clean on every circuit; it
reduces routed area by 59--85\% (mean 71\%) against APPlace at equal or
higher mean post-layout FoM, and ablations isolate the contributions
of the action space and the tree search.
\end{itemize}

\section{Related Work}
\label{sec:related}

\parhead{Circuit sizing}Sizing is commonly cast as black-box
optimization over device parameters. Bayesian optimization (BO) methods
such as MACE~\cite{mace} and the trust-region TuRBO~\cite{turbo} improve
sample efficiency, AutoCkt~\cite{autockt} applies deep reinforcement
learning (RL), and PVTSizing~\cite{pvtsizing} combines TuRBO with RL for
PVT-robust synthesis. Large language models (LLMs) now inject circuit
knowledge into this process: ADO-LLM~\cite{adollm} couples in-context
learning with BO, LEDRO~\cite{ledro} and TopoSizing~\cite{toposizing} use
LLM reasoning to reduce or structure the design space, and
AnalogCoder~\cite{analogcoder} generates circuits through code. These
methods optimize schematic-level metrics and leave layout effects to
downstream stages.

\parhead{Layout synthesis}Analog placement has been addressed by
hierarchically bounded enumeration with shape functions~\cite{plantage},
by differentiable sequence-pair optimization in DiffSP~\cite{diffsp}, and
by routing-aware placement that co-optimizes device locations and access
points~\cite{applace}. End-to-end generators such as
MAGICAL~\cite{magical} and ALIGN~\cite{align,alignprim} translate a sized
netlist into a layout through constraint extraction, primitive
generation, placement, and routing. These tools run as forward
pipelines: sizing is fixed on entry, and post-layout results are not fed
back to revise earlier decisions.

\parhead{Sizing--layout co-design}Coupling the two stages has a long
history, from simultaneous placement and module
optimization~\cite{charbon94} to constraint-based layout-driven
sizing~\cite{habal} and sizing with in-loop layout
generation~\cite{aida}. BagNet~\cite{bagnet} optimizes parameterized
layout generators against post-layout simulation with a DNN-boosted
search, Liu~et~al.~\cite{closingloop} apply BO to hierarchical layout
synthesis, post-layout-simulation-driven sizing~\cite{postbo} brings
extracted performance into parameter optimization, and row-based
synthesis~\cite{rowbased} couples sizing with height-quantized device
geometry. Budak~et~al.~\cite{ispd23} discuss the open challenges of such
joint optimization. In these flows the optimizer acts on sizing or
generator parameters, while the layout structure itself is fixed by
templates, generators, or a prescribed row fabric.

\parhead{LLM agents for analog design}LLMs have also extended automation
across stages. LayoutCopilot~\cite{layoutcopilot} uses collaborating
agents to turn designer requests into layout edits, PANDA~\cite{panda}
chains topology synthesis, sizing, placement, and routing under LLM
planning, AnalogMaster~\cite{analogmaster} generates layouts starting
from circuit images, and in-context policy improvement~\cite{icpi}
refines layout parameters from simulation feedback. Open datasets such
as LOADBench~\cite{loadbench} further support learning-based analog
design. CASCoDe differs in what the agent can act on: rather than
orchestrating tools or tuning their parameters, it exposes structural
layout decisions and the return to sizing as executable actions, and
attributes post-layout degradation to specific devices, nets, and
structures to guide each revision.

\section{CASCoDe}
\label{sec:method}
\subsection{Overview}
\label{sec:loop}
\begin{figure}[t]
    \centering
    \includegraphics[
        width=0.72\columnwidth,
    ]{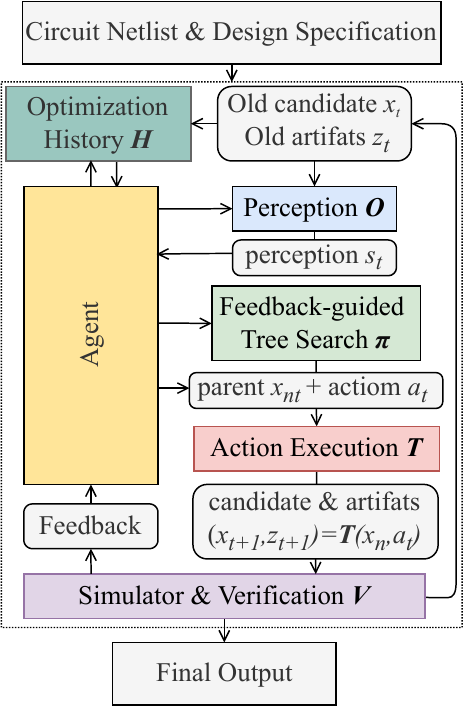}
    \caption{Closed-loop framework of CASCoDe.
    Verification $V$ checks candidate designs and their artifacts,
    while Perception $O$ organizes the accumulated evidence.
    The LLM agent implements Feedback-Guided Tree Search $\pi$,
    selecting a stored parent $x_{n_t}$ and an action sequence
    $\boldsymbol a_t$. Action Execution $T$ (detailed in
    Fig.~\ref{fig:action}) produces a new candidate $x_{t+1}$ and
    artifacts $z_{t+1}$ for subsequent evaluation.}
    \label{fig:framework}
\end{figure}

CASCoDe couples a general LLM agent with a sizer, a placer, and a router
through an execution layer (Fig.~\ref{fig:framework}). Given a circuit
netlist, testbenches, and design specifications, the
framework explores configurations $x$ specifying sizing, device
realization, placement, and routing. It minimizes routed area under a
post-layout performance constraint,
\begin{equation}
  \min_{x\in\mathcal D_{\mathrm{valid}}} A(x)
  \quad\text{subject to}\quad F_{\mathrm{post}}(x)>0,
  \label{eq:problem}
\end{equation}
where $A(x)$ is the bounding-box area of the final GDS,
$F_{\mathrm{post}}(x)$ is the FoM from RC post-layout simulation, which
is zero unless every performance metric retains at least 80\% of a
pre-layout reference (Sec.~\ref{sec:setup}), and
$\mathcal D_{\mathrm{valid}}$ contains designs satisfying physical and
functional checks. The testbenches, checks, and performance
requirements remain fixed throughout a run.

\parhead{Labor division}The provided tools perform numerical search,
physical implementation, and evaluation. Predefined actions specify
the objects and parameters the agent can modify through their
interfaces. The agent decides where to search and what to change,
schedules tool calls, and writes bridging scripts for diagnosis and
candidate generation. Pareto sizing and standardized PCell realizations
reduce the design space to structured candidates. Parameterized,
composable actions bound the agent's edits while retaining flexibility.
The same prompt and loop serve all circuits.
\begin{figure*}[t]
    \centering
    \includegraphics[width=\textwidth]{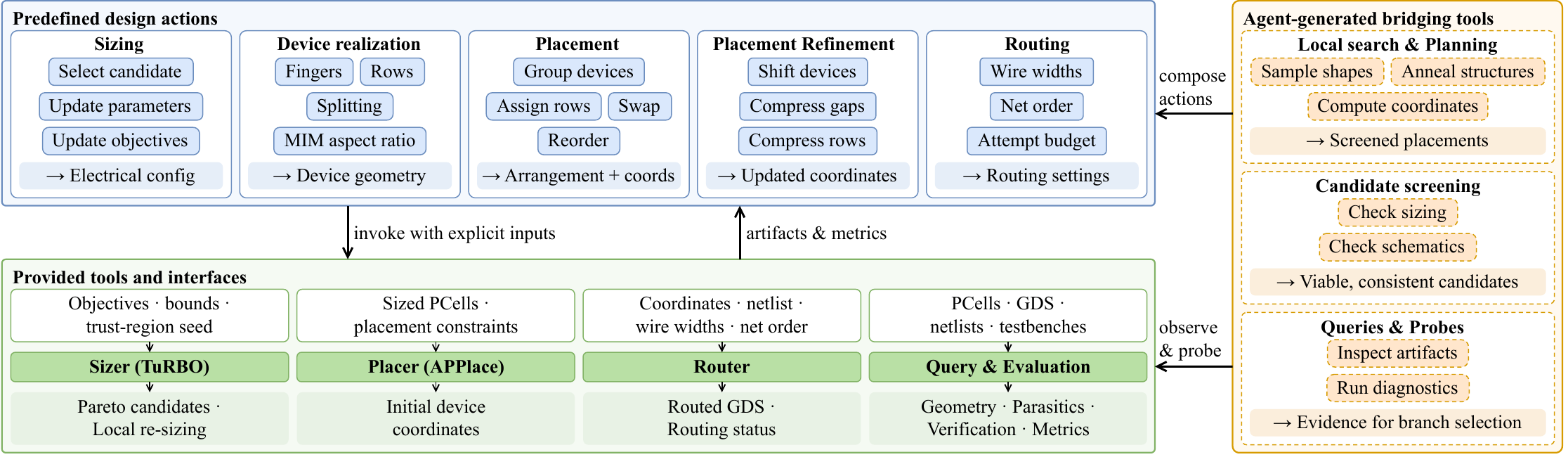}
    \caption{Action Execution $T$ in the framework of
    Fig.~\ref{fig:framework}, expanded across sizing, device
    realization, placement, routing, and post-layout evaluation.
    Structured actions are implemented through bridging scripts and
    underlying tools. Colors distinguish provided tools, reused
    scripts, and agent-generated or adapted scripts.}
    \label{fig:action}
\end{figure*}

\usetikzlibrary{arrows.meta,fit,backgrounds}

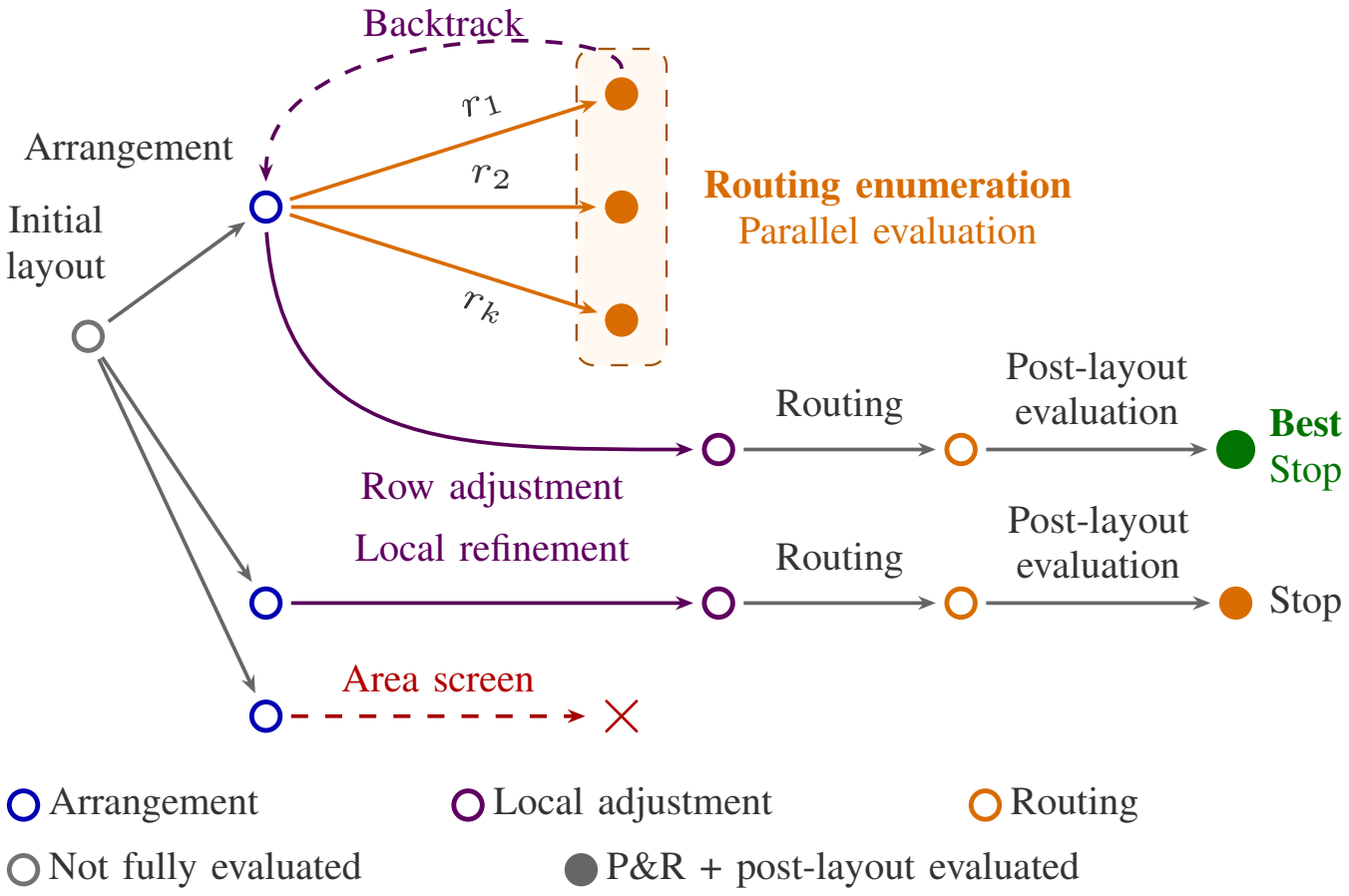
\begin{figure}[!t]
\centering
\resizebox{\columnwidth}{!}{%
\begin{tikzpicture}[
  x=1cm,y=1cm,
  state/.style={
    circle,minimum size=5pt,inner sep=0pt,line width=.85pt
  },
  initial/.style={state,draw=black!55,fill=white},
  arrangement/.style={
    state,draw=blue!70!black,fill=white
  },
  local/.style={
    state,draw=violet!75!black,fill=white
  },
  routed/.style={
    state,draw=orange!85!black,fill=white
  },
  evaluated/.style={routed,fill=orange!85!black},
  best/.style={
    state,draw=green!45!black,fill=green!45!black,
    minimum size=6pt
  },
  edge/.style={
    draw=black!60,line width=.65pt,
    -{Stealth[length=1.1mm,width=.85mm]},
    shorten >=1.5pt,shorten <=1.5pt
  },
  feedback/.style={edge,dashed,draw=violet!70!black},
  lab/.style={
    font=\fontsize{7}{8}\selectfont,
    text=black!80,align=center,inner sep=0pt
  },
  action/.style={lab,above=4pt}
]

% Initial layout and arrangement alternatives
\node[initial] (root) at (.55,0) {};
\node[lab,anchor=south] at (.35,.3) {Initial\\layout};

\node[arrangement] (a) at (1.65,.8) {};
\node[arrangement] (b) at (1.65,-1.65) {};
\node[arrangement] (c) at (1.65,-2.35) {};

\draw[edge] (root)--(a);
\draw[edge] (root)--(b);
\draw[edge] (root)--(c);

\node[lab,anchor=south] at (.8,1.03) {Arrangement};

% Enumerated routing variants, evaluated in parallel
\node[evaluated] (r1) at (3.85,1.5) {};
\node[evaluated] (r2) at (3.85,.8) {};
\node[evaluated] (rk) at (3.85,.1) {};

\draw[edge,draw=orange!85!black] (a)--
  node[lab,above=3pt,sloped,pos=.65] {$r_1$} (r1);
\draw[edge,draw=orange!85!black] (a)--
  node[lab,above=3pt,sloped,pos=.65] {$r_2$} (r2);
\draw[edge,draw=orange!85!black] (a)--
  node[lab,below=3pt,sloped,pos=.65] {$r_k$} (rk);

\begin{scope}[on background layer]
  \node[
    draw=orange!65!black,dashed,fill=orange!5,
    rounded corners=3pt,fit=(r1)(r2)(rk),
    inner xsep=5pt,inner ysep=5pt
  ] (batch) {};
\end{scope}

\node[lab,anchor=west,text=orange!85!black]
  at (4.35,.8)
  {\textbf{Routing enumeration}\\
   Parallel evaluation};

% Compact backtracking arc
\draw[feedback]
  (r1.north)
  .. controls (3.85,1.95) and (1.65,1.95) ..
  node[lab,above=2pt,pos=.5,text=violet!75!black]
  {Backtrack}
  (a.north);

% New branch after row adjustment
\node[local]  (row) at (4.45,-.7) {};
\node[routed] (rr)  at (5.95,-.7) {};
\node[best]   (win) at (7.65,-.7) {};

\draw[edge,draw=violet!70!black]
  (a.south)
  .. controls (1.75,-.7) and (2.8,-.7) ..
  (row.west);

\node[lab,anchor=north,text=violet!75!black]
  at (3.05,-.83) {Row adjustment};

\draw[edge] (row)--
  node[action] {Routing} (rr);
\draw[edge] (rr)--
  node[action] {Post-layout\\evaluation} (win);

\node[lab,anchor=west,text=green!45!black]
  at (7.85,-.7) {\textbf{Best}\\Stop};

% Another trajectory: local refinement
\node[local]     (refine) at (4.45,-1.65) {};
\node[routed]    (rl)     at (5.95,-1.65) {};
\node[evaluated] (vl)     at (7.65,-1.65) {};

\draw[edge,draw=violet!70!black]
  (b.east)
  .. controls (2.35,-1.65) and (2.9,-1.65) ..
  (refine.west);

\node[lab,anchor=south,text=violet!75!black]
  at (3.05,-1.4) {Local refinement};

\draw[edge] (refine)--
  node[action] {Routing} (rl);
\draw[edge] (rl)--
  node[action] {Post-layout\\evaluation} (vl);

\node[lab,anchor=west] at (7.85,-1.65) {Stop};

% Area-screened branch
\node[text=red!70!black,font=\large,inner sep=0pt]
  (pruned) at (3.85,-2.35) {$\times$};

\draw[edge,dashed,draw=red!65!black] (c)--
  node[action,text=red!70!black] {Area screen} (pruned);

% Legend: action levels
\node[arrangement] at (.15,-2.9) {};
\node[lab,anchor=west] at (.3,-2.9) {Arrangement};

\node[local] at (2.9,-2.9) {};
\node[lab,anchor=west] at (3.05,-2.9) {Local adjustment};

\node[routed] at (6.1,-2.9) {};
\node[lab,anchor=west] at (6.25,-2.9) {Routing};

% Legend: evaluation status
\node[initial] at (.15,-3.3) {};
\node[lab,anchor=west] at (.3,-3.3) {Not fully evaluated};

\node[state,draw=black!60,fill=black!60] at (3.6,-3.3) {};
\node[lab,anchor=west] at (3.75,-3.3)
  {P\&R + post-layout evaluated};

\end{tikzpicture}%
}
\caption{Tree search over the back-end design space.}
\label{fig:search_tree}
\end{figure}

\parhead{Framework components}The execution layer provides
\emph{Perception} $O$, \emph{Action Execution} $T$, and
\emph{Verification} $V$. The agent implements
\emph{Feedback-Guided Tree Search} $\pi$. Let $\mathcal H_t$ denote
the search history before expansion $t$, including stored designs,
parent links, actions, and tool results. One candidate expansion is
\begin{equation}
\begin{aligned}
  s_t &= O(\mathcal H_t),\\
  (n_t,\boldsymbol a_t) &= \pi(s_t,\mathcal H_t),\\
  (x_{t+1},z_{t+1}) &= T(x_{n_t},\boldsymbol a_t),\\
  v_{t+1} &= V(x_{t+1},z_{t+1}).
\end{aligned}
\label{eq:agent_loop}
\end{equation}
Here $s_t$ is the observation, $n_t$ identifies a stored parent, and
$\boldsymbol a_t$ is an ordered action sequence. Execution produces a
candidate $x_{t+1}$ and artifacts $z_{t+1}$, including layout, netlists,
and simulation outputs. Verification returns check outcomes and
performance in $v_{t+1}$. The new record extends $\mathcal H_t$.
Diagnostic requests can enrich the history before an expansion, and
independent expansions can execute in parallel.

\subsection{Perception}
\label{sec:perception}

Perception $O$ organizes the artifacts in $\mathcal H_t$ into
observations $s_t$ at three levels. The agent can supplement these
observations with targeted probes.

\parhead{Design-level outcomes}Area, FoM, circuit metrics, and check
outcomes summarize each candidate. Comparing the sizing schematic,
realized schematic, capacitance-only extraction, and RC extraction
helps separate realization effects from capacitive and resistive
degradation. Parent--child comparisons reveal the benefit of each edit.

\parhead{Device-level geometry}PCell footprints, pins, and keep-outs
expose packing constraints. Comparing the placed-cell envelope with
the routed GDS localizes unused space and routing overhang. Geometric
queries associate violations with nearby devices or gaps.

\parhead{Net-level parasitics}Extracted nodes are mapped to circuit
nets. Capacitance, resistive paths, and device-terminal voltage drops
identify connections that warrant electrical diagnosis.

\parhead{Active probing}The agent invokes or writes scripts to obtain
targeted observations. For example, a resistive probe scales the
extracted resistances $R_n$ of net $n$ by $0<\kappa\ll1$ and measures
\begin{equation}
  \Delta F_n =
  F_{\mathrm{post}}(x\mid R_n\leftarrow\kappa R_n)
  - F_{\mathrm{post}}(x).
  \label{eq:resistive_probe}
\end{equation}
A positive response indicates potential benefit from reducing that
net's resistance and guides routing or placement edits. Probe results
enrich $\mathcal H_t$. Verification uses the unmodified extracted circuit.

\subsection{Action Execution}
\label{sec:actions}

Action Execution $T$ realizes an action sequence on a selected parent
and returns the resulting configuration and artifacts. It connects
provided tools, predefined design actions, and agent-generated
bridging tools (Fig.~\ref{fig:action}).

\parhead{Provided tools}The sizer returns Pareto candidates and
supports trust-region re-sizing after realization changes. The placer
supplies initial arrangements using standardized PCells. The router
accepts explicit coordinates and routing configurations. These fast
solvers support repeated evaluation. Script interfaces expose PCell
geometry, layout queries, and configuration updates.

\parhead{Structured actions}An action $a=(\omega,\mathcal I,\theta)$
specifies an operation $\omega$, target objects $\mathcal I$, and
parameters $\theta$. The sequence
$\boldsymbol a_t=(a_{t,1},\ldots,a_{t,k_t})$ combines $k_t$ such
edits into one candidate. For example, device splitting can be followed
by row reassignment, a swap, and gap compression. Execution applies
these edits to $x_{n_t}$ and invokes the tools needed to implement and
evaluate the result. Configuration editing and tool execution thus
remain distinct within $T$.

\parhead{Bridging tools}The agent builds floorplan generators,
local searchers, and diagnostic probes as needed. Local searchers use
randomized sampling or simulated annealing to explore row assignments,
group orderings, and device shapes. A geometry planner converts these
choices into exact coordinates and checks packing constraints.
Geometric ranking retains a small set for physical evaluation,
\[
  \text{Local search}
  \;\longrightarrow\;\text{Plan and screen}
  \;\longrightarrow\;\text{P\&R}.
\]
The agent defines the search scope and selects candidates using
feedback. Generators extend $T$ and probes extend $O$, while $V$
remains fixed. These scripts can be reused across branches and
promoted into the reusable tool set for later runs.

\subsection{Feedback-Guided Tree Search}
\label{sec:search}

Feedback-Guided Tree Search $\pi$ selects the parent $n_t$ and action
sequence $\boldsymbol a_t$ in Eq.~\eqref{eq:agent_loop}. Exploration
proceeds through floorplan branches, placement variants, and routing
attempts (Fig.~\ref{fig:search_tree}). Parent links and results in
$\mathcal H_t$ support expansion, comparison, and backtracking.

\parhead{Floorplan branches}Functionally screened sizing candidates
seed alternative floorplan forms. The agent assesses their ideal
packing potential using PCell geometry. For configuration $x$ and
floorplan form $f$, let $B_i(x,f)$ be the footprint of device $i$
positioned by the planner under the specified spacing constraints.
The estimated placement area is
\begin{equation}
  \widehat A_{\mathrm{plan}}(x,f)=
  \operatorname{area}\!\left[
    \operatorname{bbox}\!\left(\bigcup_i B_i(x,f)\right)
  \right].
  \label{eq:minarea}
\end{equation}
This estimate covers rows, stacked devices, and side columns within
one representation. Compact realizations indicate the area reduction
available before routing. The agent prioritizes promising forms
relative to the incumbent area $A^\star$ and retains alternatives as
separate branches. Routed area $A(x)$ determines actual improvement.

\parhead{Placement and routing expansion}Local search and agent-directed
edits produce placement variants within each branch. Realization
changes are screened by
$|F_{\mathrm{real}}-F_{\mathrm{size}}|\le\epsilon_{\mathrm{real}}$,
where the two FoMs correspond to the realized and sizing schematics.
Only shortlisted configurations proceed to full evaluation. Each
placement supports parallel routing attempts with different wire
widths and net orders. Promising attempts are compared using their
own extracted circuits and performance checks. Independent branches,
placement variants, and routing attempts can execute in parallel.

\parhead{Feedback and backtracking}Perception turns completed
results into evidence for the next expansion. Excessive whitespace
motivates placement refinement, routing overhang motivates placement
or routing changes, and performance loss motivates probes,
realization changes, or local re-sizing. If an attempt is infeasible
or falls short of expectations, the agent returns to a stored parent
and explores another modification. Repeated failures or exhausted
improvement opportunities terminate that branch and redirect the
search to another placement or floorplan. Gap compression is judged
after routing because routing-grid effects can offset geometric gains.

\parhead{Verification and termination}\label{sec:verification}
Verification $V$ evaluates candidates using DRC, LVS, functional
checks, and RC post-layout simulation under the fixed protocol.
Search stops when the budget is exhausted or two consecutive rounds
yield no qualifying area improvement. The selected design is rebuilt
and verified before delivery.
\section{Experimental Results}
\label{sec:results}

% Preamble:
% \usepackage{booktabs,amssymb,bm,adjustbox,xcolor}

\begin{table*}[t]
\centering
\caption{Layout and post-layout performance comparison. Area is in $\mu\mathrm{m}^{2}$.}
\label{tab:layout_comparison}
\begingroup
\definecolor{failbg}{HTML}{F6D5D0}
\setlength{\tabcolsep}{3.5pt}
\renewcommand{\arraystretch}{1.18}
\scriptsize
\begin{adjustbox}{max width=\textwidth}
\begin{tabular}{@{}c*{4}{cccc}@{}}
\toprule
\textbf{Circuit}
& \multicolumn{4}{c}{\textbf{MAGICAL}\textsuperscript{*}}
& \multicolumn{4}{c}{\textbf{ALIGN}\textsuperscript{$\dagger$}}
& \multicolumn{4}{c}{\textbf{APPlace}\textsuperscript{$\ddagger$}}
& \multicolumn{4}{c}{\textbf{CASCoDe}} \\
\cmidrule(lr){2-5}\cmidrule(lr){6-9}
\cmidrule(lr){10-13}\cmidrule(lr){14-17}
& DRC/LVS & Area & FoM & Density
& DRC/LVS & Area & FoM & Density
& DRC/LVS & Area & FoM & Density
& DRC/LVS & Area & FoM & Density \\
\midrule
1s\_5t
& $84/\checkmark$ & 1,050 & \textbf{7.24} & 0.16
& $0/\checkmark$ & 549 & 6.85 & 0.37
& $0/\checkmark$ & 884 & 6.94 & 0.36
& $0/\checkmark$ & \textbf{233}{\tiny\,$\pm$10\%} & 6.64 & \textbf{0.71} \\
1s\_fcse
& $254/\checkmark$ & 3,162 & \cellcolor{failbg}6.21 & 0.35
& $17/\times$ & 2,102 & \cellcolor{failbg}5.71\textsuperscript{$\dagger$} & 0.52
& $0/\checkmark$ & 2,186 & 6.71 & 0.54
& $0/\checkmark$ & \textbf{806}{\tiny\,$\pm$11\%} & \textbf{7.10} & \textbf{0.71} \\
1s\_fcfd
& $120/\checkmark$ & 13,411 & 6.75 & 0.33
& $114/\checkmark$ & 11,530 & \cellcolor{failbg}6.63 & 0.40
& $0/\checkmark$ & 14,270 & \textbf{6.79} & 0.41
& $0/\checkmark$ & \textbf{4,054}{\tiny\,$\pm$7\%} & 6.66 & \textbf{0.61} \\
1s\_iwil
& $150/\checkmark$ & 2,719 & 6.94 & 0.25
& $20/\checkmark$ & 1,885 & 6.96 & 0.34
& $0/\checkmark$ & 2,454 & \textbf{6.97} & 0.33
& $0/\checkmark$ & \textbf{447}{\tiny\,$\pm$9\%} & 6.93 & \textbf{0.64} \\
1s\_ntele
& $207/\checkmark$ & 1,887 & \textbf{6.97} & 0.20
& $63/\checkmark$ & 1,481 & \textbf{6.97} & 0.28
& $0/\checkmark$ & 1,559 & \textbf{6.97} & 0.34
& $0/\checkmark$ & \textbf{455}{\tiny\,$\pm$17\%} & 6.76 & \textbf{0.72} \\
1s\_nwil
& $166/\checkmark$ & 2,705 & 6.70 & 0.41
& $134/\checkmark$ & 1,680 & 6.81 & 0.57
& $0/\checkmark$ & 2,969 & 6.82 & 0.34
& $0/\checkmark$ & \textbf{541}{\tiny\,$\pm$14\%} & \textbf{6.93} & \textbf{0.82} \\
1s\_sym
& $60/\checkmark$ & 11,373 & \textbf{7.05} & 0.37
& $97/\times$ & 5,805 & \cellcolor{failbg}5.32\textsuperscript{$\dagger$} & 0.59
& $0/\checkmark$ & 8,701 & 6.99 & 0.39
& $0/\checkmark$ & \textbf{1,326}{\tiny\,$\pm$6\%} & 6.97 & \textbf{0.70} \\
1s\_tele
& $226/\checkmark$ & 1,974 & 6.95 & 0.20
& $88/\checkmark$ & 1,148 & 6.94 & 0.33
& $0/\checkmark$ & 1,429 & \textbf{6.97} & 0.42
& $0/\checkmark$ & \textbf{587}{\tiny\,$\pm$12\%} & 6.69 & \textbf{0.68} \\
1s\_twil
& $226/\checkmark$ & 2,650 & \textbf{6.92} & 0.25
& $88/\checkmark$ & 1,967 & 6.75 & 0.41
& $0/\checkmark$ & 2,513 & 6.81 & 0.45
& $0/\checkmark$ & \textbf{1,040}{\tiny\,$\pm$5\%} & 6.79 & \textbf{0.64} \\
1s\_wsc
& $193/\checkmark$ & 4,802 & \cellcolor{failbg}6.66 & 0.31
& $58/\times$ & 3,137 & \cellcolor{failbg}4.98\textsuperscript{$\dagger$} & 0.52
& $0/\checkmark$ & 4,425 & \textbf{6.75} & 0.39
& $0/\checkmark$ & \textbf{774}{\tiny\,$\pm$9\%} & 6.71 & \textbf{0.74} \\
\cmidrule(lr){1-17}
\textbf{1s Summary}
& 0\% & 4,573 & 80\% & 0.28
& 10\% & 3,128 & 60\% & 0.43
& \textbf{100\%} & 4,139 & \textbf{100\%} & 0.40
& \textbf{100\%} & \textbf{1,026}{\tiny\,$\pm$10\%} & \textbf{100\%} & \textbf{0.70} \\
\midrule
2s\_se
& $112/\checkmark$ & 4,014 & \cellcolor{failbg}6.31 & 0.35
& $48/\checkmark$ & 1,935 & 6.81 & 0.55
& $0/\checkmark$ & 3,183 & 6.87 & 0.41
& $0/\checkmark$ & \textbf{877}{\tiny\,$\pm$10\%} & \textbf{7.12} & \textbf{0.89} \\
2s\_miller
& $447/\checkmark$ & 4,238 & 6.79 & 0.29
& $47/\checkmark$ & 4,462 & \textbf{6.84} & 0.46
& $0/\checkmark$ & 4,535 & 6.75 & 0.32
& $0/\checkmark$ & \textbf{1,347}{\tiny\,$\pm$7\%} & 6.74 & \textbf{0.76} \\
2s\_nfc
& $363/\checkmark$ & 9,921 & \cellcolor{failbg}1.99 & 0.33
& $250/\times$ & 4,544 & \cellcolor{failbg}6.31\textsuperscript{$\dagger$} & 0.61
& $0/\checkmark$ & 8,682 & \cellcolor{failbg}6.66 & 0.34
& $0/\checkmark$ & \textbf{3,195}{\tiny\,$\pm$4\%} & \textbf{6.93} & \textbf{0.73} \\
2s\_pcas
& $337/\times$ & 3,726 & \cellcolor{failbg}4.31\textsuperscript{*} & 0.26
& $384/\checkmark$ & 1,884 & \cellcolor{failbg}6.68 & 0.43
& $0/\checkmark$ & 2,780 & \cellcolor{failbg}6.65 & 0.34
& $0/\checkmark$ & \textbf{802}{\tiny\,$\pm$5\%} & \textbf{6.70} & \textbf{0.81} \\
2s\_wsc
& $236/\checkmark$ & 22,532 & \cellcolor{failbg}5.30 & 0.45
& $134/\checkmark$ & 3,320 & 6.94 & 0.53
& $0/\checkmark$ & 10,479 & 6.79 & 0.27
& $0/\checkmark$ & \textbf{1,934}{\tiny\,$\pm$6\%} & \textbf{6.99} & \textbf{0.75} \\
\cmidrule(lr){1-17}
\textbf{2s Summary}
& 0\% & 8,886 & 20\% & 0.34
& 0\% & 3,229 & 60\% & 0.52
& \textbf{100\%} & 5,932 & 60\% & 0.34
& \textbf{100\%} & \textbf{1,631}{\tiny\,$\pm$6\%} & \textbf{100\%} & \textbf{0.79} \\
\midrule
3s\_acbc
& $214/\checkmark$ & 13,519 & \cellcolor{failbg}5.89 & 0.37
& $24/\checkmark$ & 7,237 & \cellcolor{failbg}5.87 & 0.60
& $20/\checkmark$ & 14,095 & \cellcolor{failbg}6.23 & 0.37
& $0/\checkmark$ & \textbf{5,736}{\tiny\,$\pm$3\%} & \textbf{7.71} & \textbf{0.79} \\
3s\_affc
& $269/\times$ & 36,928 & \cellcolor{failbg}4.89\textsuperscript{*} & 0.25
& $124/\checkmark$ & 11,757 & \cellcolor{failbg}6.37 & 0.63
& $34/\checkmark$ & 16,636 & \cellcolor{failbg}6.48 & 0.51
& $0/\checkmark$ & \textbf{5,453}{\tiny\,$\pm$3\%} & \textbf{7.16} & \textbf{0.79} \\
3s\_dacfc
& $307/\times$ & 29,526 & \cellcolor{failbg}5.63\textsuperscript{*} & 0.41
& $137/\checkmark$ & 20,807 & \cellcolor{failbg}6.16 & 0.48
& $41/\times$ & 39,913 & \cellcolor{failbg}6.52\textsuperscript{$\ddagger$} & 0.32
& $0/\checkmark$ & \textbf{8,425}{\tiny\,$\pm$4\%} & \textbf{7.02} & \textbf{0.64} \\
3s\_smc
& $377/\checkmark$ & 27,281 & \cellcolor{failbg}5.32 & 0.28
& $15/\checkmark$ & 8,215 & \textbf{7.45} & 0.51
& $16/\checkmark$ & 12,915 & 6.77 & 0.44
& $0/\checkmark$ & \textbf{5,219}{\tiny\,$\pm$5\%} & 6.96 & \textbf{0.85} \\
3s\_tcfc
& -- & -- & -- & --
& $250/\checkmark$ & 20,548 & \textbf{7.09} & 0.55
& $27/\checkmark$ & 24,699 & \cellcolor{failbg}6.69 & 0.50
& $0/\checkmark$ & \textbf{5,989}{\tiny\,$\pm$4\%} & 6.79 & \textbf{0.69} \\
\cmidrule(lr){1-17}
\textbf{3s Summary}
& 0\% & 26,814 & 0\% & 0.33
& 0\% & 13,713 & 40\% & 0.55
& 0\% & 21,652 & 20\% & 0.43
& \textbf{100\%} & \textbf{6,164}{\tiny\,$\pm$4\%} & \textbf{100\%} & \textbf{0.75} \\
\midrule
\textbf{All (20)}
& 0\% & 10,390 & 45\% & 0.31
& 5\% & 5,800 & 55\% & 0.48
& 75\% & 8,965 & 70\% & 0.39
& \textbf{100\%} & \textbf{2,462}{\tiny\,$\pm$8\%} & \textbf{100\%} & \textbf{0.74} \\
\textbf{Ratio}
& & 1.16 & & 0.79
& & 0.65 & & 1.23
& & 1.00 & & 1.00
& & \textbf{0.27} & & \textbf{1.90} \\
\bottomrule
\end{tabular}
\end{adjustbox}
\par\vspace{6pt}
\begin{minipage}{\textwidth}
\footnotesize
DRC/LVS shows the DRC violation count and LVS result
($\checkmark$ pass, $\times$ fail); -- denotes unavailable data.
Shaded FoM cells fail the pass criterion of Sec.~\ref{sec:setup}.

\textsuperscript{*}\textsuperscript{$\dagger$}\textsuperscript{$\ddagger$}
FoM after minimal LVS repair (MAGICAL, ALIGN, APPlace). MAGICAL produced
no layout for \texttt{3s\_tcfc}.

Summary rows give the share of clean (zero-DRC, LVS-passing) layouts
under DRC/LVS and of passing layouts under FoM.
CASCoDe values are means of three runs, whose shares count every run;
$\pm$ is the range of routed area over the runs relative to the mean.
Area and density are arithmetic means over available layouts; Ratio
normalizes mean area and density to APPlace.
Bold marks the best area, FoM, and density in each row and the highest
shares in each summary (ties included).
\end{minipage}
\endgroup
\end{table*}
% Regenerate with:
%   python scripts/make_main_table_figure.py tab/main_tab_1.tex figures/fig_main_table.pdf
\begin{figure*}[t]
\centering
\includegraphics[width=\textwidth]{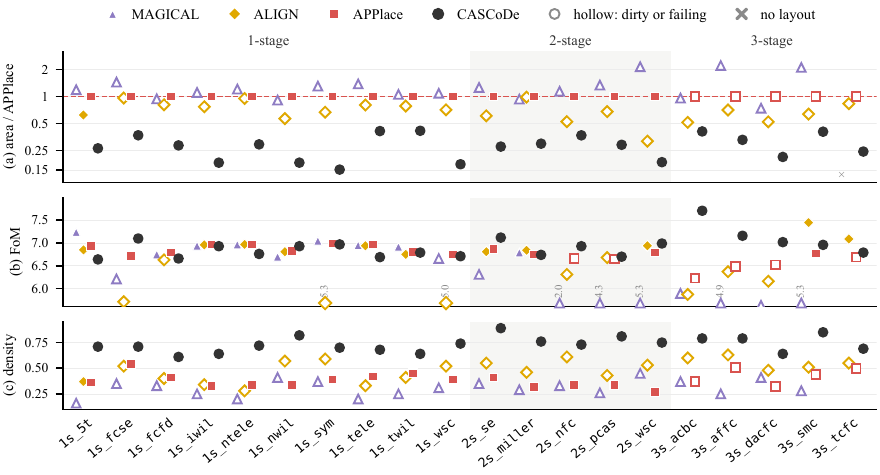}
\caption{Visual summary of Table~\ref{tab:layout_comparison} for every
circuit: (a) routed area relative to APPlace (log scale), (b) post-layout
FoM, and (c) device density. Hollow markers in (a) and (c) are layouts
that are not DRC- and LVS-clean; in (b) they fail the performance
requirement, which is checked per metric, so a hollow marker can sit
above a filled one. FoM values below the plotted range are pinned to its
floor and printed. Shaded: two-stage class.}
\label{fig:main_table}
\end{figure*}

\subsection{Experimental Setup}
\label{sec:setup}

\parhead{Implementation}The agent is Claude Opus 5.5 at high reasoning
effort, running on an 8-core Ryzen~9 9950X3D workstation with 16\,GB of
memory. All layouts are in SKY130, and post-layout metrics are simulated with ngspice-43
on RC-extracted netlists.

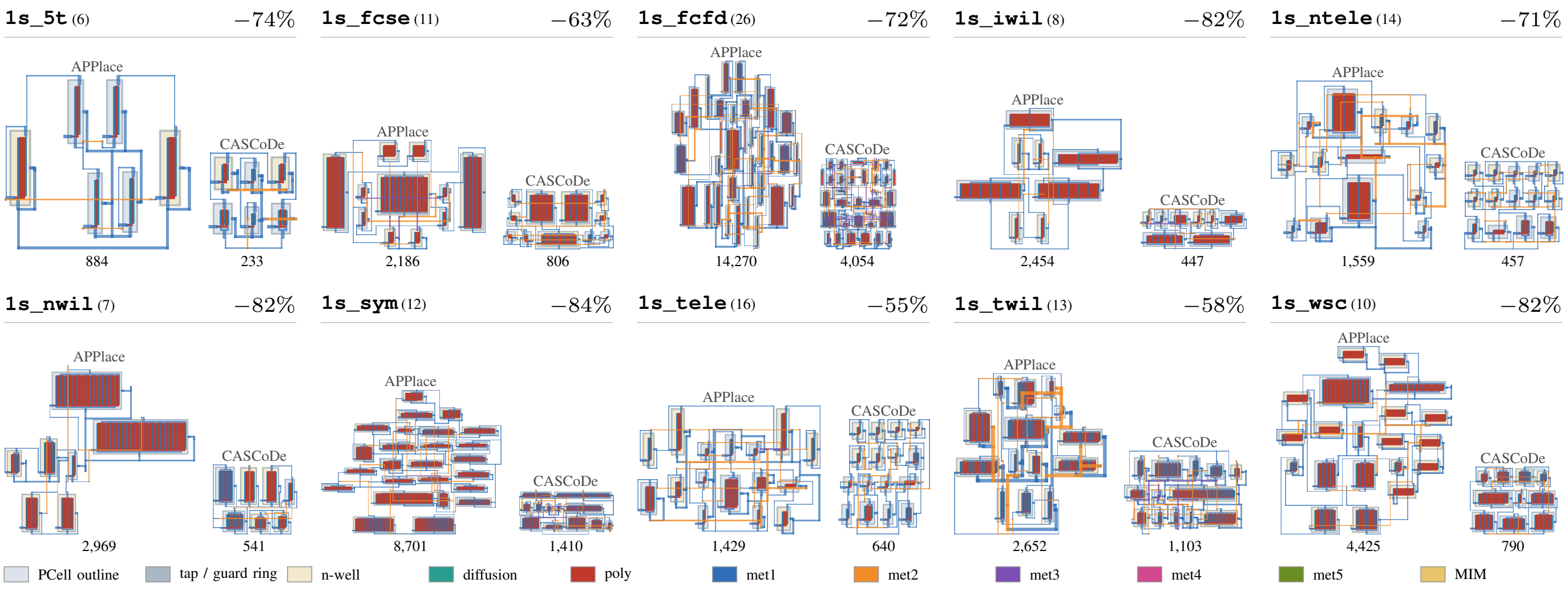
\begin{figure*}[t]
\centering
\begin{tikzpicture}[x=1pt,y=1pt,font=\tiny,lab/.style={inner sep=0pt,anchor=south,text=black!70},ar/.style={inner sep=0pt,anchor=north}]
\node[inner sep=0pt,anchor=north west,font=\scriptsize] at (0.00,0.00) {\textbf{\texttt{1s\_5t}}\,{\tiny(6)}};
\node[inner sep=0pt,anchor=north east,font=\scriptsize] at (96.40,0.00) {$-74\%$};
\draw[black!25,line width=0.3pt] (0.00,-9.00) -- (96.40,-9.00);
\node[inner sep=0pt,anchor=south west] at (0.00,-79.00) {\includegraphics[width=61.30pt]{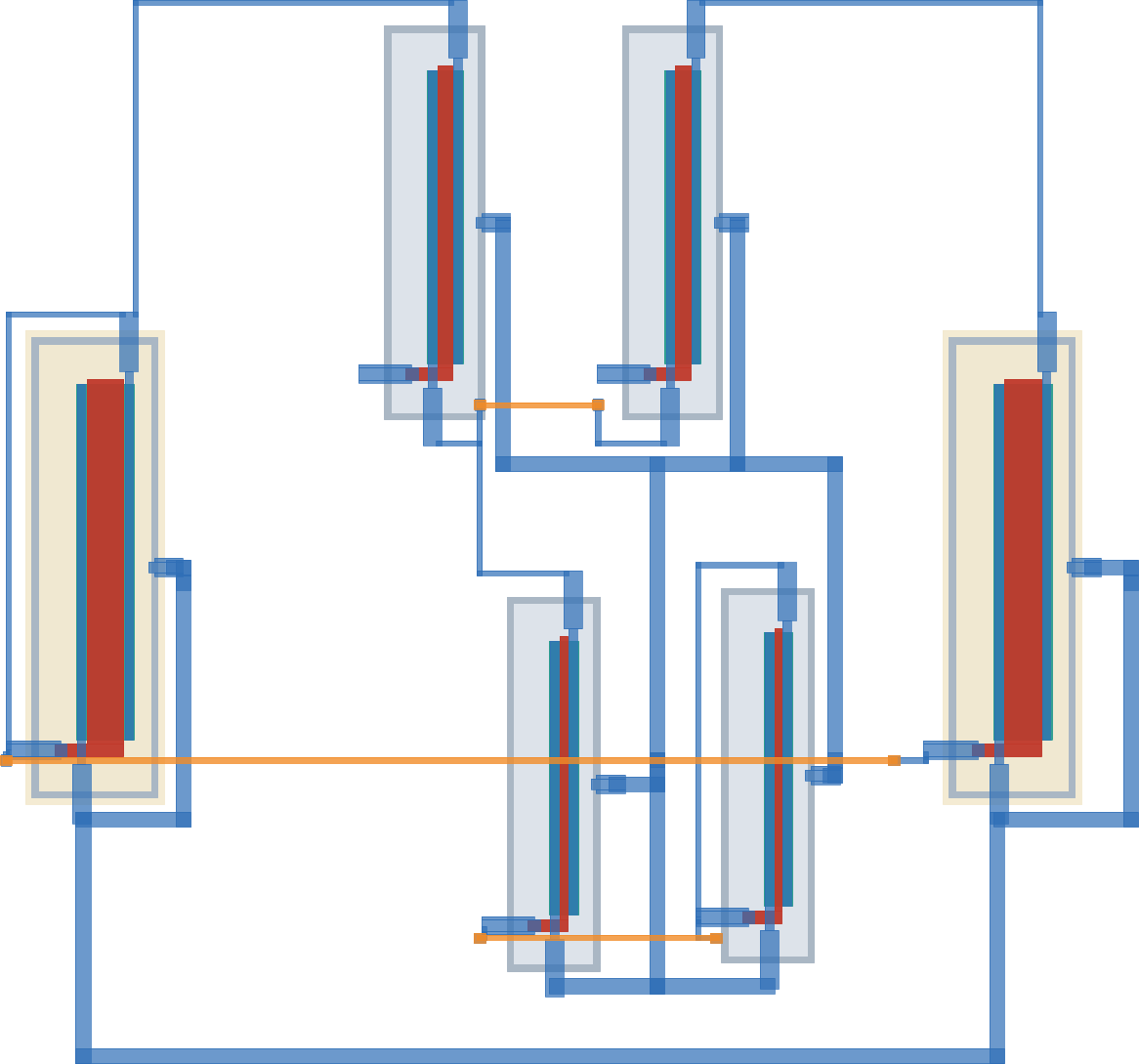}};
\node[lab] at (30.65,-20.52) {APPlace};
\node[ar] at (30.65,-80.80) {884};
\node[inner sep=0pt,anchor=south west] at (67.30,-79.00) {\includegraphics[width=29.10pt]{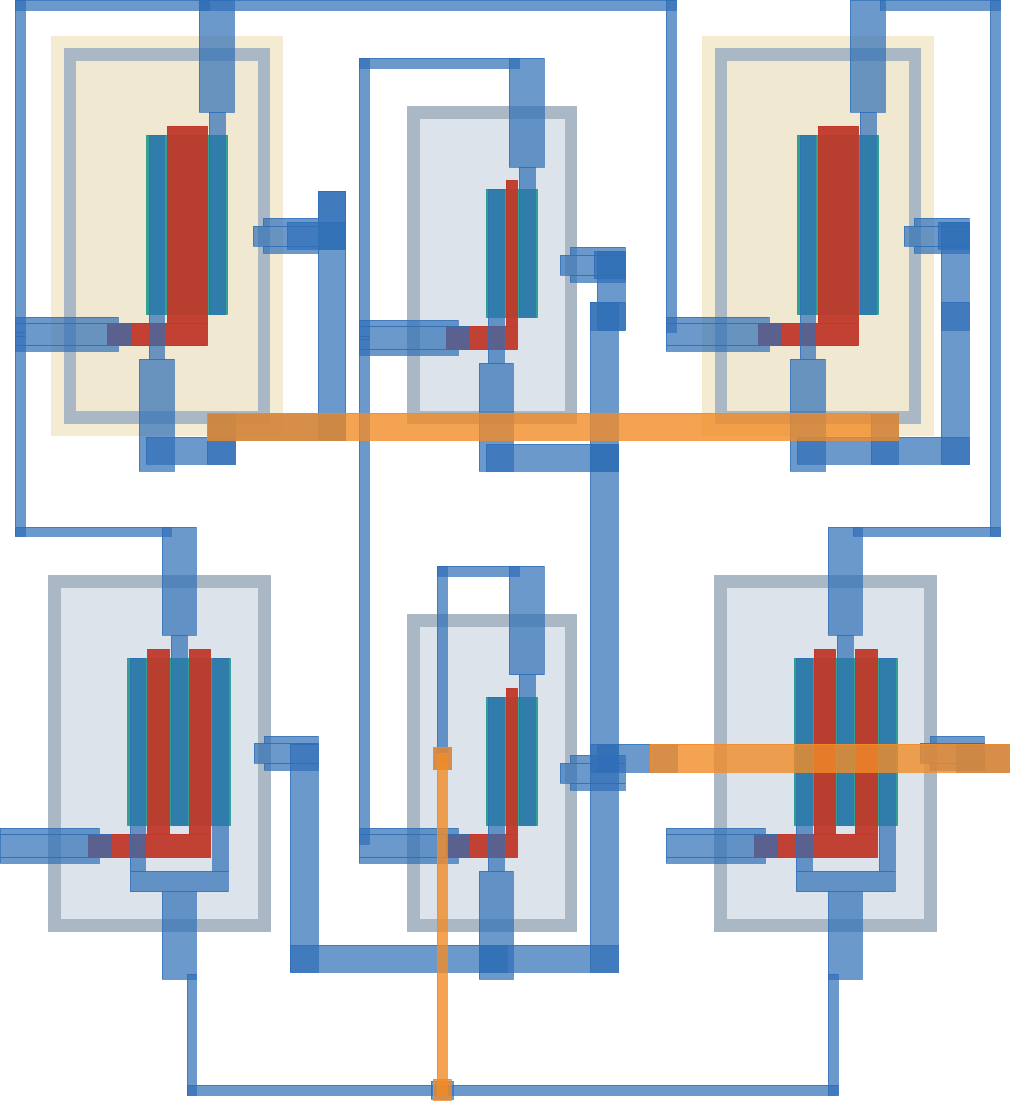}};
\node[lab] at (81.85,-46.05) {CASCoDe};
\node[ar] at (81.85,-80.80) {233};
\node[inner sep=0pt,anchor=north west,font=\scriptsize] at (104.40,0.00) {\textbf{\texttt{1s\_fcse}}\,{\tiny(11)}};
\node[inner sep=0pt,anchor=north east,font=\scriptsize] at (200.80,0.00) {$-63\%$};
\draw[black!25,line width=0.3pt] (104.40,-9.00) -- (200.80,-9.00);
\node[inner sep=0pt,anchor=south west] at (104.40,-79.00) {\includegraphics[width=54.13pt]{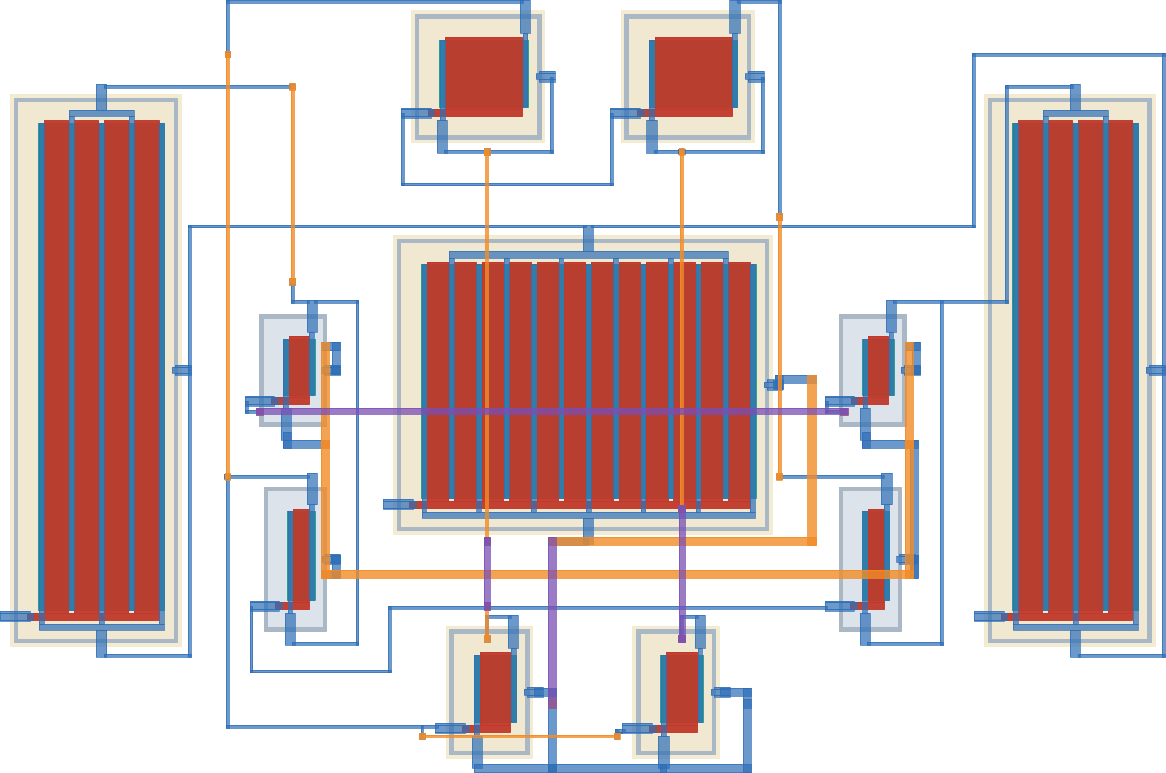}};
\node[lab] at (131.47,-41.90) {APPlace};
\node[ar] at (131.47,-80.80) {2{,}186};
\node[inner sep=0pt,anchor=south west] at (164.53,-79.00) {\includegraphics[width=36.27pt]{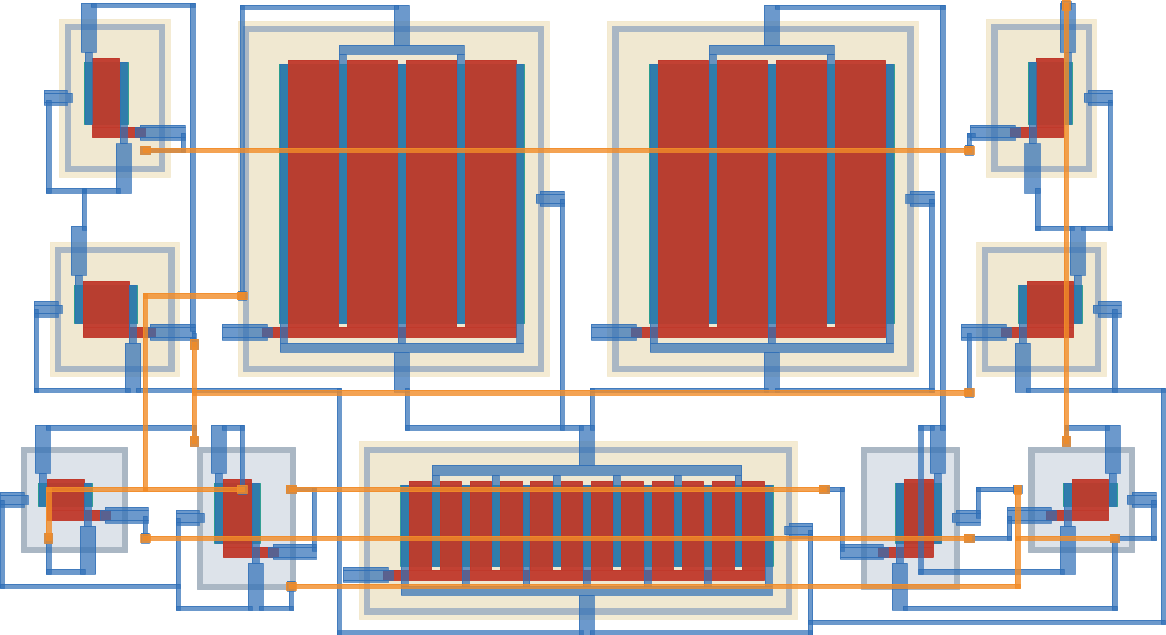}};
\node[lab] at (182.67,-58.04) {CASCoDe};
\node[ar] at (182.67,-80.80) {806};
\node[inner sep=0pt,anchor=north west,font=\scriptsize] at (208.80,0.00) {\textbf{\texttt{1s\_fcfd}}\,{\tiny(26)}};
\node[inner sep=0pt,anchor=north east,font=\scriptsize] at (305.20,0.00) {$-72\%$};
\draw[black!25,line width=0.3pt] (208.80,-9.00) -- (305.20,-9.00);
\node[inner sep=0pt,anchor=south west] at (219.94,-79.00) {\includegraphics[width=42.95pt]{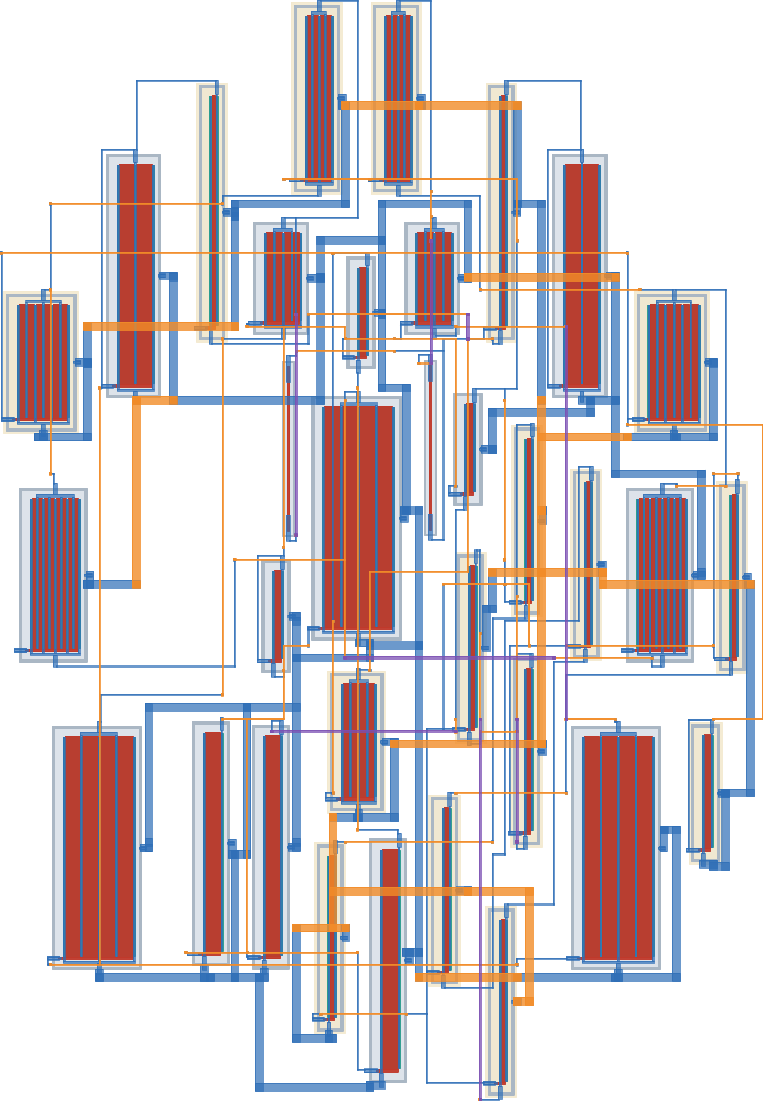}};
\node[lab] at (241.42,-15.80) {APPlace};
\node[ar] at (241.42,-80.80) {14{,}270};
\node[inner sep=0pt,anchor=south west] at (268.90,-79.00) {\includegraphics[width=25.16pt]{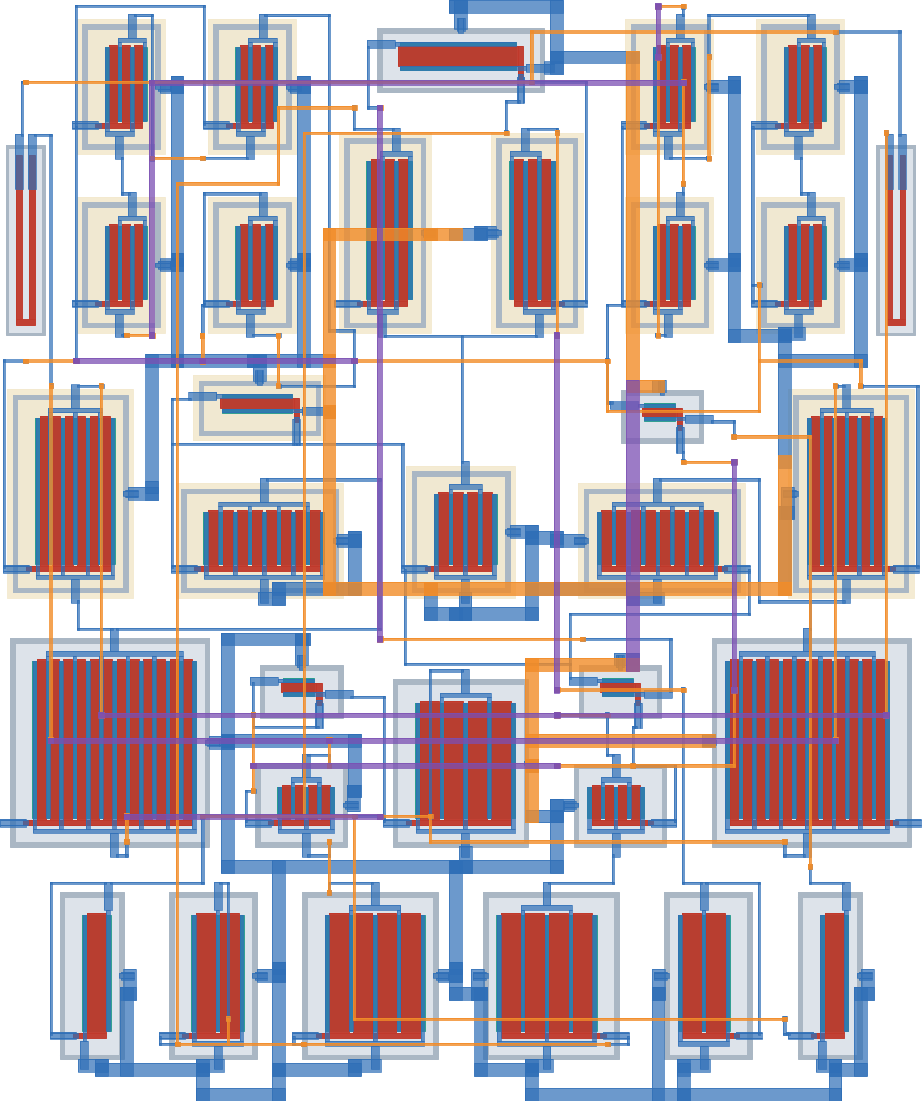}};
\node[lab] at (281.48,-47.73) {CASCoDe};
\node[ar] at (281.48,-80.80) {4{,}054};
\node[inner sep=0pt,anchor=north west,font=\scriptsize] at (313.20,0.00) {\textbf{\texttt{1s\_iwil}}\,{\tiny(8)}};
\node[inner sep=0pt,anchor=north east,font=\scriptsize] at (409.60,0.00) {$-82\%$};
\draw[black!25,line width=0.3pt] (313.20,-9.00) -- (409.60,-9.00);
\node[inner sep=0pt,anchor=south west] at (313.20,-79.00) {\includegraphics[width=55.30pt]{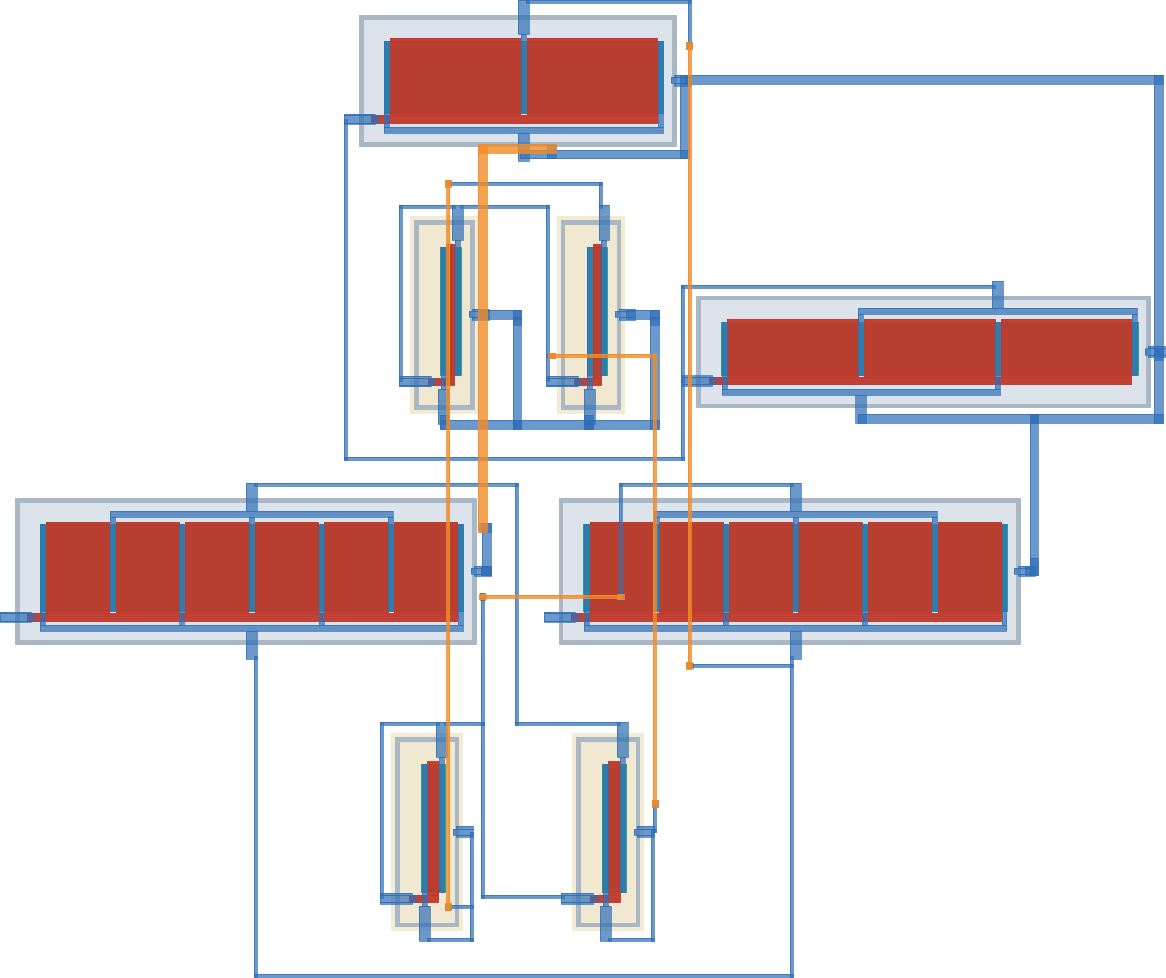}};
\node[lab] at (340.85,-31.41) {APPlace};
\node[ar] at (340.85,-80.80) {2{,}454};
\node[inner sep=0pt,anchor=south west] at (374.50,-79.00) {\includegraphics[width=35.10pt]{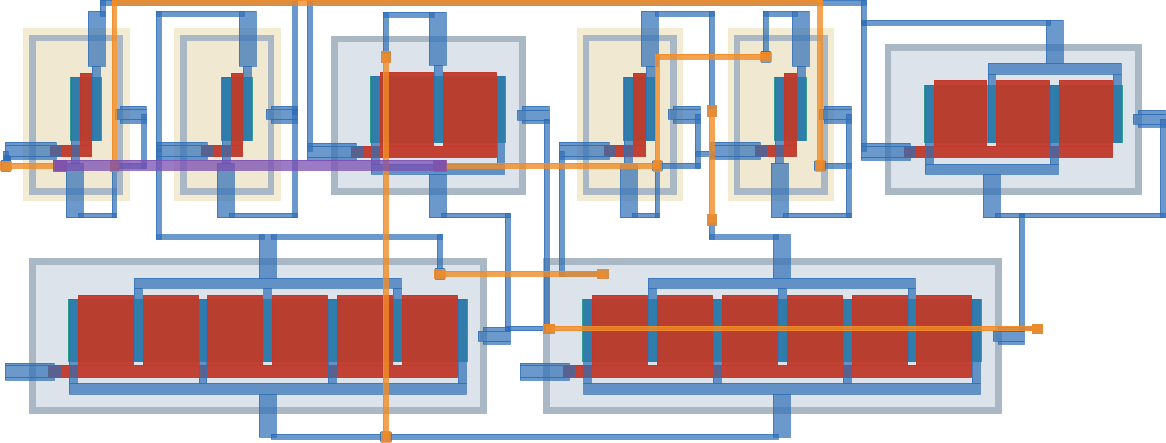}};
\node[lab] at (392.05,-64.48) {CASCoDe};
\node[ar] at (392.05,-80.80) {447};
\node[inner sep=0pt,anchor=north west,font=\scriptsize] at (417.60,0.00) {\textbf{\texttt{1s\_ntele}}\,{\tiny(14)}};
\node[inner sep=0pt,anchor=north east,font=\scriptsize] at (514.00,0.00) {$-71\%$};
\draw[black!25,line width=0.3pt] (417.60,-9.00) -- (514.00,-9.00);
\node[inner sep=0pt,anchor=south west] at (417.60,-79.00) {\includegraphics[width=57.83pt]{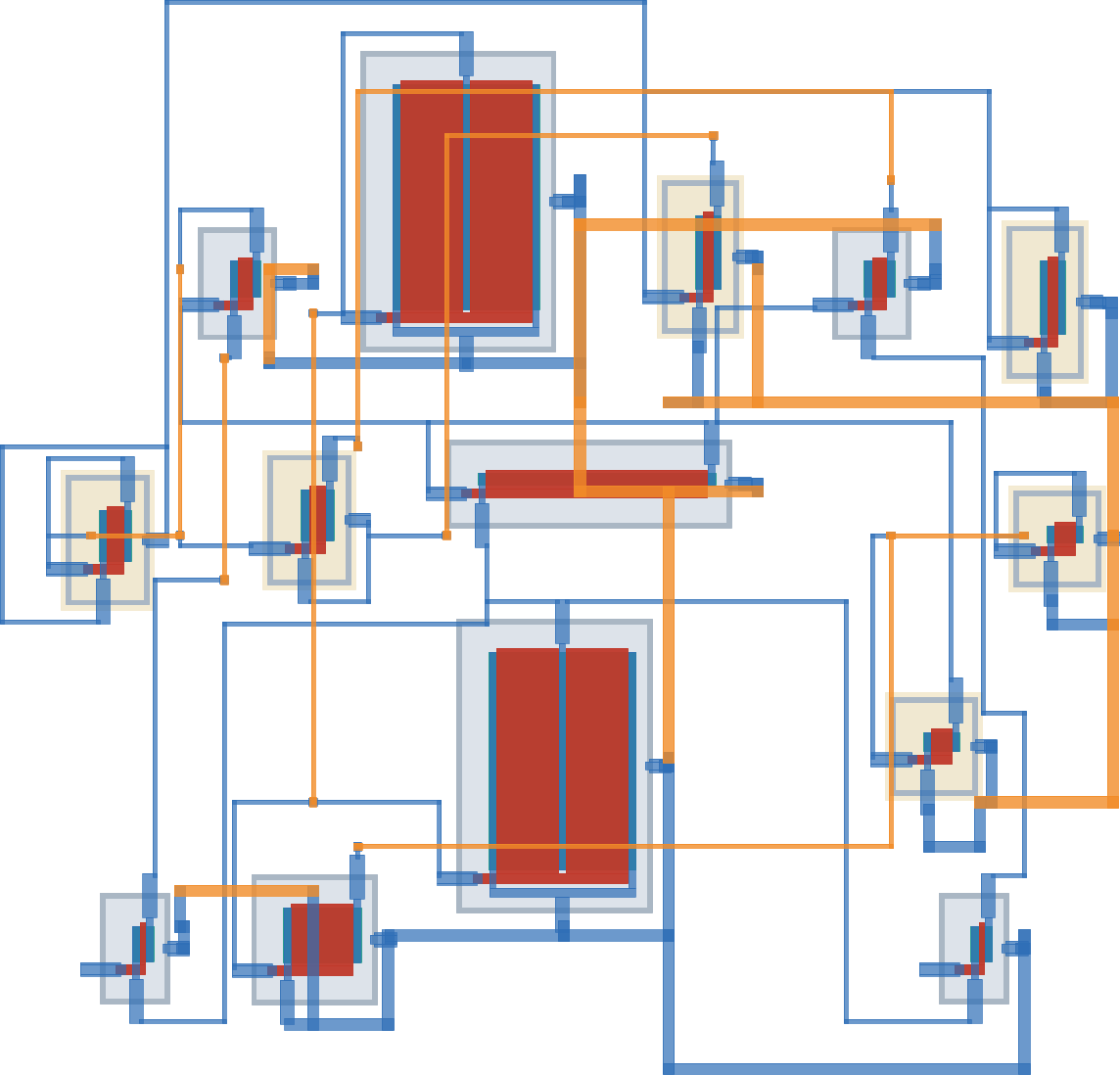}};
\node[lab] at (446.51,-22.33) {APPlace};
\node[ar] at (446.51,-80.80) {1{,}559};
\node[inner sep=0pt,anchor=south west] at (481.43,-79.00) {\includegraphics[width=32.57pt]{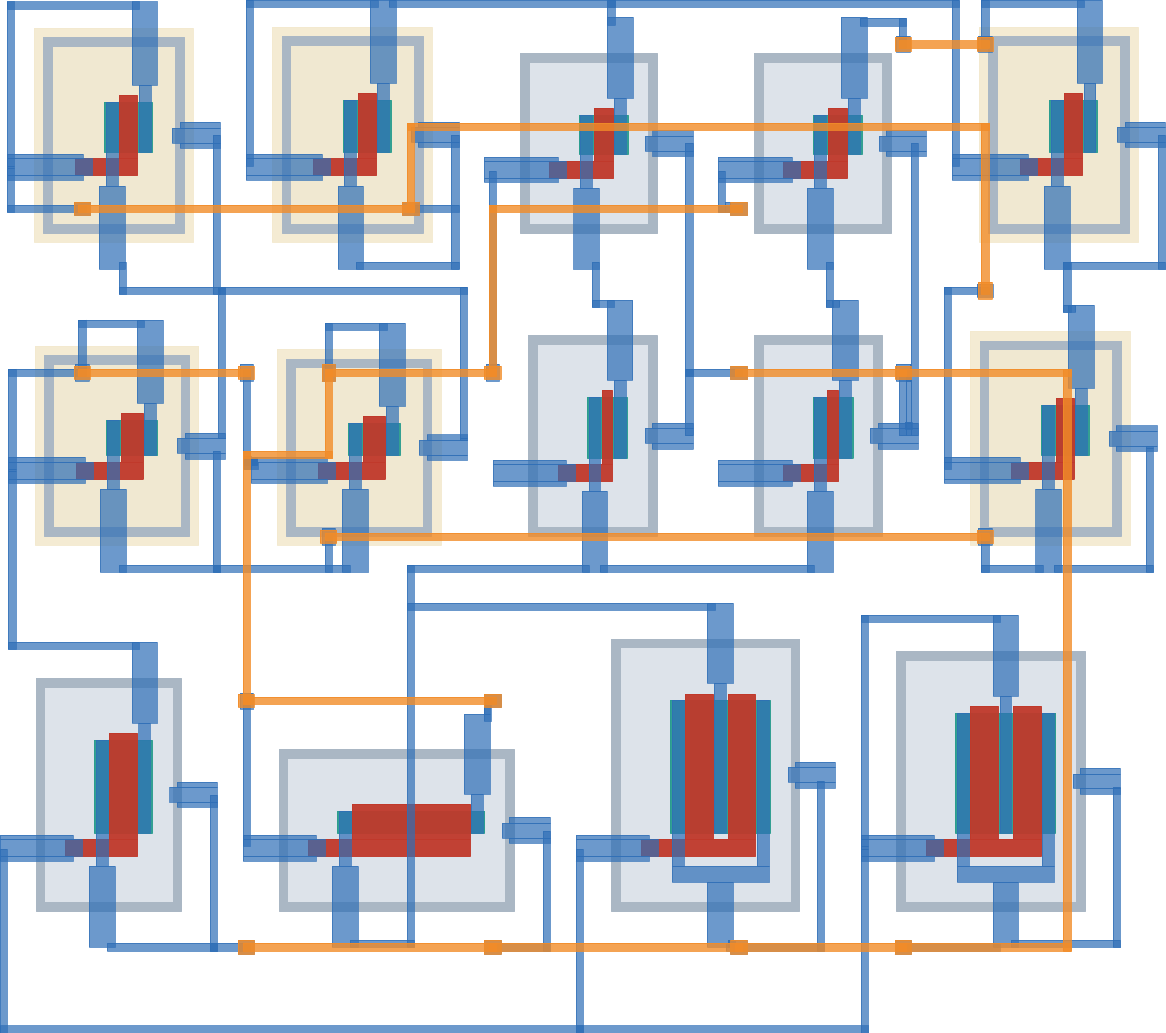}};
\node[lab] at (497.71,-48.93) {CASCoDe};
\node[ar] at (497.71,-80.80) {457};
\node[inner sep=0pt,anchor=north west,font=\scriptsize] at (0.00,-94.00) {\textbf{\texttt{1s\_nwil}}\,{\tiny(7)}};
\node[inner sep=0pt,anchor=north east,font=\scriptsize] at (96.40,-94.00) {$-82\%$};
\draw[black!25,line width=0.3pt] (0.00,-103.00) -- (96.40,-103.00);
\node[inner sep=0pt,anchor=south west] at (0.00,-173.00) {\includegraphics[width=62.66pt]{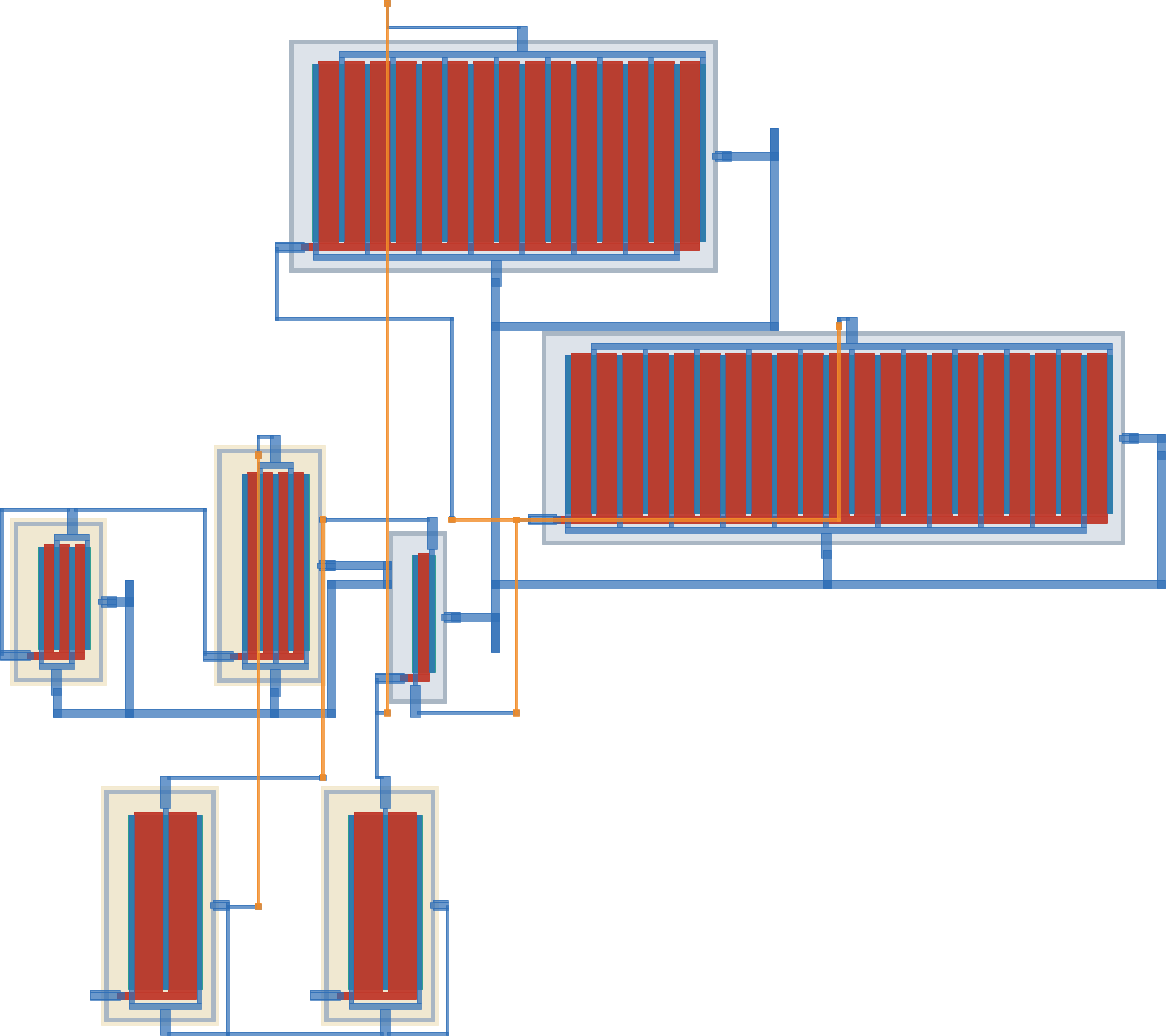}};
\node[lab] at (31.33,-116.12) {APPlace};
\node[ar] at (31.33,-174.80) {2{,}969};
\node[inner sep=0pt,anchor=south west] at (68.66,-173.00) {\includegraphics[width=27.74pt]{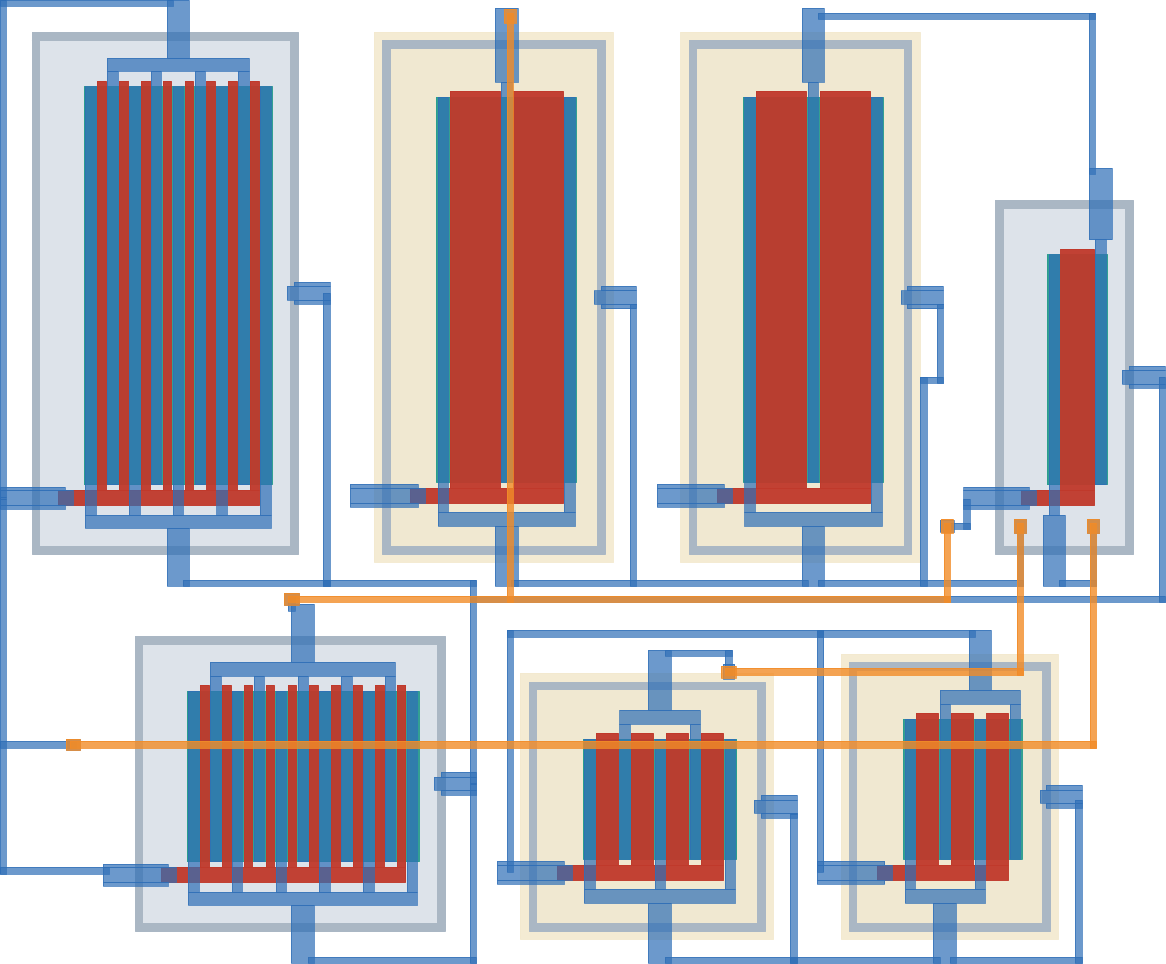}};
\node[lab] at (82.53,-148.87) {CASCoDe};
\node[ar] at (82.53,-174.80) {541};
\node[inner sep=0pt,anchor=north west,font=\scriptsize] at (104.40,-94.00) {\textbf{\texttt{1s\_sym}}\,{\tiny(12)}};
\node[inner sep=0pt,anchor=north east,font=\scriptsize] at (200.80,-94.00) {$-84\%$};
\draw[black!25,line width=0.3pt] (104.40,-103.00) -- (200.80,-103.00);
\node[inner sep=0pt,anchor=south west] at (104.40,-173.00) {\includegraphics[width=59.29pt]{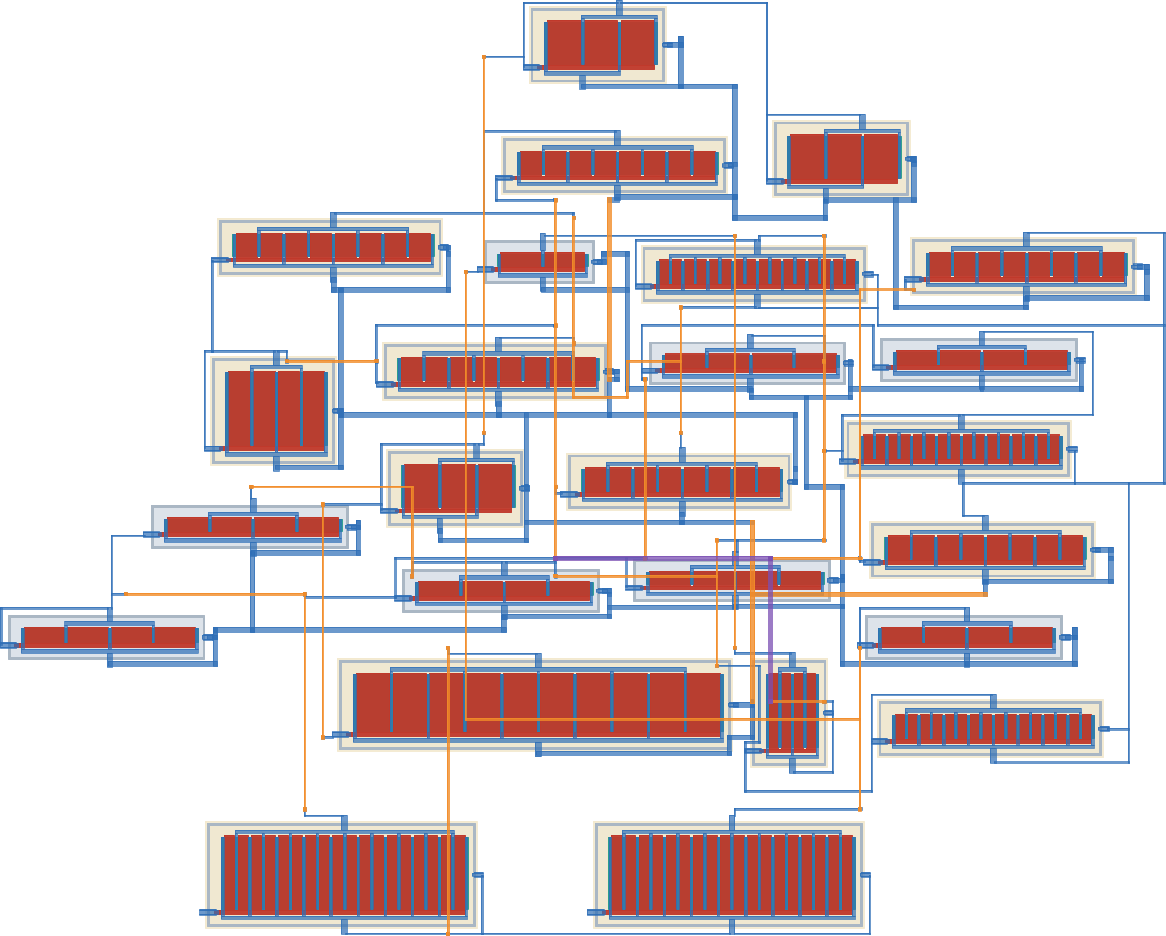}};
\node[lab] at (134.05,-124.17) {APPlace};
\node[ar] at (134.05,-174.80) {8{,}701};
\node[inner sep=0pt,anchor=south west] at (169.69,-173.00) {\includegraphics[width=31.11pt]{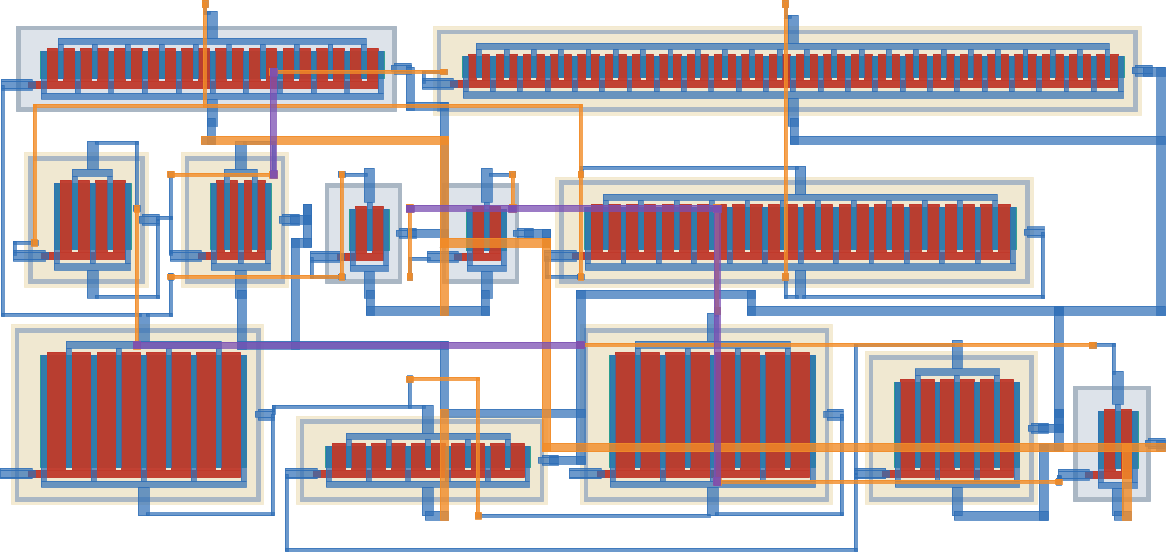}};
\node[lab] at (185.25,-157.08) {CASCoDe};
\node[ar] at (185.25,-174.80) {1{,}410};
\node[inner sep=0pt,anchor=north west,font=\scriptsize] at (208.80,-94.00) {\textbf{\texttt{1s\_tele}}\,{\tiny(16)}};
\node[inner sep=0pt,anchor=north east,font=\scriptsize] at (305.20,-94.00) {$-55\%$};
\draw[black!25,line width=0.3pt] (208.80,-103.00) -- (305.20,-103.00);
\node[inner sep=0pt,anchor=south west] at (208.80,-173.00) {\includegraphics[width=60.26pt]{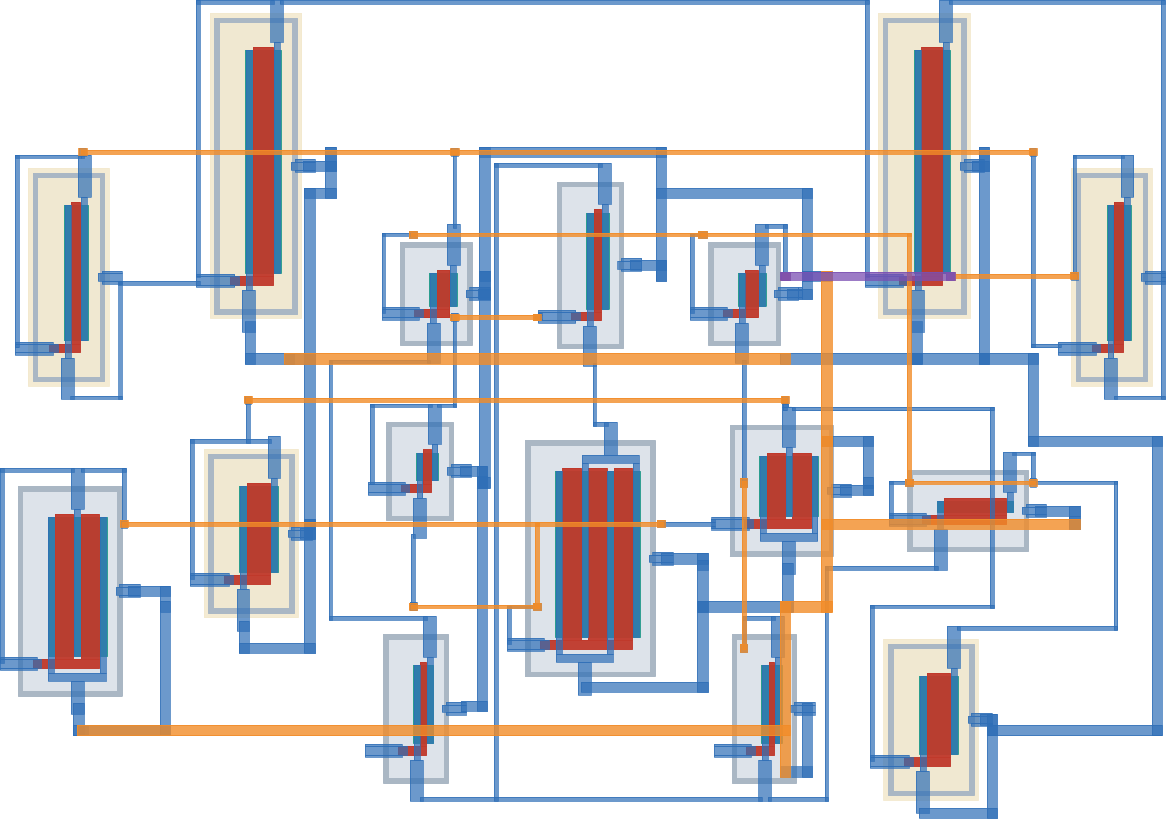}};
\node[lab] at (238.93,-129.48) {APPlace};
\node[ar] at (238.93,-174.80) {1{,}429};
\node[inner sep=0pt,anchor=south west] at (275.06,-173.00) {\includegraphics[width=30.14pt]{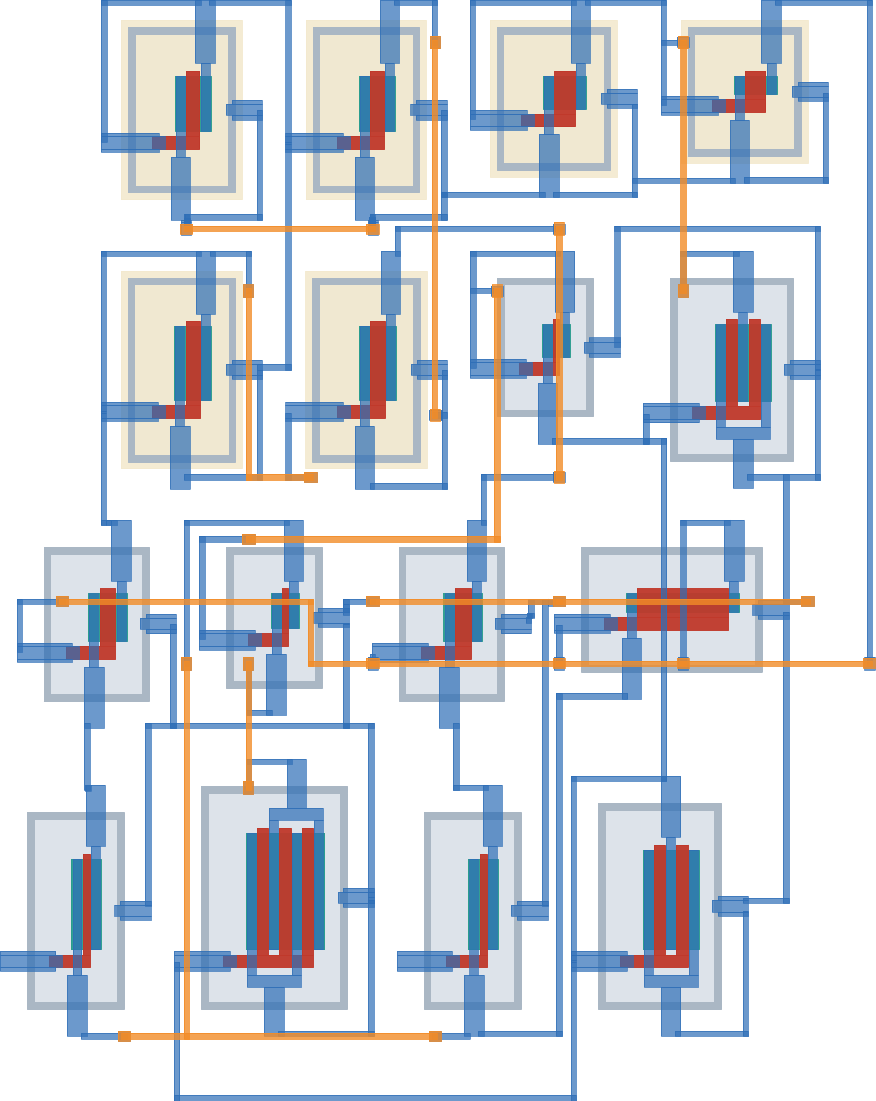}};
\node[lab] at (290.13,-133.91) {CASCoDe};
\node[ar] at (290.13,-174.80) {640};
\node[inner sep=0pt,anchor=north west,font=\scriptsize] at (313.20,-94.00) {\textbf{\texttt{1s\_twil}}\,{\tiny(13)}};
\node[inner sep=0pt,anchor=north east,font=\scriptsize] at (409.60,-94.00) {$-58\%$};
\draw[black!25,line width=0.3pt] (313.20,-103.00) -- (409.60,-103.00);
\node[inner sep=0pt,anchor=south west] at (313.20,-173.00) {\includegraphics[width=50.16pt]{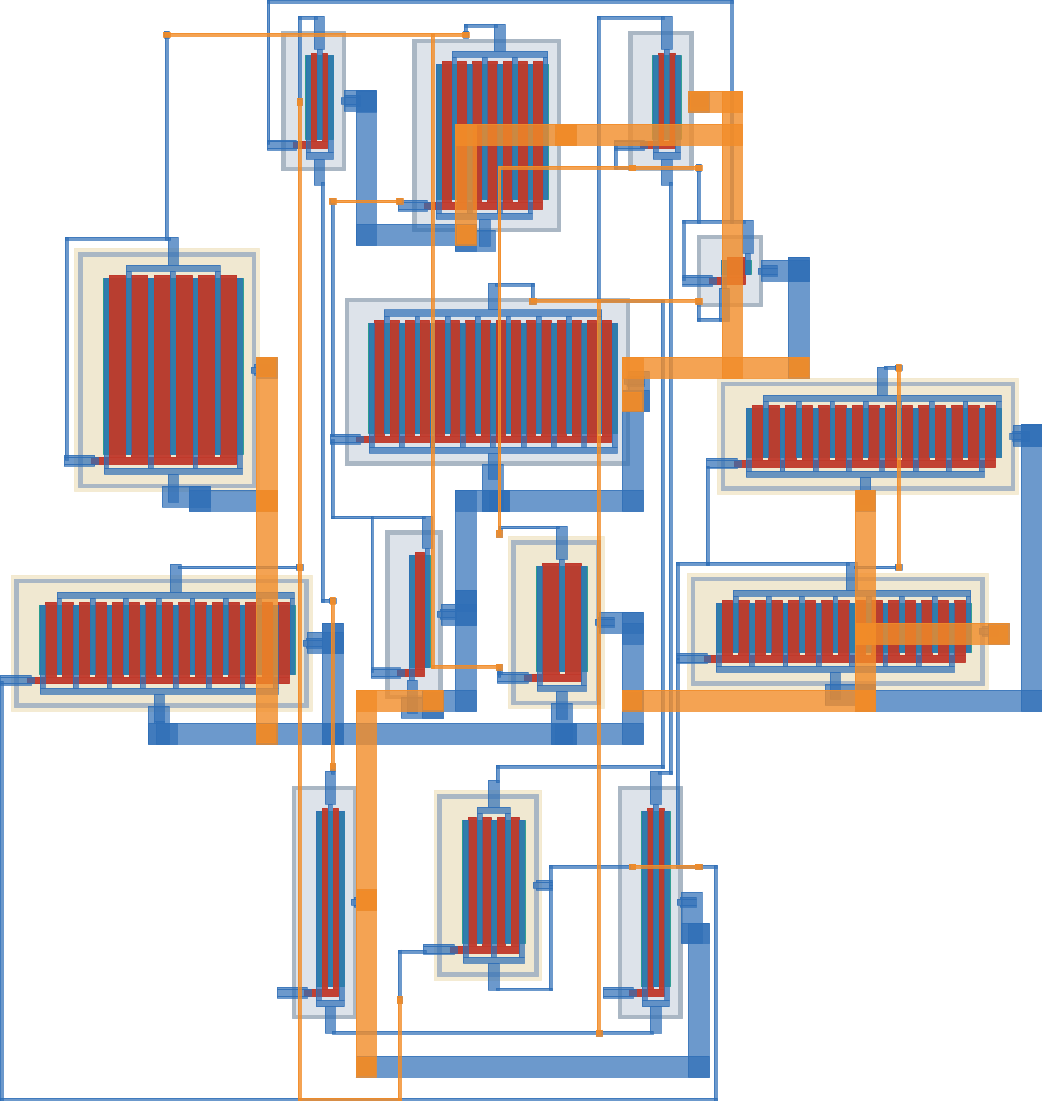}};
\node[lab] at (338.28,-118.80) {APPlace};
\node[ar] at (338.28,-174.80) {2{,}652};
\node[inner sep=0pt,anchor=south west] at (369.36,-173.00) {\includegraphics[width=40.24pt]{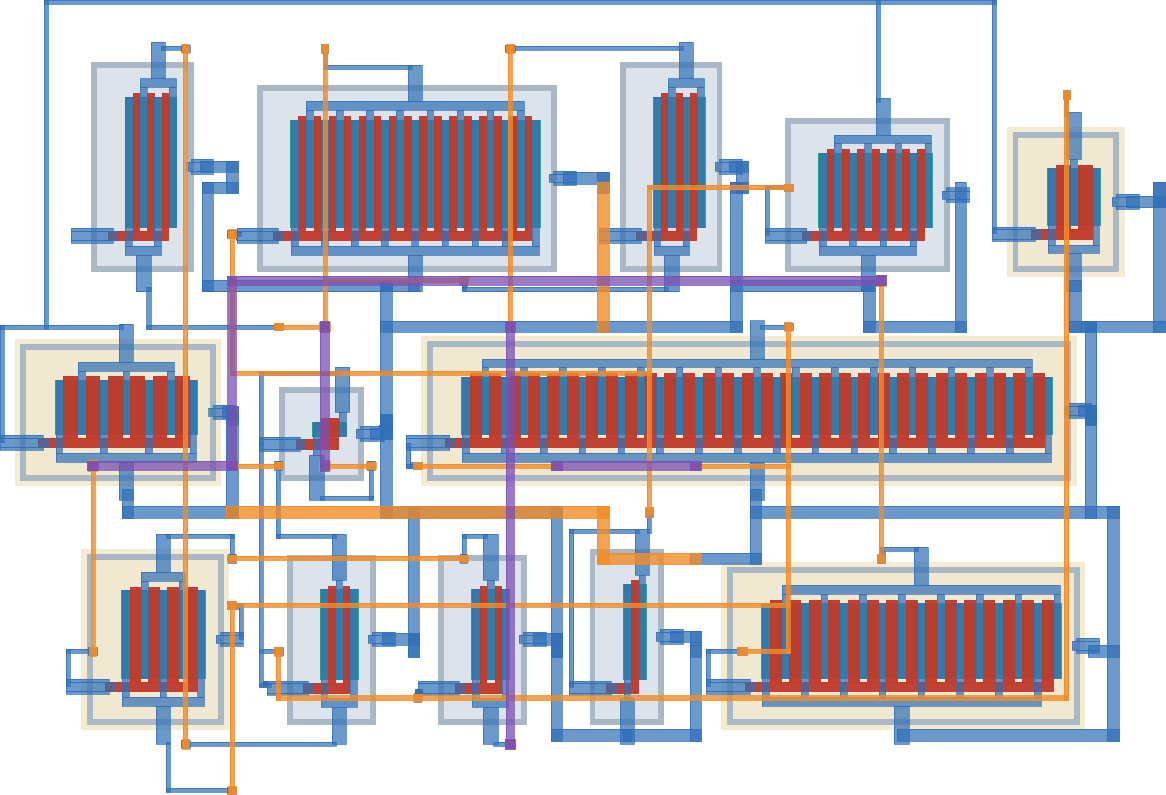}};
\node[lab] at (389.48,-144.34) {CASCoDe};
\node[ar] at (389.48,-174.80) {1{,}103};
\node[inner sep=0pt,anchor=north west,font=\scriptsize] at (417.60,-94.00) {\textbf{\texttt{1s\_wsc}}\,{\tiny(10)}};
\node[inner sep=0pt,anchor=north east,font=\scriptsize] at (514.00,-94.00) {$-82\%$};
\draw[black!25,line width=0.3pt] (417.60,-103.00) -- (514.00,-103.00);
\node[inner sep=0pt,anchor=south west] at (419.56,-173.00) {\includegraphics[width=57.75pt]{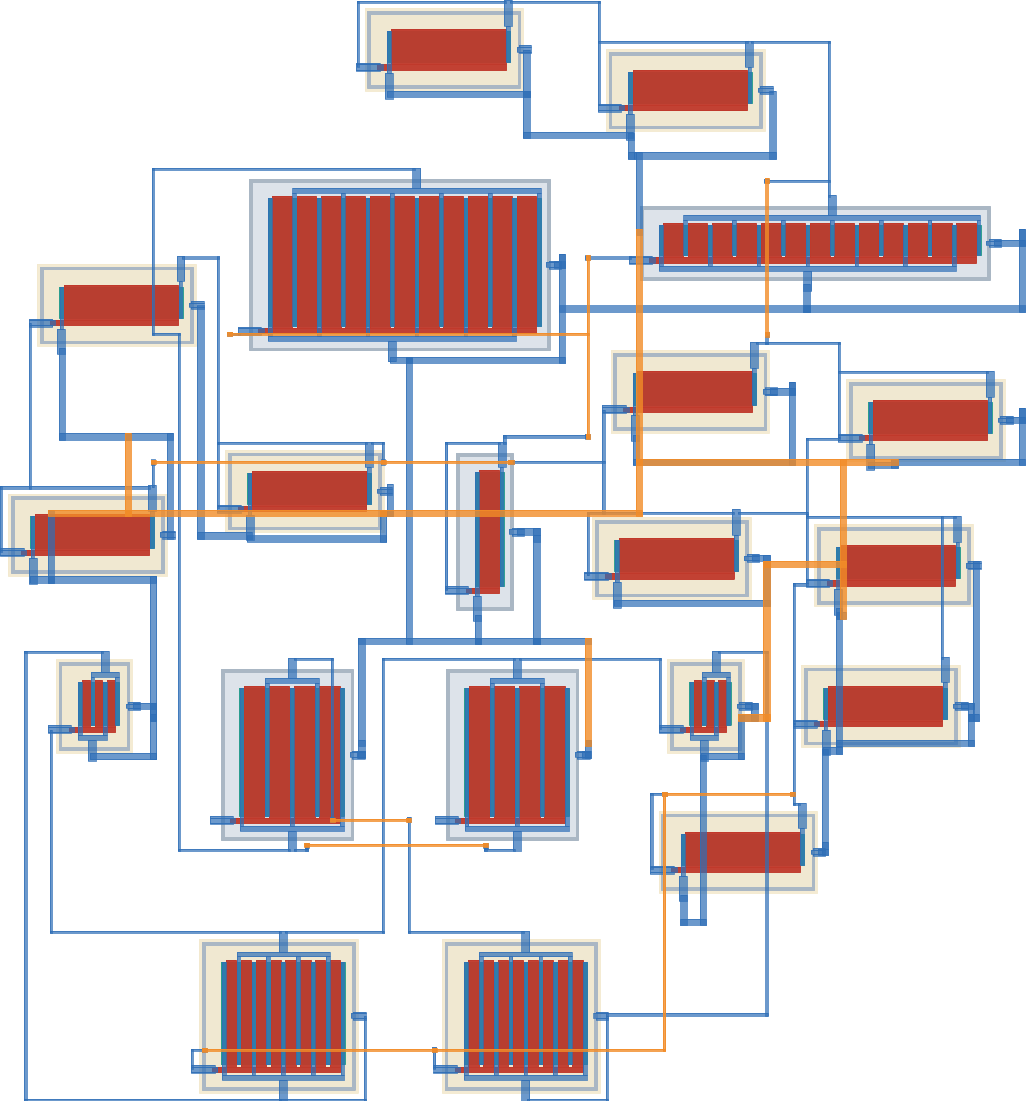}};
\node[lab] at (448.43,-109.80) {APPlace};
\node[ar] at (448.43,-174.80) {4{,}425};
\node[inner sep=0pt,anchor=south west] at (483.30,-173.00) {\includegraphics[width=28.74pt]{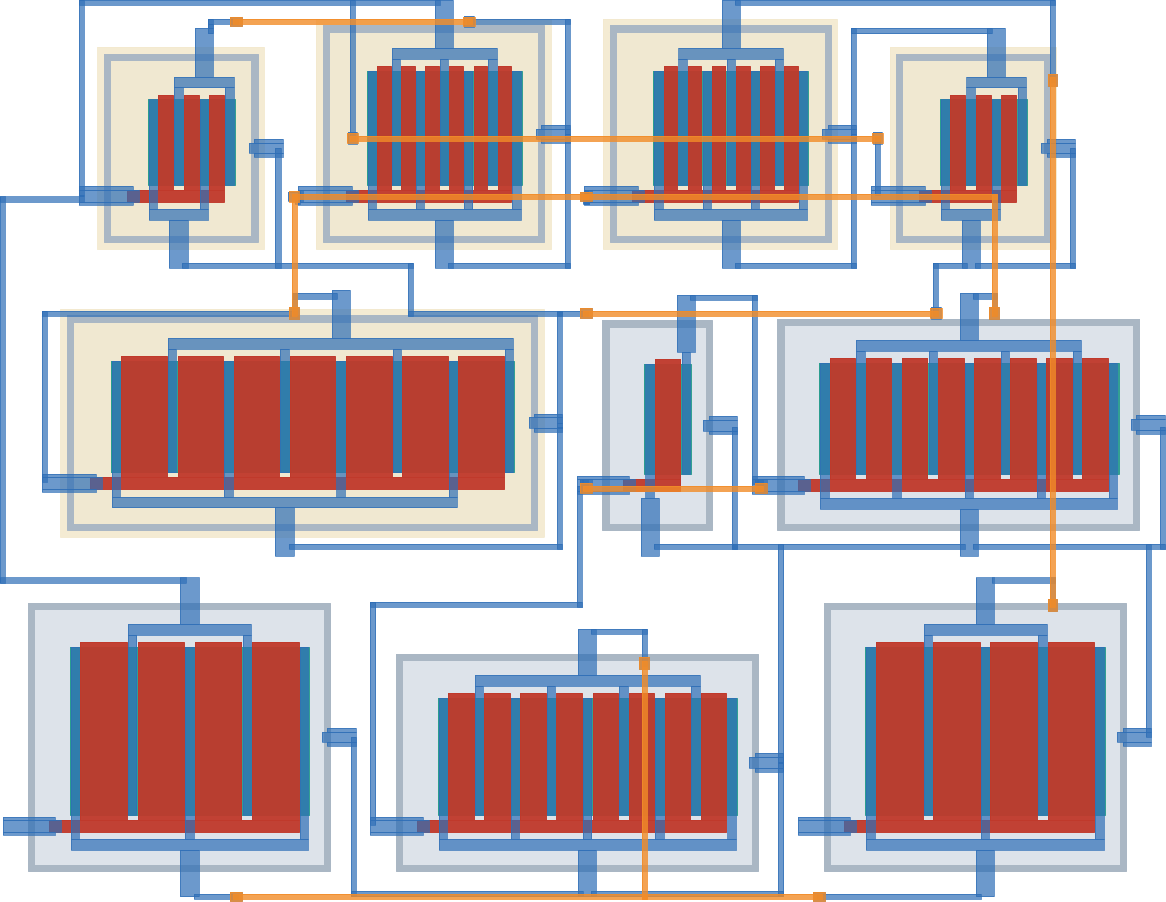}};
\node[lab] at (497.67,-149.57) {CASCoDe};
\node[ar] at (497.67,-174.80) {790};
\fill[fill={rgb,255:red,221;green,227;blue,234},draw=black!40,line width=0.3pt] (0.00,-188.50) rectangle (8.00,-183.50);
\node[inner sep=0pt,anchor=west] at (11.00,-186.00) {PCell outline};
\fill[fill={rgb,255:red,170;green,183;blue,196},draw=black!40,line width=0.3pt] (46.73,-188.50) rectangle (54.73,-183.50);
\node[inner sep=0pt,anchor=west] at (57.73,-186.00) {tap / guard ring};
\fill[fill={rgb,255:red,243;green,233;blue,207},draw=black!40,line width=0.3pt] (93.45,-188.50) rectangle (101.45,-183.50);
\node[inner sep=0pt,anchor=west] at (104.45,-186.00) {n-well};
\fill[fill={rgb,255:red,42;green,157;blue,143},draw=black!40,line width=0.3pt] (140.18,-188.50) rectangle (148.18,-183.50);
\node[inner sep=0pt,anchor=west] at (151.18,-186.00) {diffusion};
\fill[fill={rgb,255:red,192;green,57;blue,43},draw=black!40,line width=0.3pt] (186.91,-188.50) rectangle (194.91,-183.50);
\node[inner sep=0pt,anchor=west] at (197.91,-186.00) {poly};
\fill[fill={rgb,255:red,47;green,111;blue,183},draw=black!40,line width=0.3pt] (233.64,-188.50) rectangle (241.64,-183.50);
\node[inner sep=0pt,anchor=west] at (244.64,-186.00) {met1};
\fill[fill={rgb,255:red,242;green,140;blue,40},draw=black!40,line width=0.3pt] (280.36,-188.50) rectangle (288.36,-183.50);
\node[inner sep=0pt,anchor=west] at (291.36,-186.00) {met2};
\fill[fill={rgb,255:red,123;green,79;blue,179},draw=black!40,line width=0.3pt] (327.09,-188.50) rectangle (335.09,-183.50);
\node[inner sep=0pt,anchor=west] at (338.09,-186.00) {met3};
\fill[fill={rgb,255:red,209;green,73;blue,143},draw=black!40,line width=0.3pt] (373.82,-188.50) rectangle (381.82,-183.50);
\node[inner sep=0pt,anchor=west] at (384.82,-186.00) {met4};
\fill[fill={rgb,255:red,107;green,142;blue,35},draw=black!40,line width=0.3pt] (420.55,-188.50) rectangle (428.55,-183.50);
\node[inner sep=0pt,anchor=west] at (431.55,-186.00) {met5};
\fill[fill={rgb,255:red,233;green,196;blue,106},draw=black!40,line width=0.3pt] (467.27,-188.50) rectangle (475.27,-183.50);
\node[inner sep=0pt,anchor=west] at (478.27,-186.00) {MIM};
\end{tikzpicture}
\caption{Routed layouts of APPlace (left of each pair) and CASCoDe (right) on identical PCells: one-stage amplifiers. Header: circuit, number of netlist devices, and area reduction; below each layout: routed bounding-box area in \areaunit. The two layouts of a circuit share one scale; every pair is scaled to about the same height. The CASCoDe layout shown is one delivered run; Table~\ref{tab:layout_comparison} reports three-run means. li1, contacts, pins, labels and well taps are not drawn.}
\label{fig:layout_compare_1}
\end{figure*}

\begin{figure*}[t]
\centering
\begin{tikzpicture}[x=1pt,y=1pt,font=\tiny,lab/.style={inner sep=0pt,anchor=south,text=black!70},ar/.style={inner sep=0pt,anchor=north}]
\node[inner sep=0pt,anchor=north west,font=\scriptsize] at (0.00,0.00) {\textbf{\texttt{2s\_se}}\,{\tiny(9)}};
\node[inner sep=0pt,anchor=north east,font=\scriptsize] at (96.40,0.00) {$-72\%$};
\draw[black!25,line width=0.3pt] (0.00,-9.00) -- (96.40,-9.00);
\node[inner sep=0pt,anchor=south west] at (7.00,-79.00) {\includegraphics[width=58.90pt]{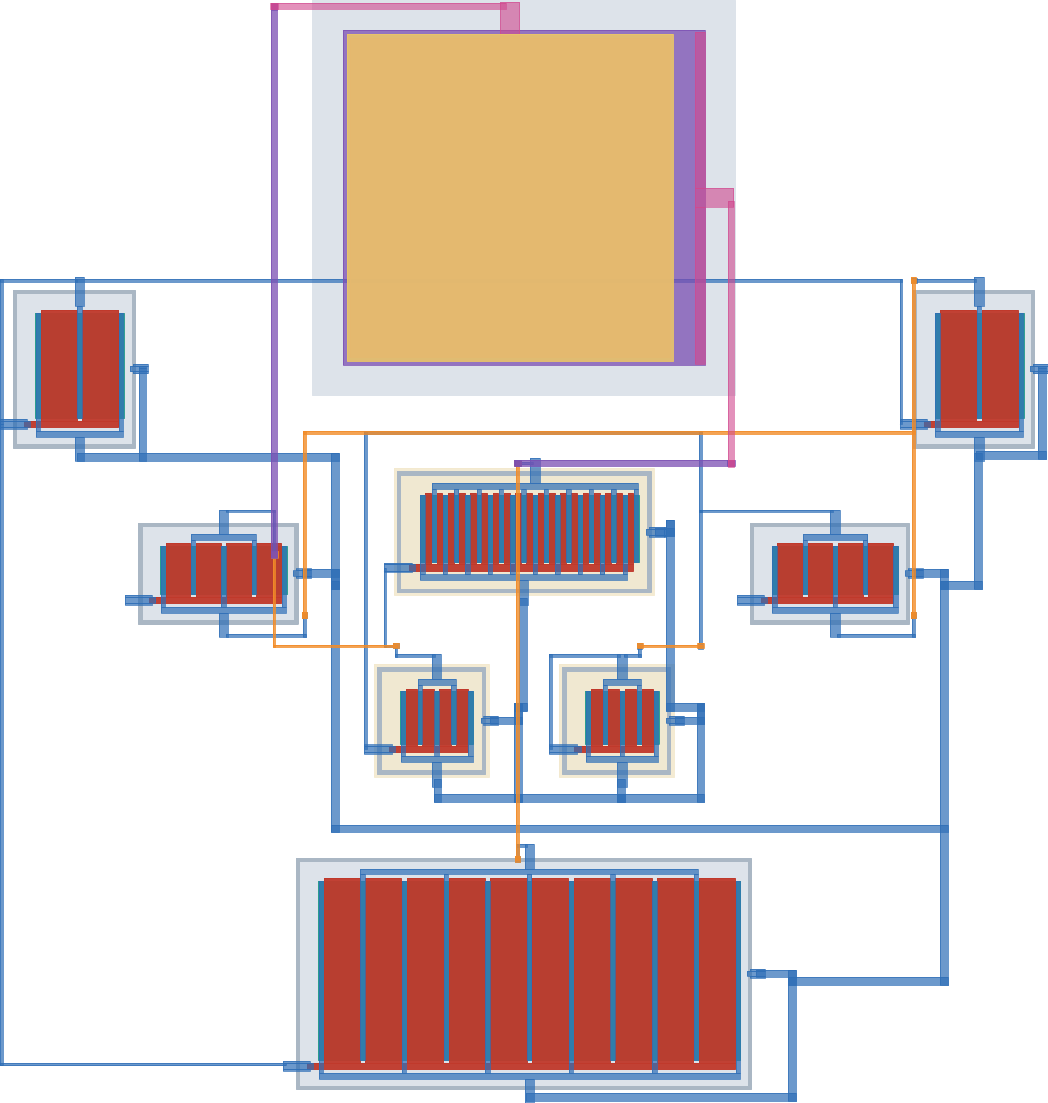}};
\node[lab] at (36.45,-15.80) {APPlace};
\node[ar] at (36.45,-80.80) {3{,}183};
\node[inner sep=0pt,anchor=south west] at (71.90,-79.00) {\includegraphics[width=17.50pt]{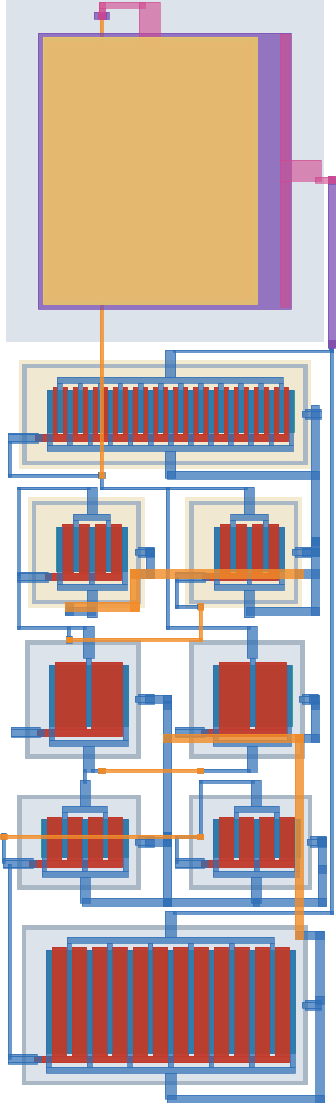}};
\node[lab] at (80.65,-20.26) {CASCoDe};
\node[ar] at (80.65,-80.80) {877};
\node[inner sep=0pt,anchor=north west,font=\scriptsize] at (104.40,0.00) {\textbf{\texttt{2s\_miller}}\,{\tiny(17)}};
\node[inner sep=0pt,anchor=north east,font=\scriptsize] at (200.80,0.00) {$-70\%$};
\draw[black!25,line width=0.3pt] (104.40,-9.00) -- (200.80,-9.00);
\node[inner sep=0pt,anchor=south west] at (104.40,-79.00) {\includegraphics[width=58.74pt]{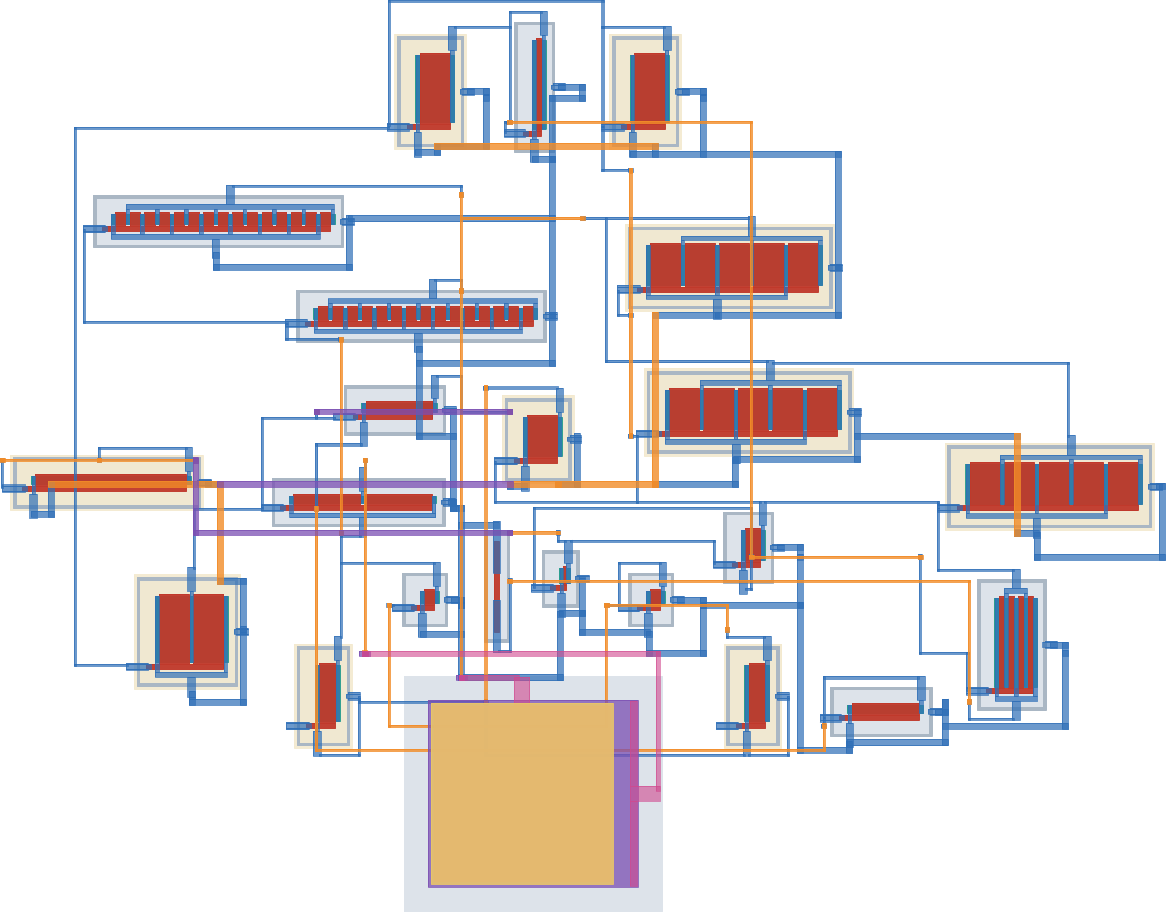}};
\node[lab] at (133.77,-31.84) {APPlace};
\node[ar] at (133.77,-80.80) {4{,}535};
\node[inner sep=0pt,anchor=south west] at (169.14,-79.00) {\includegraphics[width=31.66pt]{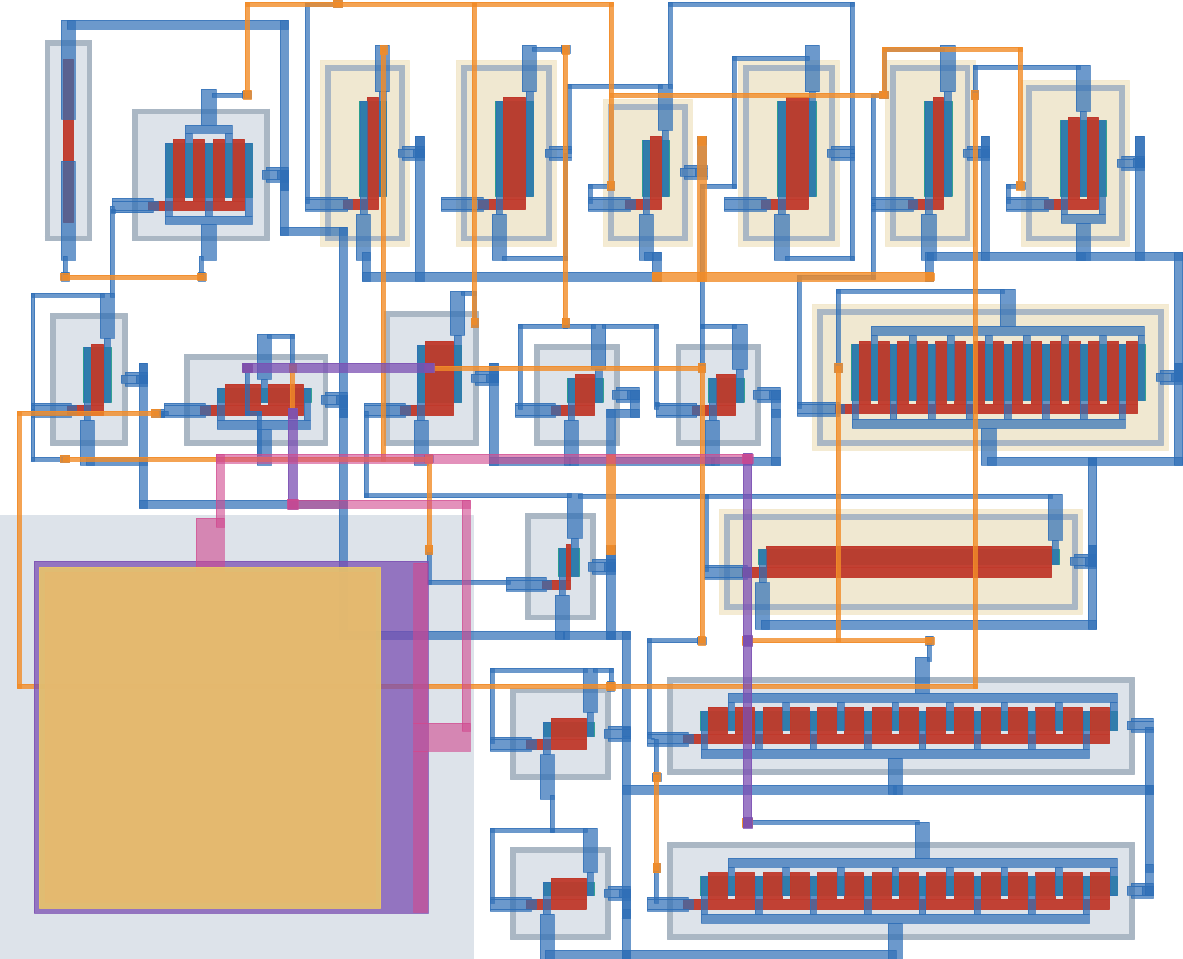}};
\node[lab] at (184.97,-52.12) {CASCoDe};
\node[ar] at (184.97,-80.80) {1{,}382};
\node[inner sep=0pt,anchor=north west,font=\scriptsize] at (208.80,0.00) {\textbf{\texttt{2s\_nfc}}\,{\tiny(19)}};
\node[inner sep=0pt,anchor=north east,font=\scriptsize] at (305.20,0.00) {$-63\%$};
\draw[black!25,line width=0.3pt] (208.80,-9.00) -- (305.20,-9.00);
\node[inner sep=0pt,anchor=south west] at (208.80,-79.00) {\includegraphics[width=54.70pt]{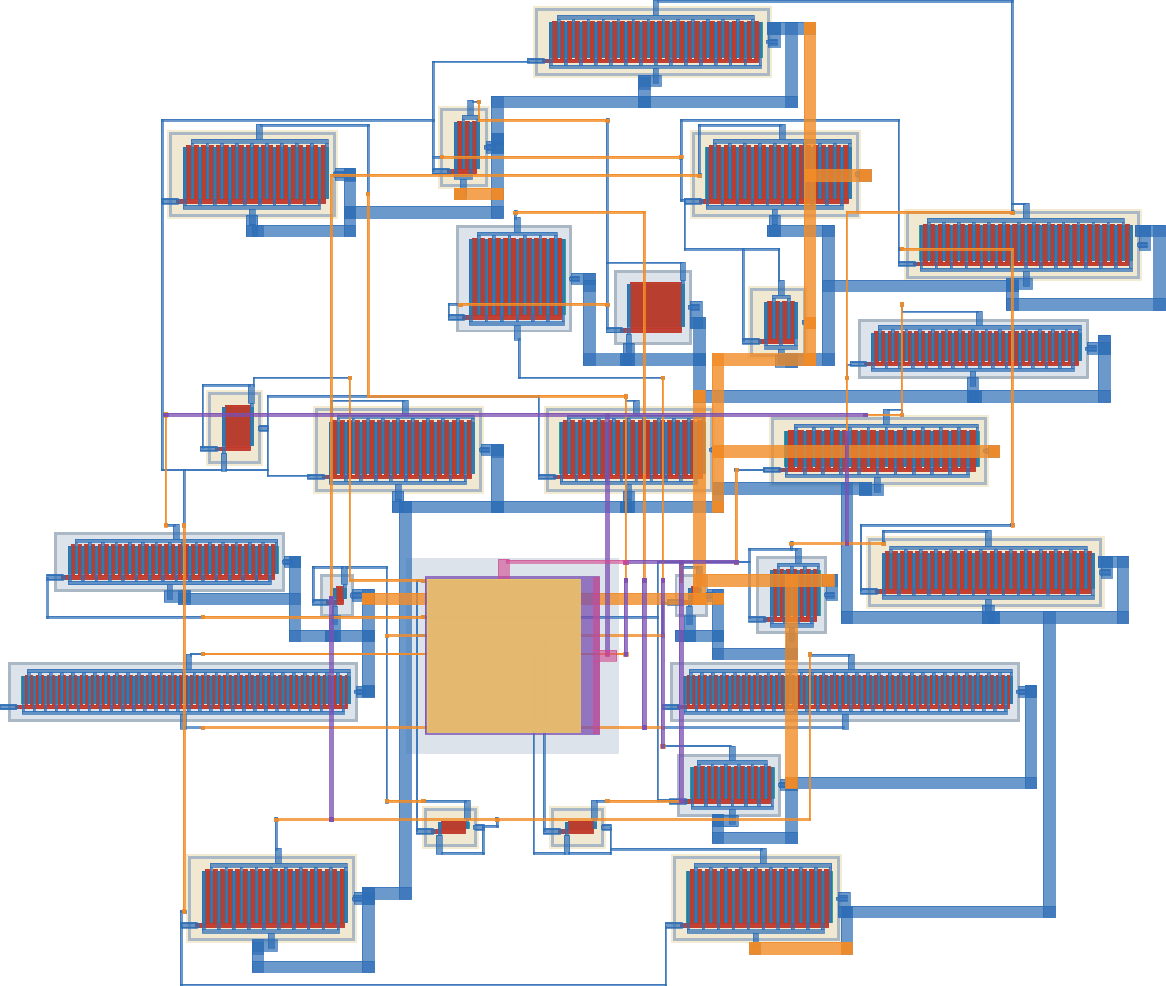}};
\node[lab] at (236.15,-31.51) {APPlace};
\node[ar] at (236.15,-80.80) {8{,}682};
\node[inner sep=0pt,anchor=south west] at (269.50,-79.00) {\includegraphics[width=35.70pt]{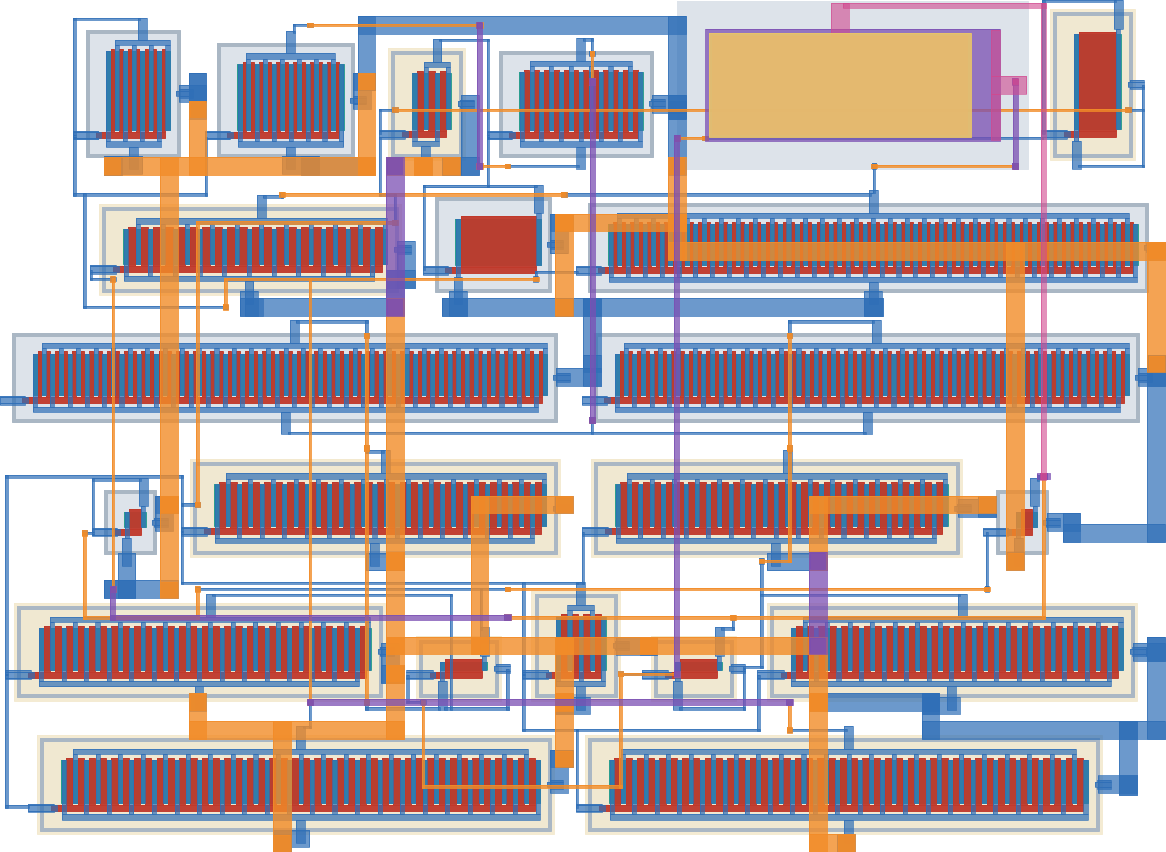}};
\node[lab] at (287.35,-51.70) {CASCoDe};
\node[ar] at (287.35,-80.80) {3{,}195};
\node[inner sep=0pt,anchor=north west,font=\scriptsize] at (313.20,0.00) {\textbf{\texttt{2s\_pcas}}\,{\tiny(15)}};
\node[inner sep=0pt,anchor=north east,font=\scriptsize] at (409.60,0.00) {$-71\%$};
\draw[black!25,line width=0.3pt] (313.20,-9.00) -- (409.60,-9.00);
\node[inner sep=0pt,anchor=south west] at (313.20,-79.00) {\includegraphics[width=54.67pt]{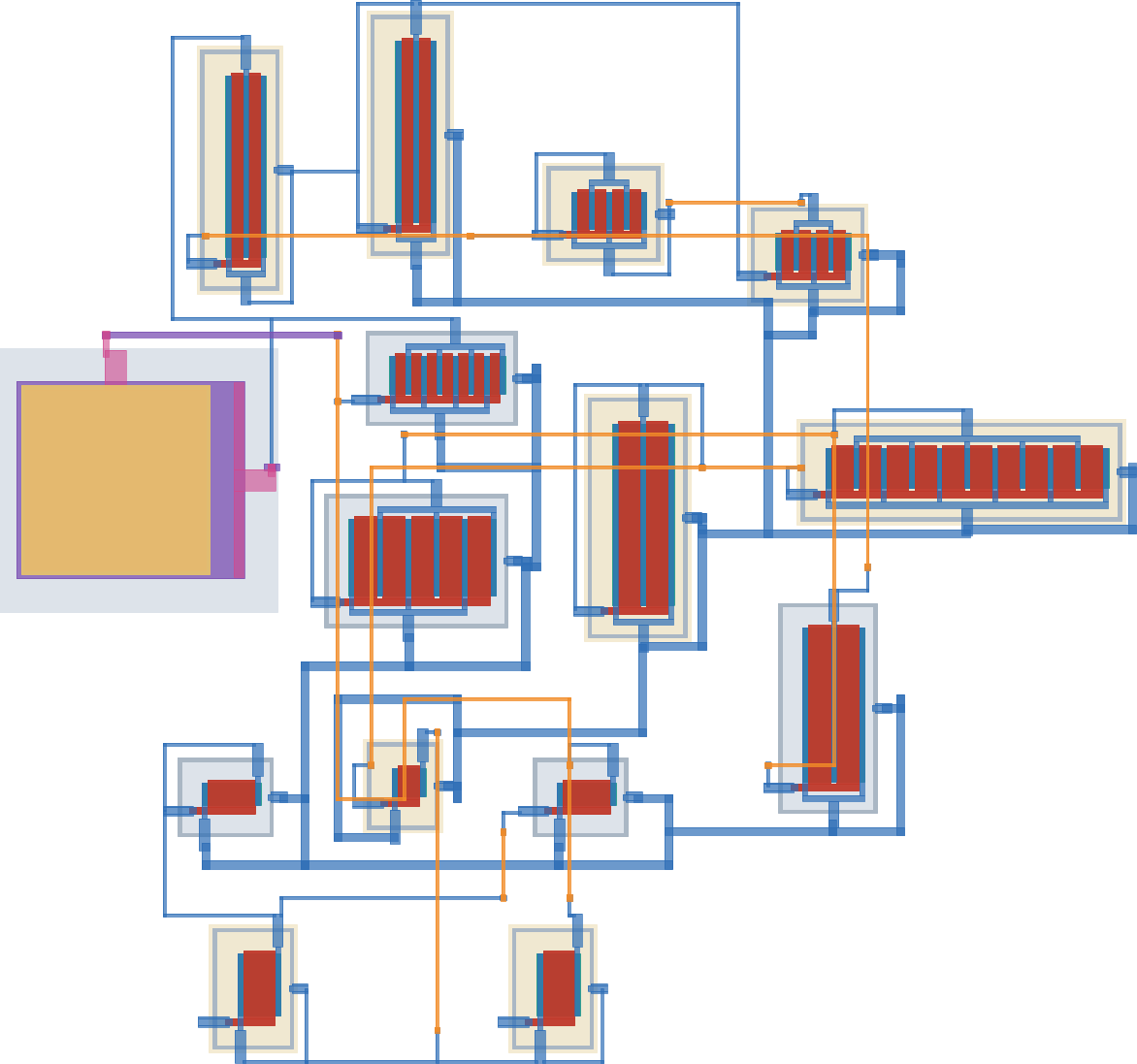}};
\node[lab] at (340.53,-26.64) {APPlace};
\node[ar] at (340.53,-80.80) {2{,}780};
\node[inner sep=0pt,anchor=south west] at (373.87,-79.00) {\includegraphics[width=35.73pt]{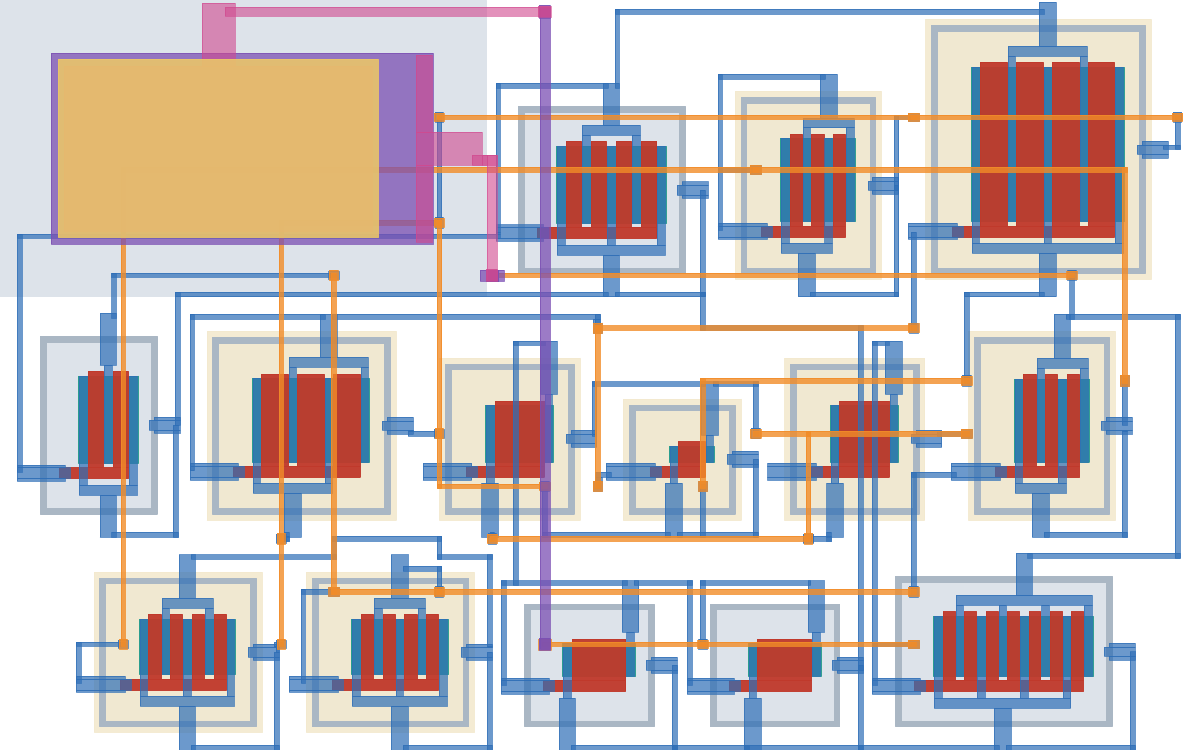}};
\node[lab] at (391.73,-55.15) {CASCoDe};
\node[ar] at (391.73,-80.80) {802};
\node[inner sep=0pt,anchor=north west,font=\scriptsize] at (417.60,0.00) {\textbf{\texttt{2s\_wsc}}\,{\tiny(14)}};
\node[inner sep=0pt,anchor=north east,font=\scriptsize] at (514.00,0.00) {$-81\%$};
\draw[black!25,line width=0.3pt] (417.60,-9.00) -- (514.00,-9.00);
\node[inner sep=0pt,anchor=south west] at (417.60,-79.00) {\includegraphics[width=62.74pt]{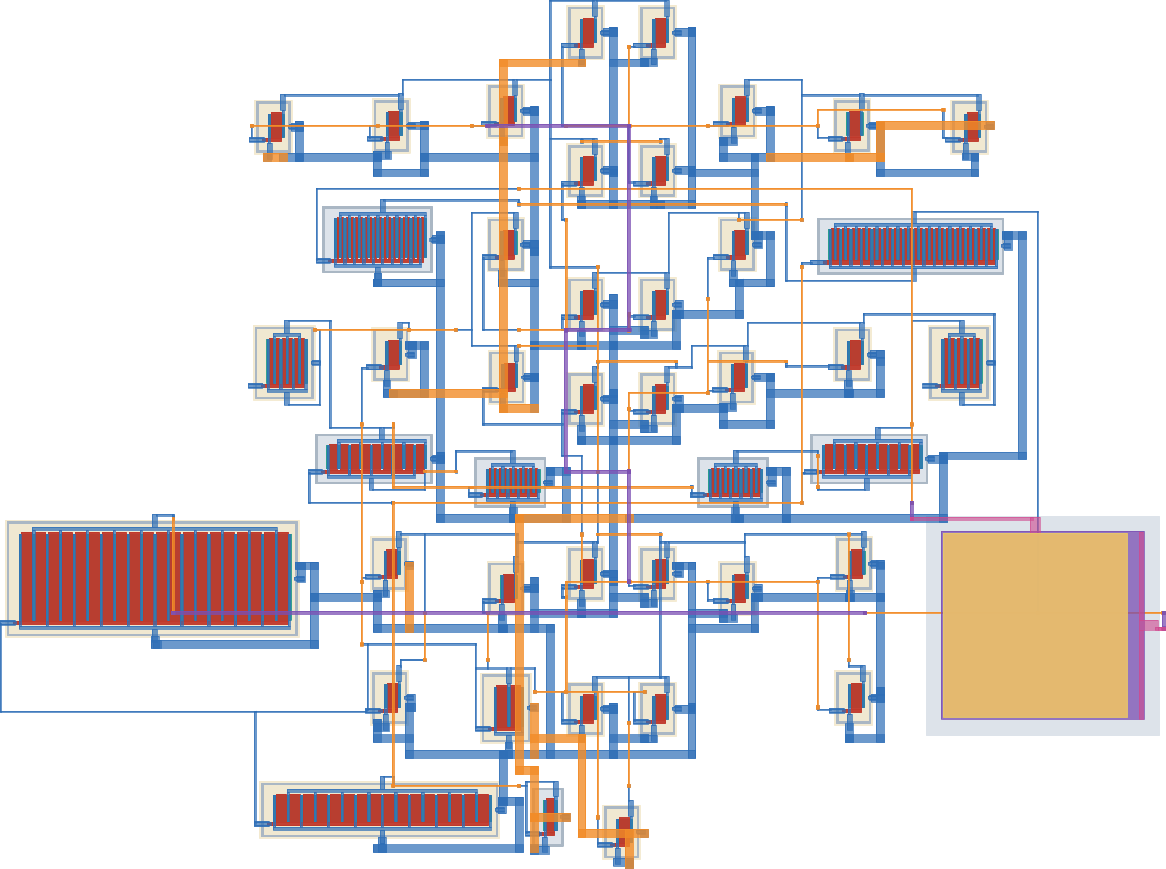}};
\node[lab] at (448.97,-31.05) {APPlace};
\node[ar] at (448.97,-80.80) {10{,}479};
\node[inner sep=0pt,anchor=south west] at (486.34,-79.00) {\includegraphics[width=27.66pt]{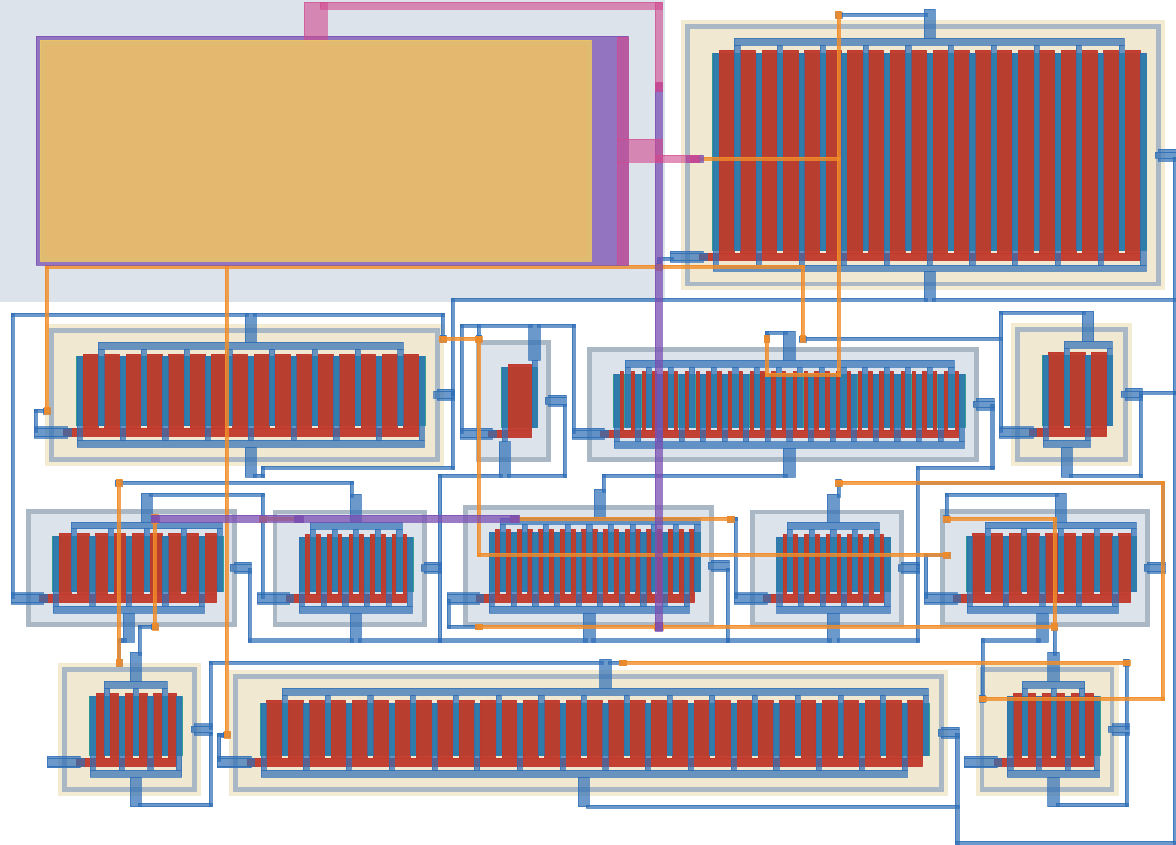}};
\node[lab] at (500.17,-57.94) {CASCoDe};
\node[ar] at (500.17,-80.80) {1{,}939};
\node[inner sep=0pt,anchor=north west,font=\scriptsize] at (0.00,-94.00) {\textbf{\texttt{3s\_acbc}}\,{\tiny(29)}};
\node[inner sep=0pt,anchor=north east,font=\scriptsize] at (96.40,-94.00) {$-59\%$};
\draw[black!25,line width=0.3pt] (0.00,-103.00) -- (96.40,-103.00);
\node[inner sep=0pt,anchor=south west] at (6.15,-173.00) {\includegraphics[width=44.39pt]{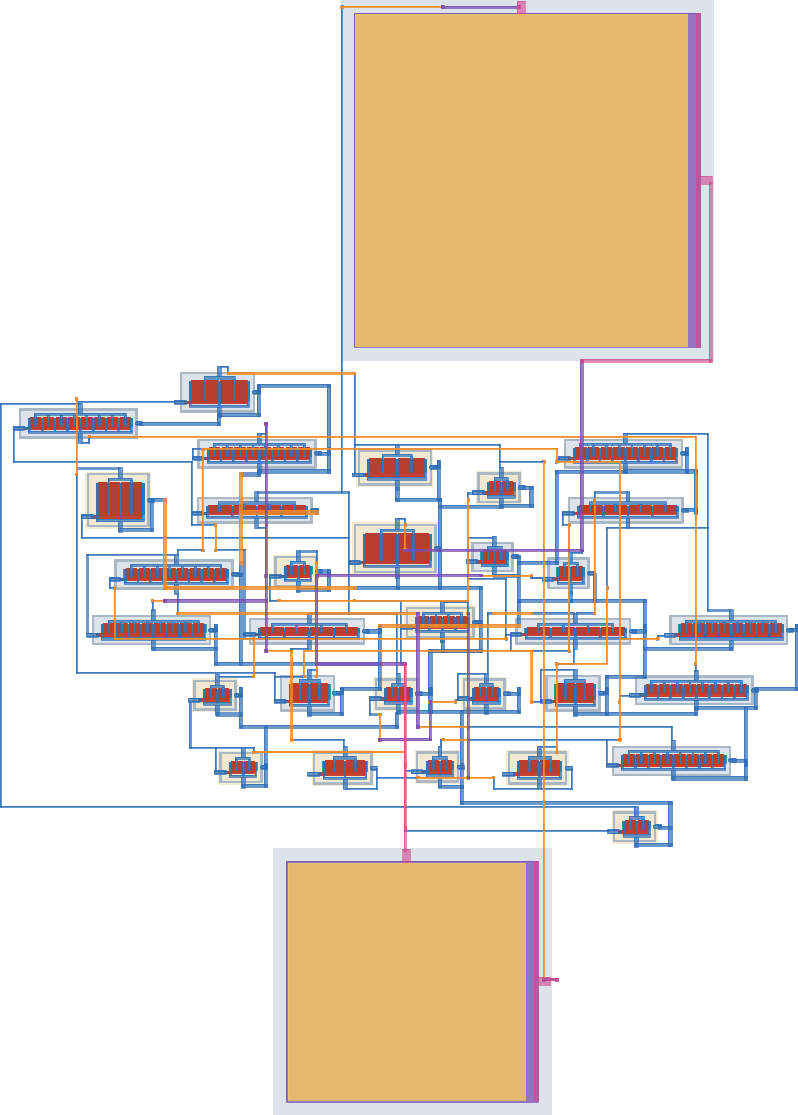}};
\node[lab] at (28.35,-109.80) {APPlace};
\node[ar] at (28.35,-174.80) {14{,}095};
\node[inner sep=0pt,anchor=south west] at (56.54,-173.00) {\includegraphics[width=33.71pt]{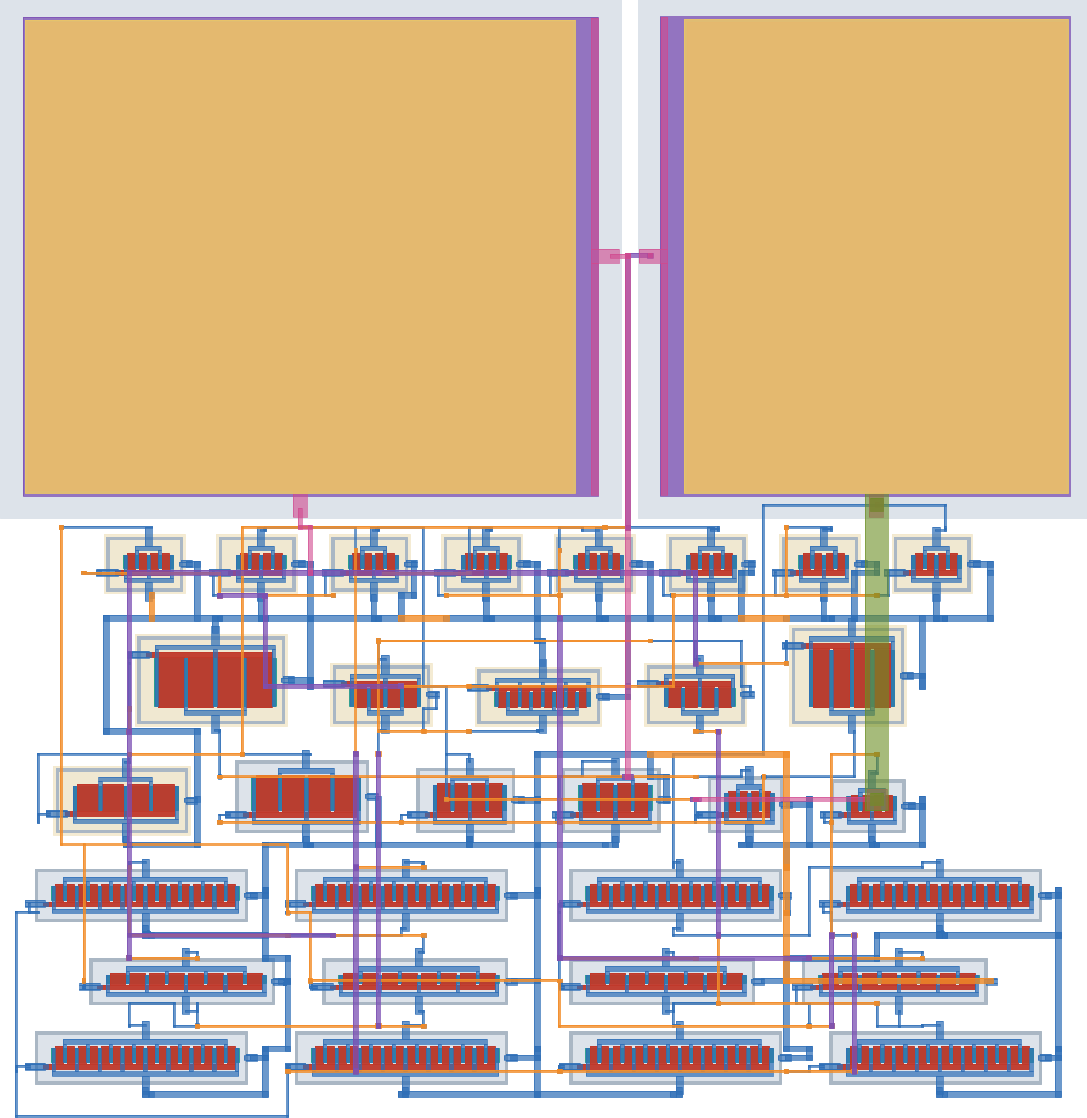}};
\node[lab] at (73.40,-137.15) {CASCoDe};
\node[ar] at (73.40,-174.80) {5{,}795};
\node[inner sep=0pt,anchor=north west,font=\scriptsize] at (104.40,-94.00) {\textbf{\texttt{3s\_affc}}\,{\tiny(32)}};
\node[inner sep=0pt,anchor=north east,font=\scriptsize] at (200.80,-94.00) {$-67\%$};
\draw[black!25,line width=0.3pt] (104.40,-103.00) -- (200.80,-103.00);
\node[inner sep=0pt,anchor=south west] at (104.40,-173.00) {\includegraphics[width=59.12pt]{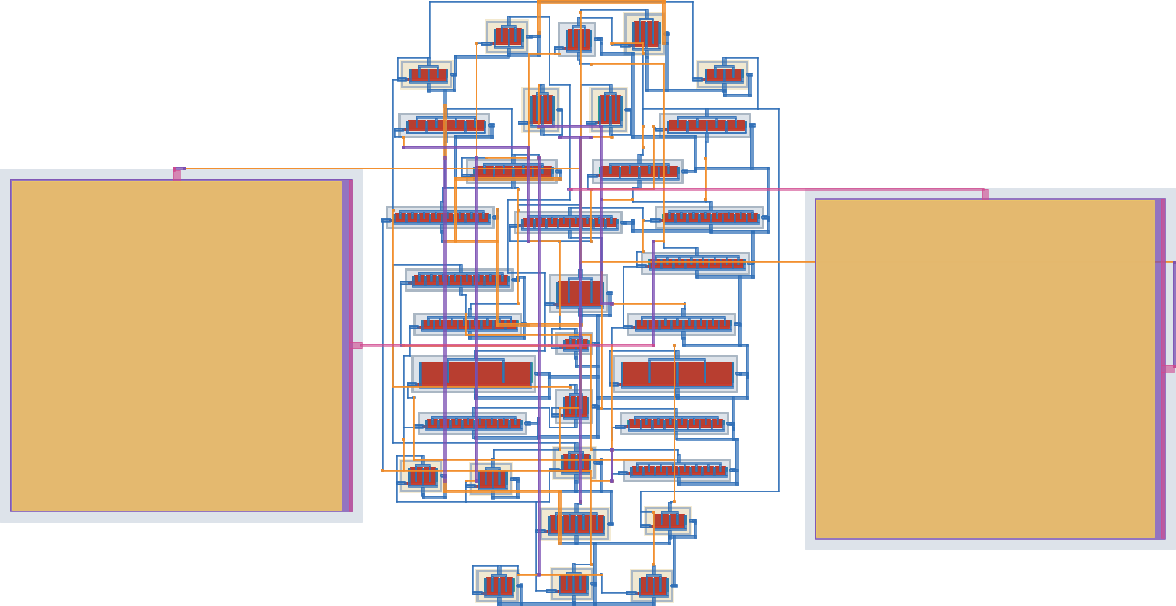}};
\node[lab] at (133.96,-141.34) {APPlace};
\node[ar] at (133.96,-174.80) {16{,}636};
\node[inner sep=0pt,anchor=south west] at (169.52,-173.00) {\includegraphics[width=31.28pt]{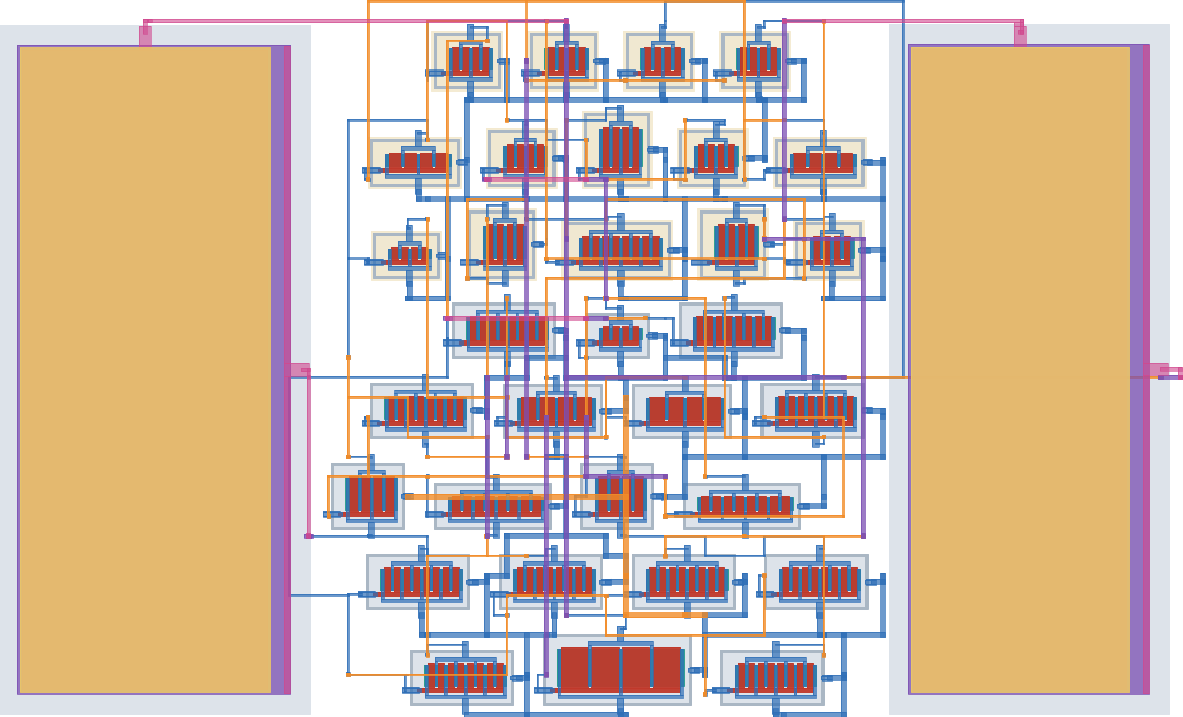}};
\node[lab] at (185.16,-152.84) {CASCoDe};
\node[ar] at (185.16,-174.80) {5{,}445};
\node[inner sep=0pt,anchor=north west,font=\scriptsize] at (208.80,-94.00) {\textbf{\texttt{3s\_dacfc}}\,{\tiny(39)}};
\node[inner sep=0pt,anchor=north east,font=\scriptsize] at (305.20,-94.00) {$-78\%$};
\draw[black!25,line width=0.3pt] (208.80,-103.00) -- (305.20,-103.00);
\node[inner sep=0pt,anchor=south west] at (210.76,-173.00) {\includegraphics[width=62.56pt]{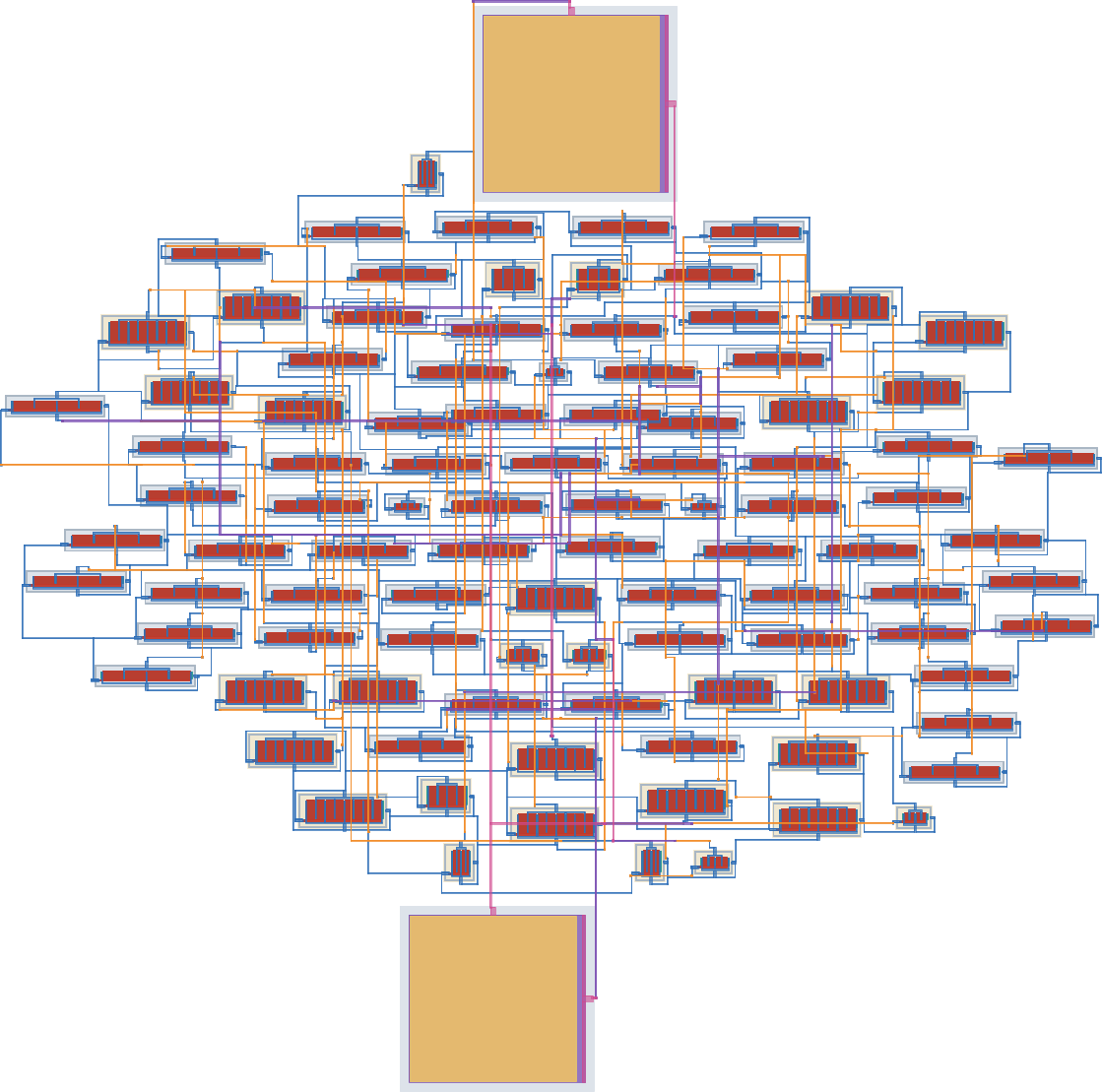}};
\node[lab] at (242.04,-109.80) {APPlace};
\node[ar] at (242.04,-174.80) {39{,}913};
\node[inner sep=0pt,anchor=south west] at (279.32,-173.00) {\includegraphics[width=23.93pt]{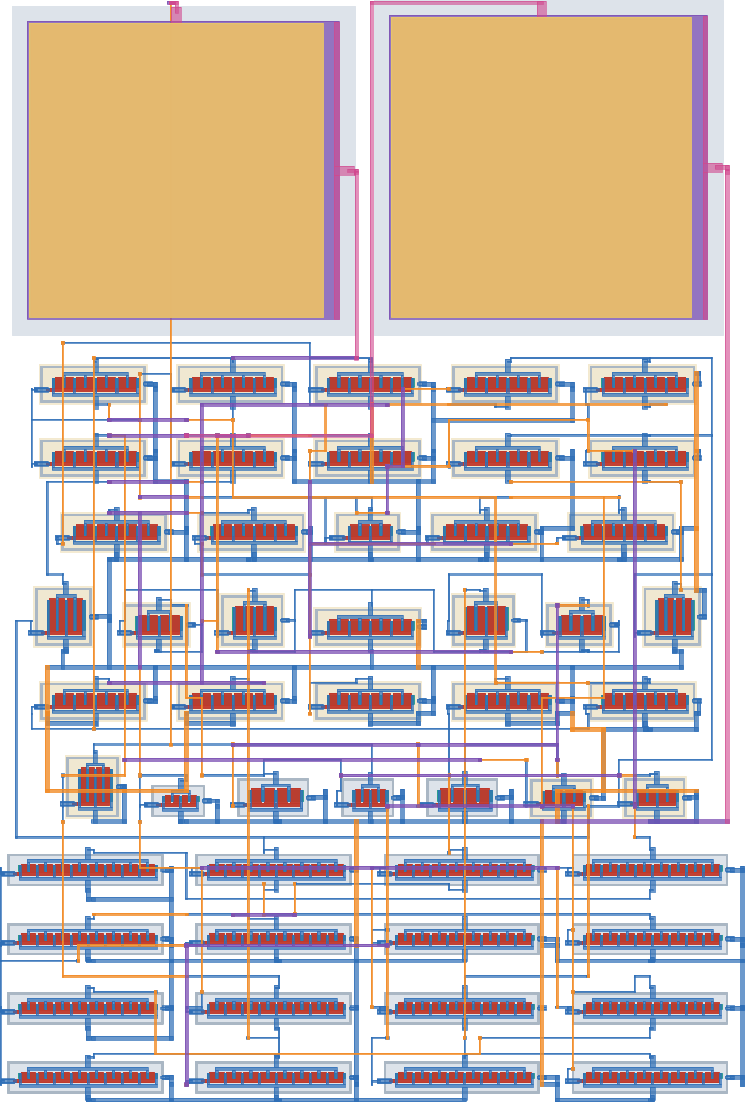}};
\node[lab] at (291.28,-136.39) {CASCoDe};
\node[ar] at (291.28,-174.80) {8{,}785};
\node[inner sep=0pt,anchor=north west,font=\scriptsize] at (313.20,-94.00) {\textbf{\texttt{3s\_smc}}\,{\tiny(25)}};
\node[inner sep=0pt,anchor=north east,font=\scriptsize] at (409.60,-94.00) {$-59\%$};
\draw[black!25,line width=0.3pt] (313.20,-103.00) -- (409.60,-103.00);
\node[inner sep=0pt,anchor=south west] at (313.20,-173.00) {\includegraphics[width=57.36pt]{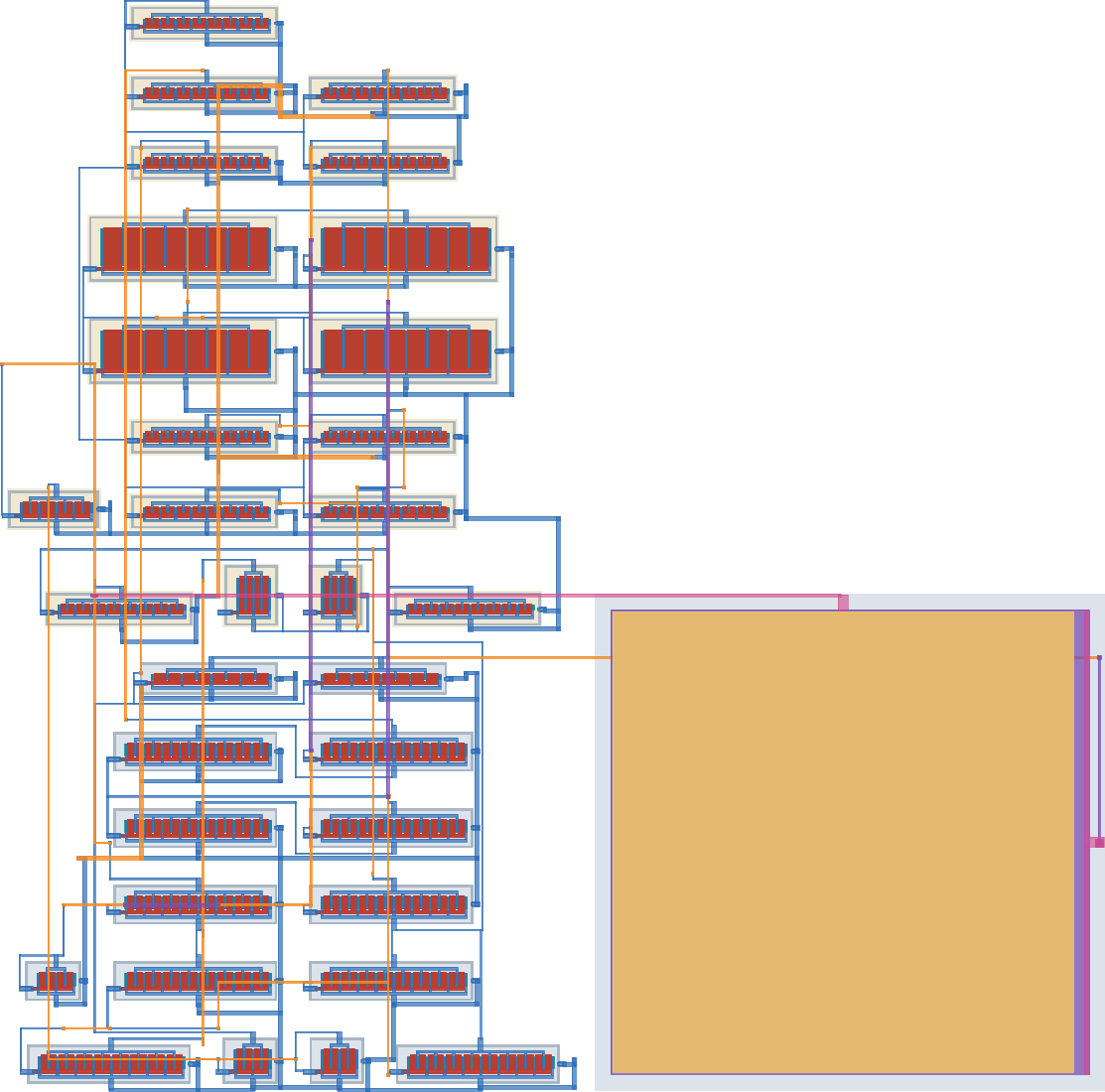}};
\node[lab] at (341.88,-115.11) {APPlace};
\node[ar] at (341.88,-174.80) {12{,}915};
\node[inner sep=0pt,anchor=south west] at (376.56,-173.00) {\includegraphics[width=33.04pt]{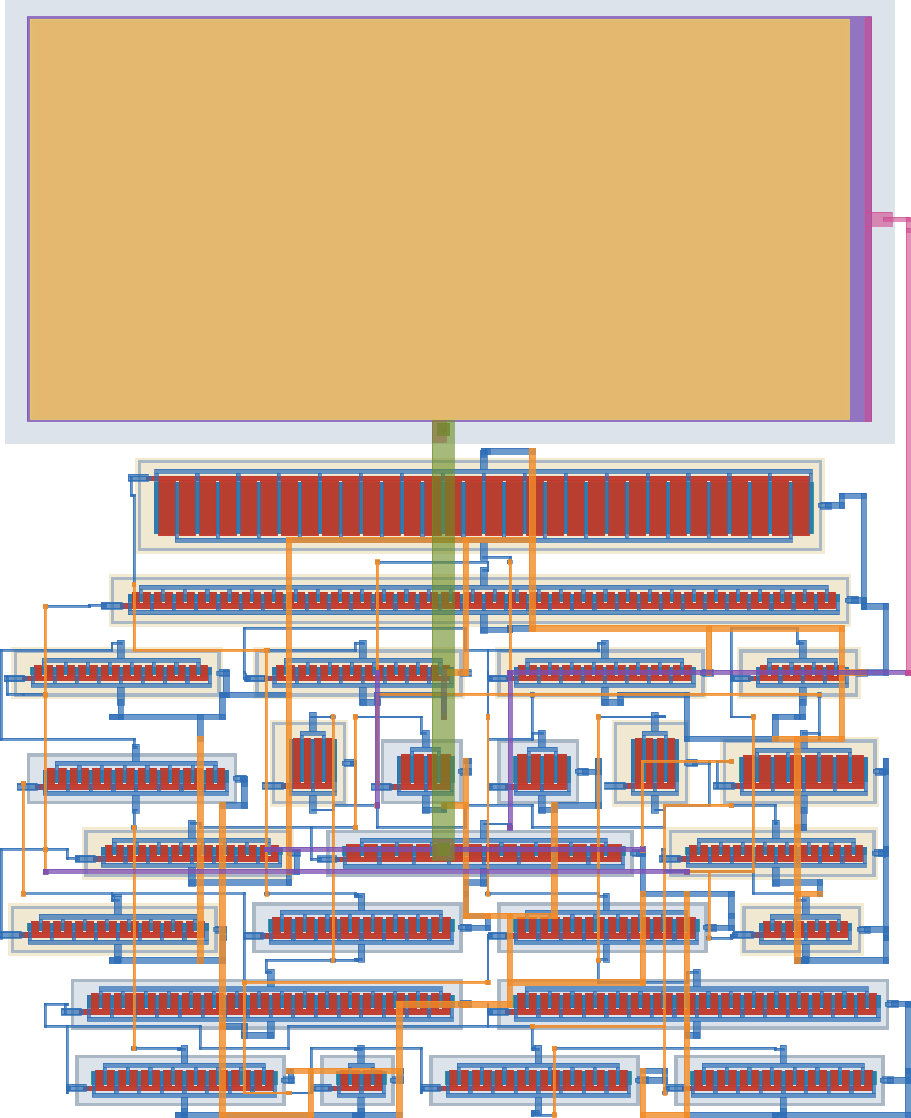}};
\node[lab] at (393.08,-131.25) {CASCoDe};
\node[ar] at (393.08,-174.80) {5{,}246};
\node[inner sep=0pt,anchor=north west,font=\scriptsize] at (417.60,-94.00) {\textbf{\texttt{3s\_tcfc}}\,{\tiny(34)}};
\node[inner sep=0pt,anchor=north east,font=\scriptsize] at (514.00,-94.00) {$-77\%$};
\draw[black!25,line width=0.3pt] (417.60,-103.00) -- (514.00,-103.00);
\node[inner sep=0pt,anchor=south west] at (417.60,-173.00) {\includegraphics[width=62.66pt]{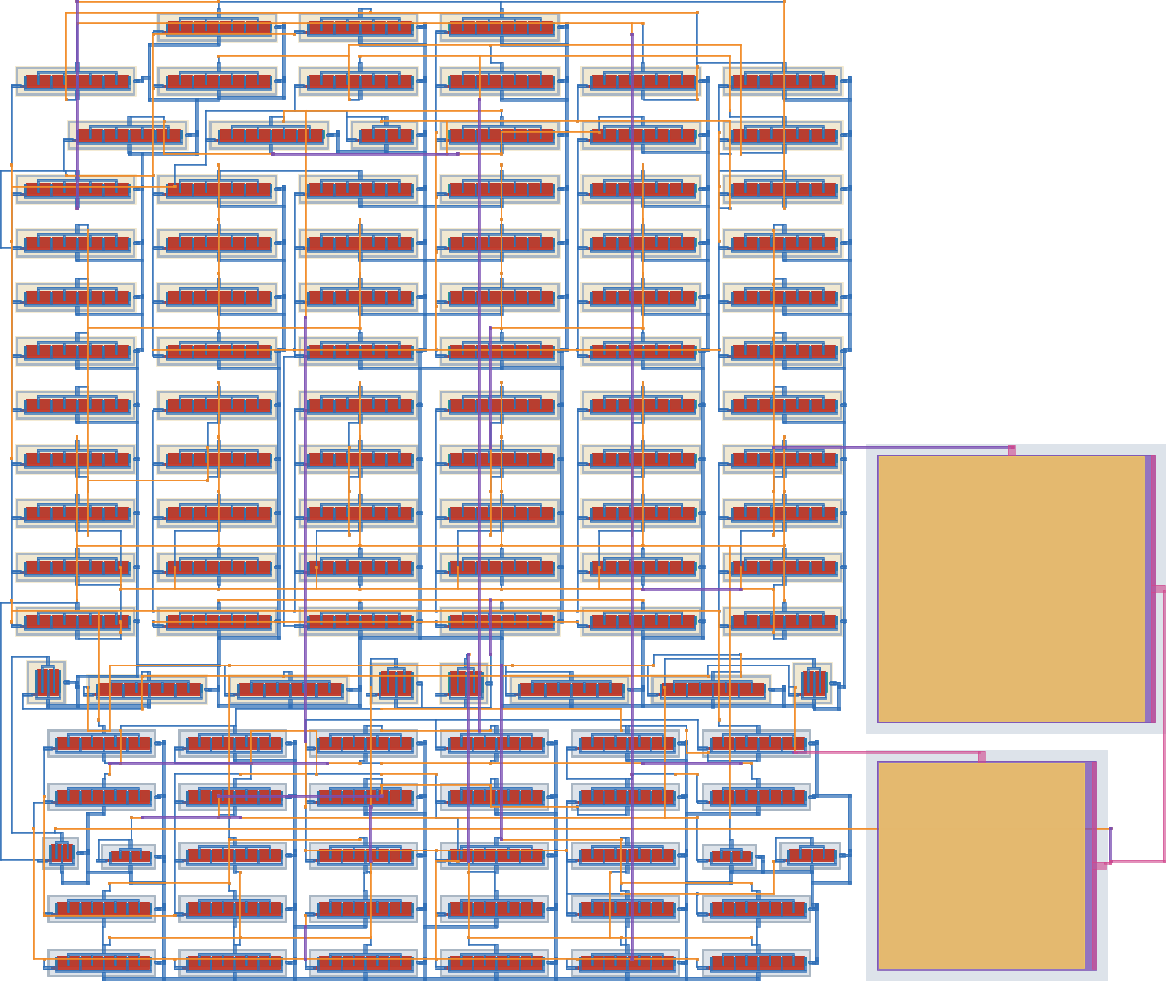}};
\node[lab] at (448.93,-119.09) {APPlace};
\node[ar] at (448.93,-174.80) {24{,}699};
\node[inner sep=0pt,anchor=south west] at (486.26,-173.00) {\includegraphics[width=27.74pt]{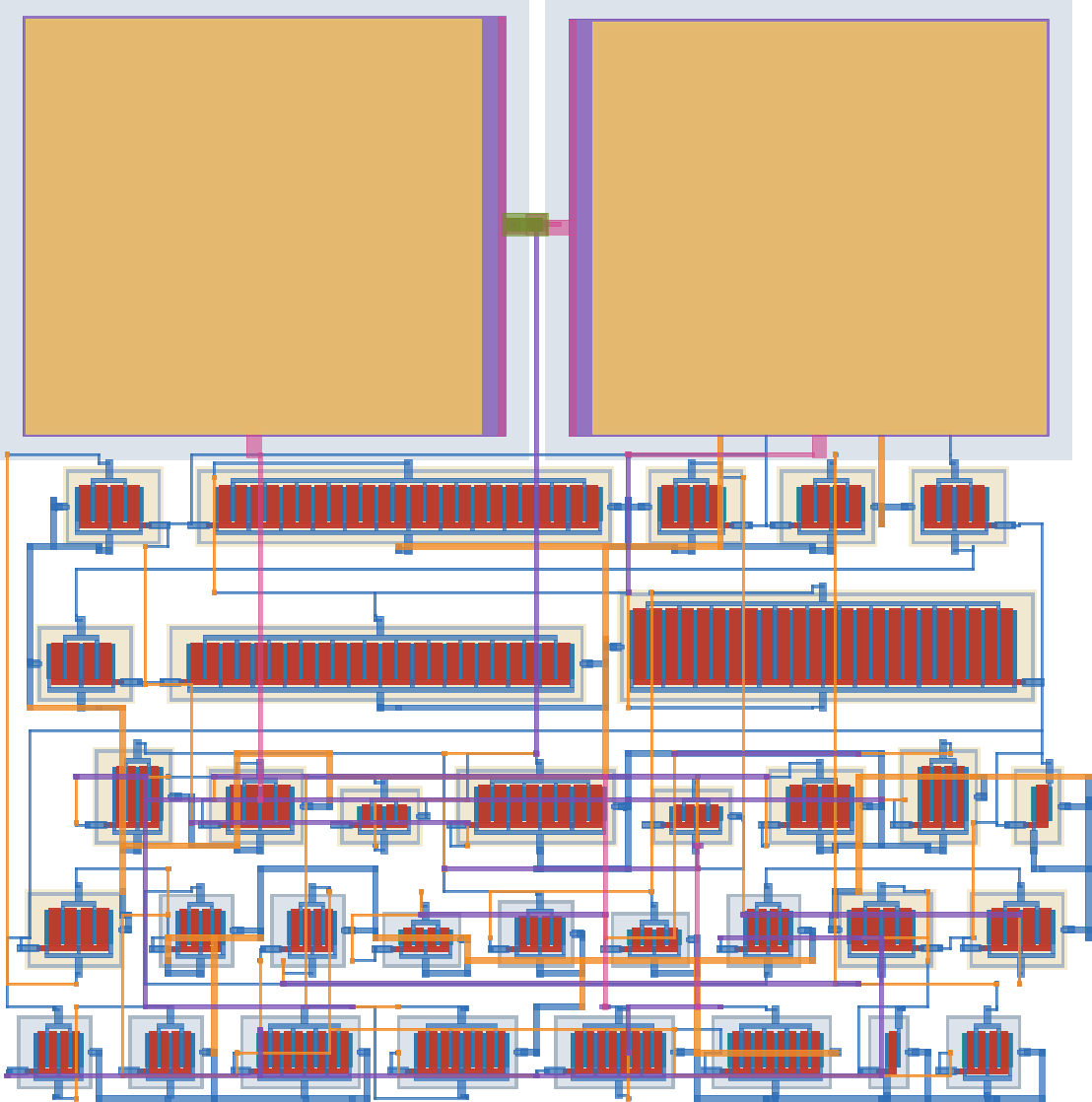}};
\node[lab] at (500.13,-143.80) {CASCoDe};
\node[ar] at (500.13,-174.80) {5{,}703};
\fill[fill={rgb,255:red,221;green,227;blue,234},draw=black!40,line width=0.3pt] (0.00,-188.50) rectangle (8.00,-183.50);
\node[inner sep=0pt,anchor=west] at (11.00,-186.00) {PCell outline};
\fill[fill={rgb,255:red,170;green,183;blue,196},draw=black!40,line width=0.3pt] (46.73,-188.50) rectangle (54.73,-183.50);
\node[inner sep=0pt,anchor=west] at (57.73,-186.00) {tap / guard ring};
\fill[fill={rgb,255:red,243;green,233;blue,207},draw=black!40,line width=0.3pt] (93.45,-188.50) rectangle (101.45,-183.50);
\node[inner sep=0pt,anchor=west] at (104.45,-186.00) {n-well};
\fill[fill={rgb,255:red,42;green,157;blue,143},draw=black!40,line width=0.3pt] (140.18,-188.50) rectangle (148.18,-183.50);
\node[inner sep=0pt,anchor=west] at (151.18,-186.00) {diffusion};
\fill[fill={rgb,255:red,192;green,57;blue,43},draw=black!40,line width=0.3pt] (186.91,-188.50) rectangle (194.91,-183.50);
\node[inner sep=0pt,anchor=west] at (197.91,-186.00) {poly};
\fill[fill={rgb,255:red,47;green,111;blue,183},draw=black!40,line width=0.3pt] (233.64,-188.50) rectangle (241.64,-183.50);
\node[inner sep=0pt,anchor=west] at (244.64,-186.00) {met1};
\fill[fill={rgb,255:red,242;green,140;blue,40},draw=black!40,line width=0.3pt] (280.36,-188.50) rectangle (288.36,-183.50);
\node[inner sep=0pt,anchor=west] at (291.36,-186.00) {met2};
\fill[fill={rgb,255:red,123;green,79;blue,179},draw=black!40,line width=0.3pt] (327.09,-188.50) rectangle (335.09,-183.50);
\node[inner sep=0pt,anchor=west] at (338.09,-186.00) {met3};
\fill[fill={rgb,255:red,209;green,73;blue,143},draw=black!40,line width=0.3pt] (373.82,-188.50) rectangle (381.82,-183.50);
\node[inner sep=0pt,anchor=west] at (384.82,-186.00) {met4};
\fill[fill={rgb,255:red,107;green,142;blue,35},draw=black!40,line width=0.3pt] (420.55,-188.50) rectangle (428.55,-183.50);
\node[inner sep=0pt,anchor=west] at (431.55,-186.00) {met5};
\fill[fill={rgb,255:red,233;green,196;blue,106},draw=black!40,line width=0.3pt] (467.27,-188.50) rectangle (475.27,-183.50);
\node[inner sep=0pt,anchor=west] at (478.27,-186.00) {MIM};
\end{tikzpicture}
\caption{Routed layouts of APPlace (left of each pair) and CASCoDe (right) on identical PCells: two- and three-stage amplifiers. Header: circuit, number of netlist devices, and area reduction; below each layout: routed bounding-box area in \areaunit. The two layouts of a circuit share one scale; every pair is scaled to about the same height. The CASCoDe layout shown is one delivered run; Table~\ref{tab:layout_comparison} reports three-run means. li1, contacts, pins, labels and well taps are not drawn.}
\label{fig:layout_compare_2}
\end{figure*}
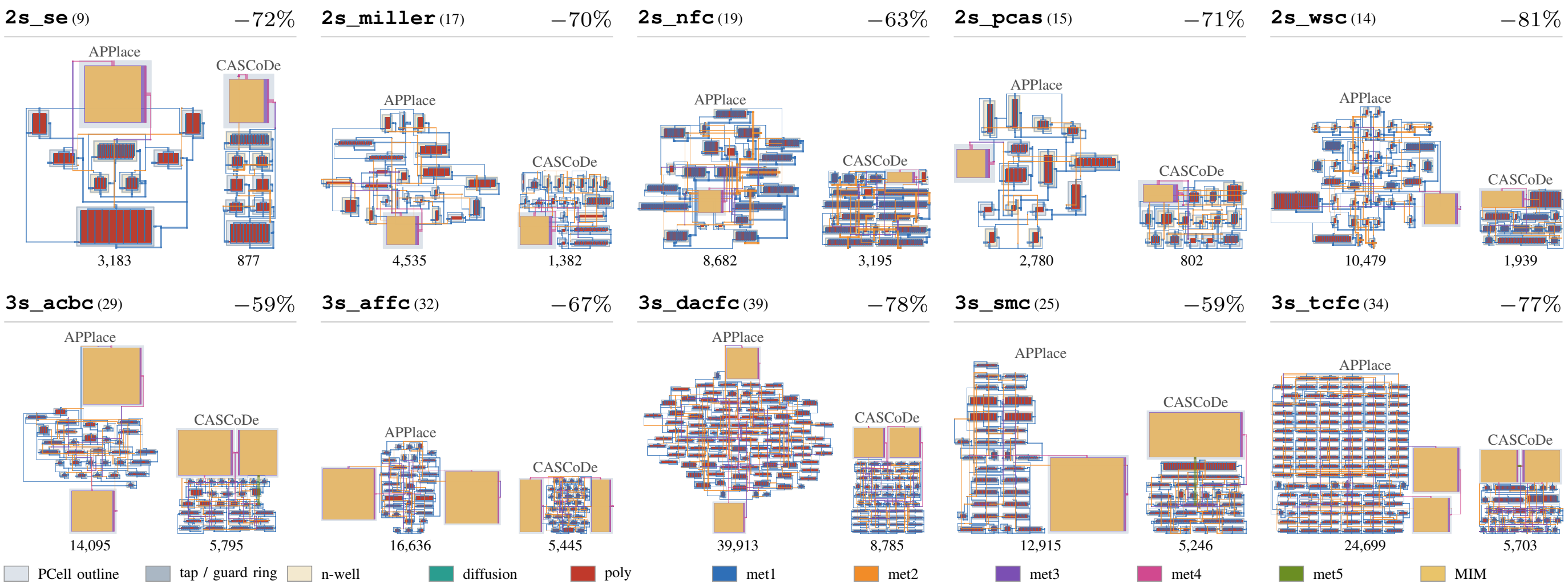

\parhead{Benchmarks}Table~\ref{tab:layout_comparison} lists 20 SKY130
amplifiers: ten one-stage OTAs (\texttt{1s}), five two-stage
(\texttt{2s}) and five three-stage (\texttt{3s}) amplifiers, with 6--39
devices each (Figs.~\ref{fig:layout_compare_1}
and~\ref{fig:layout_compare_2}). The two- and three-stage designs
carry MIM compensation capacitors.

\parhead{Compared methods}APPlace~\cite{applace} is a conventional row-based P\&R
flow on the same PCells as CASCoDe. MAGICAL~\cite{magical} (the
authors' SKY130 branch) and ALIGN~\cite{align} (ALIGN-public with its
SKY130 PDK) run as released after minimal fixes for crashes and PDK
mismatches; their DRC counts omit violations caused by PDK-mismatched
device generation and keep only genuine, unrepairable ones, and
layouts failing LVS are minimally repaired before simulation.

\parhead{Metrics}We report the DRC violation count, LVS result, routed
bounding-box area, and device density, the summed PCell bounding-box
area divided by the routed area.

\parhead{Performance requirement}Because CASCoDe optimizes end to end,
from sizing to the routed layout, every flow is held to the same
pre-layout reference. The reference is one sizing point, the smallest-area initial sample within the top 5\% of every
metric, whose pre-layout simulation gives values $m_k^{\mathrm{ref}}$
for seven metrics. Each post-layout metric $m_k$ is expressed as a ratio
oriented so that $r_k\ge1$ means at least the reference performance:
$r_k=m_k/m_k^{\mathrm{ref}}$ for Gain, GBW, SR, CMRR, and PSRR, and
$r_k=m_k^{\mathrm{ref}}/m_k$ for the supply current $I_{\mathrm{dc}}$
and Noise. A layout must keep every metric at no less than 80\% of the
reference, $r_k\ge0.8$ for all $k$, and meet the phase-margin
requirement $\mathrm{PM}\ge\theta_{\mathrm{PM}}$, with
$\theta_{\mathrm{PM}}=60^{\circ}$ for one-stage and $45^{\circ}$ for
two- and three-stage amplifiers.

Among layouts that meet this requirement, the FoM ranks overall
performance. Each ratio is clipped to $[0,2]$ before summation, and the
phase-margin requirement adds one point:
\begin{align}
\mathrm{FoM}&=\sum_{k=1}^{7}\min\{2,\max\{0,r_k\}\}
+\mathbb 1[\mathrm{PM}\ge\theta_{\mathrm{PM}}],\label{eq:score}\\
F_{\mathrm{post}}&=\mathrm{FoM}\cdot
\mathbb 1\bigl[\mathrm{PM}\ge\theta_{\mathrm{PM}}\text{ and }\min\nolimits_k r_k\ge0.8\bigr].\label{eq:fpost}
\end{align}
The two bounds keep a single anomalous metric from dominating the sum
in either direction, and stop a design from collecting credit by
over-optimizing one metric far beyond the reference. A layout
\emph{passes} when $F_{\mathrm{post}}>0$, i.e., when every metric meets
the 80\% requirement and the phase margin holds; this is the constraint
of Eq.~\eqref{eq:problem}. Because a ratio above 1 can offset a weak
metric in the sum, the FoM value alone does not decide passing: a
layout whose FoM exceeds $1+7\times0.8=6.6$ still fails if any single
metric falls below 0.8. Tables report the FoM and shade failing
layouts.

\subsection{End-to-End Comparison}
Eq.~\eqref{eq:problem} asks for the smallest layout that meets the
performance requirement, so every flow is given that same
problem: each baseline receives every sizing CASCoDe used during its
run plus the 20 best initial samples, runs five times on each, and
reports its smallest-area passing layout, or its highest-FoM layout if
none passes; CASCoDe reports its delivered design, averaged over three
runs, whose routed area varies by 3--17\% (mean 8\%) across runs.
Table~\ref{tab:layout_comparison} lists every layout, and
Fig.~\ref{fig:main_table} plots its area, FoM, and density per circuit.

% Generated by scripts/make_layout_compare.py -- do not edit by hand.

% Generated by scripts/make_layout_compare.py -- do not edit by hand.

CASCoDe is the only
flow that is DRC- and LVS-clean on all 20 circuits: APPlace is clean on
15, MAGICAL on none, and ALIGN on one, and they fail LVS on one, three,
and four circuits. CASCoDe also meets the performance criterion on all
20, against 14, 9, and 11, although each baseline picks its best layout
over all sizings. Under this criterion it is smaller than every
baseline on every circuit, reducing area by 59--85\% (mean 71\%)
against APPlace on identical PCells. Its mean FoM
(6.92) exceeds every baseline (6.76, 6.04, and 6.52) and is the highest
on nine circuits; where a baseline scores higher, the gap is at most
0.30 against APPlace and 0.60 against a DRC-dirty MAGICAL or ALIGN
layout.

\parhead{Per-circuit view}Fig.~\ref{fig:main_table} makes the per-circuit
pattern of Table~\ref{tab:layout_comparison} visible at a glance. In
panel~(a), CASCoDe forms a separate band at 0.15--0.41$\times$ the
APPlace area on every circuit, while MAGICAL lies above the APPlace line
on 14 circuits. ALIGN is the closest baseline in area (0.32--0.98$\times$),
but its markers are hollow on 19 of 20 circuits: its smaller footprint
comes with DRC or LVS violations. Even against the smallest baseline
layout of each circuit, dirty ones included, CASCoDe needs only
0.23--0.79$\times$ the area. Panel~(b) shows that this saving does not
trade away performance: every CASCoDe layout passes (FoM
6.64--7.71), whereas the baselines have 25 failing layouts (ten for
MAGICAL, nine for ALIGN, and six for APPlace), 19 of them in the two-
and three-stage classes. The panel also shows why passing is checked
per metric rather than by FoM: five baseline layouts with FoM between
6.63 and 6.69 fail because a single metric or the phase margin misses
its requirement, while CASCoDe layouts with similar FoM, such as
\texttt{1s\_5t} at 6.64, meet all of them. Panel~(c) explains the area gap: CASCoDe reaches a device
density of 0.61--0.89, above every baseline layout (at most 0.63), and
exceeds the densest baseline of each circuit by 0.11 (\texttt{1s\_sym})
to 0.38 (\texttt{2s\_pcas}).

\subsection{Layout Analysis}
\label{sec:layout}
Figs.~\ref{fig:layout_compare_1} and~\ref{fig:layout_compare_2} draw the
APPlace and CASCoDe layouts of every circuit side by side; the two
layouts of a circuit share one scale, so their footprints compare
directly. The area gap builds up in three steps.

\parhead{(1) Regular arrangement}The first saving comes from
regularity. APPlace keeps the row structure of the conventional P\&R
flow, but its rows are loosely filled and separated by wide routing
channels, and large devices protrude from the device array: the wide
multi-finger devices of \texttt{1s\_iwil} and \texttt{1s\_nwil} and the
large blocks of \texttt{1s\_ntele} and \texttt{1s\_wsc} each extend the
bounding box well beyond the rest of the layout. CASCoDe first chooses
an arrangement, i.e., which devices form rows, stacks, or side columns
(Sec.~\ref{sec:search}), and only then realizes it as aligned rows.
Devices of similar height share a row and rows share a common width, so
every CASCoDe layout is close to a filled rectangle with no device
sticking out. This step removes most of the whitespace before any
fine-grained tuning: it accounts for 62\% of the reduction
(Table~\ref{tab:sizing_share}) and lifts device density to 0.61--0.89
(Table~\ref{tab:layout_comparison}).

\parhead{(2) Adaptation to circuit structure}The same action set adapts
to the structure of each circuit instead of applying one template.
\emph{Symmetric circuits.} In the differential and fully differential
amplifiers, such as \texttt{1s\_5t}, \texttt{1s\_sym}, and
\texttt{1s\_fcfd}, matched devices are moved as groups, so the compact
arrangement keeps the two halves of the circuit mirrored about a common
axis. \emph{Large passive devices.} In the two- and three-stage
amplifiers, the MIM compensation capacitors are several times larger
than any transistor. APPlace places them as separate blocks at the
periphery and spreads the active devices between them, whereas CASCoDe
abuts them to the active array: along the top edge in \texttt{2s\_se},
\texttt{3s\_acbc}, \texttt{3s\_dacfc}, \texttt{3s\_smc}, and
\texttt{3s\_tcfc}, in a corner in \texttt{2s\_miller} and
\texttt{2s\_pcas}, and on both flanks in \texttt{3s\_affc}. The active
array is shaped to the capacitor edge, so the capacitor no longer
leaves empty space around it; on \texttt{2s\_se} the whole amplifier
becomes a single narrow column under its capacitor. \emph{Dominant wide
devices.} When a few wide devices dominate the footprint, as in
\texttt{1s\_iwil}, \texttt{1s\_nwil}, and \texttt{1s\_fcse}, device
splitting and shape changes fold them into the rows rather than leaving
them as separate long blocks. Each case is handled by a different
sequence drawn from the same predefined actions, which is what lets one
unchanged loop serve all 20 circuits.

\begin{figure*}[t]
\centering
\includegraphics[width=\textwidth]{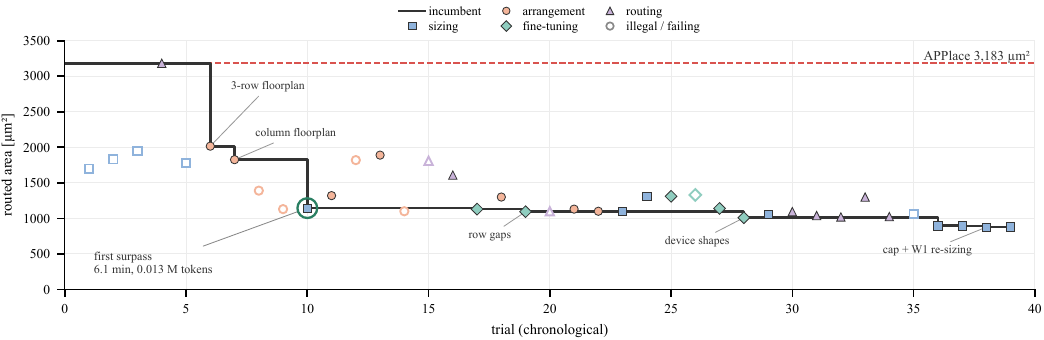}
\caption{One CASCoDe run on \texttt{2s\_se}: trials coloured by action
kind (hollow: illegal or failing), the incumbent staircase from the
APPlace layout with each improving action labelled, and the first
layout that beats every baseline circled.}
\label{fig:trajectory}
\end{figure*}

\parhead{(3) Discrete refinement}Once the arrangement is set, the
remaining gain comes from many small discrete decisions, each kept only
if the layout stays DRC- and LVS-clean and passes the post-layout
criterion. Geometric fine-tuning compresses row gaps and changes device
shapes, and local re-sizing shrinks devices that have slack in the
post-layout metrics; each accounts for 19\% of the reduction
(Table~\ref{tab:sizing_share}). These edits are routing-aware: gap
compression is judged only after routing, because the routing grid can
offset a geometric gain; routing attempts vary wire widths and net
orders; and resistive probes (Eq.~\eqref{eq:resistive_probe}) point to
nets whose resistance limits the FoM. The run of
Fig.~\ref{fig:trajectory} shows the progression: after the floorplan
actions reach 1,144\,\areaunit, row-gap tuning, device-shape changes,
and a final capacitor and transistor re-sizing bring the area to
877\,\areaunit, and every incumbent along the way is legal and passing.

% Required packages: booktabs, multirow
\begin{table}[t]
\centering
\caption{Comparison at identical sizing}
\label{tab:same_sizing}
\footnotesize
\setlength{\tabcolsep}{4pt}
\renewcommand{\arraystretch}{1.15}
\begin{tabular}{@{}llrrrr@{}}
\toprule
Circuits & Metric & APPlace & MAGICAL & ALIGN & CASCoDe\\
\midrule
\multirow{3}{*}{1-stage (10)}
 & Area    & 3,343 & 3,502 & 2,318 & \textbf{1,026}\\
 & Pass    & 70\% & 60\% & 50\% & \textbf{100\%}\\
 & Density & 0.41  & 0.29  & 0.45  & \textbf{0.70}\\
\midrule
\multirow{3}{*}{2-stage (5)}
 & Area    & 5,332 & 7,692 & 2,911 & \textbf{1,631}\\
 & Pass    & 40\% & 0\% & 20\% & \textbf{100\%}\\
 & Density & 0.37  & 0.36  & 0.51  & \textbf{0.79}\\
\midrule
\multirow{3}{*}{3-stage (5)}
 & Area    & 18,933 & 20,082 & 10,328 & \textbf{6,164}\\
 & Pass    & 0\% & 0\% & 0\% & \textbf{100\%}\\
 & Density & 0.45   & 0.37   & 0.56   & \textbf{0.79}\\
\bottomrule
\end{tabular}
\end{table}

% Replaces figures/red_decomp.tex (same data).
\begin{table}[t]
\centering
\caption{Area reduction against APPlace at identical sizing per circuit
class, split by source. Areas are class means; shares are percentages
of the reduction.}
\label{tab:sizing_share}
\footnotesize
\setlength{\tabcolsep}{3.5pt}
\begin{adjustbox}{max width=\columnwidth}
\begin{tabular}{@{}lrrrrrr@{}}
\toprule
& \multicolumn{3}{c}{Area (\areaunit)}
& \multicolumn{3}{c}{Share of reduction (\%)} \\
\cmidrule(lr){2-4}\cmidrule(l){5-7}
Class & APPlace & CASCoDe & Red. & Sizing & Arrange. & Fine-tune \\
\midrule
1-stage (10) & 4,139  & 1,026 & $-$75.2\% & 25.6 & 59.8 & 14.6 \\
2-stage (5)  & 5,932  & 1,631 & $-$72.5\% & 13.9 & 66.3 & 19.8 \\
3-stage (5)  & 21,652 & 6,164 & $-$71.5\% & 17.6 & 61.7 & 20.7 \\
\midrule
All (20)     & --     & --    & --        & \textbf{18.9} & \textbf{62.0} & \textbf{19.1} \\
\bottomrule
\end{tabular}
\end{adjustbox}
\end{table}

% Replaces figures/cost.tex (data: figures/cost_data.csv).
\begin{table}[t]
\centering
\caption{Mean cost per circuit class. Baselines: runtime per run.
CASCoDe: total agent time (trials run in parallel), tokens, and trials
at the first layout that beats every baseline and at the end of the
run.}
\label{tab:cost}
\footnotesize
\setlength{\tabcolsep}{3pt}
\begin{adjustbox}{max width=\columnwidth}
\begin{tabular}{@{}lrrrrrrrrr@{}}
\toprule
& \multicolumn{3}{c}{Baselines (s)} & \multicolumn{6}{c}{CASCoDe} \\
\cmidrule(lr){2-4}\cmidrule(l){5-10}
& & & & \multicolumn{3}{c}{First win} & \multicolumn{3}{c}{Full run} \\
\cmidrule(lr){5-7}\cmidrule(l){8-10}
Class & APPlace & MAGICAL & ALIGN
& min & M tok. & trials
& min & M tok. & trials \\
\midrule
1-stage (10) & 53  & 15.1  & 232 & 3.1  & 0.017 & 4.2  & 41.7  & 0.259 & 46.4 \\
2-stage (5)  & 274 & 43.5  & 319 & 8.8  & 0.052 & 7.6  & 58.0  & 0.362 & 69.4 \\
3-stage (5)  & 601 & 121.2 & 368 & 29.2 & 0.136 & 17.8 & 233.4 & 0.734 & 81.0 \\
\midrule
All (20)     & 240 & 48.7  & 290 & 11.1 & 0.056 & 8.5  & 93.7  & 0.420 & 60.8 \\
\bottomrule
\end{tabular}
\end{adjustbox}
\end{table}

% Regenerate with:
%   python scripts/make_convergence.py figures/trajectory_data_2s_se.csv \
%     figures/fig_trajectory.png --baseline 3183 --first-surpass T010 \
%     --fs-time 6.1 --fs-tokens 0.013 --width 7.0 --height 2.3

% Required packages: booktabs
\begin{table*}[t]
\centering
\caption{Ablation study: FoM, routed area (\areaunit), time (min)/tokens (M),
and DRC count; all layouts pass LVS. Bold: best variant; --: unfinished.}
\label{tab:ablation}
\footnotesize
\setlength{\tabcolsep}{2.6pt}
\renewcommand{\arraystretch}{1.2}
\begin{adjustbox}{max width=\textwidth}
\begin{tabular}{@{}l*{5}{rrrc}@{}}
\toprule
Variant
& \multicolumn{4}{c}{\texttt{ldo}}
& \multicolumn{4}{c}{\texttt{cmp\_ct}}
& \multicolumn{4}{c}{\texttt{1s\_nwil}}
& \multicolumn{4}{c}{\texttt{2s\_pcas}}
& \multicolumn{4}{c}{\texttt{3s\_affc}} \\
\cmidrule(lr){2-5}\cmidrule(lr){6-9}\cmidrule(lr){10-13}
\cmidrule(lr){14-17}\cmidrule(lr){18-21}
& FoM & Area & Time/Tok. & DRC
& FoM & Area & Time/Tok. & DRC
& FoM & Area & Time/Tok. & DRC
& FoM & Area & Time/Tok. & DRC
& FoM & Area & Time/Tok. & DRC \\
\midrule
\textbf{CASCoDe}
& \textbf{4.19} & \textbf{1,340} & \textbf{48}/\textbf{0.33} & 0
& \textbf{3.64} & \textbf{942} & \textbf{65}/\textbf{0.49} & 0
& \textbf{6.93} & \textbf{541} & \textbf{36}/\textbf{0.25} & 0
& 6.70 & \textbf{802} & \textbf{74}/\textbf{0.44} & 0
& \textbf{7.16} & \textbf{5,453} & \textbf{232}/\textbf{0.72} & 0 \\
w/o Action
& 2.06 & 2,499 & 72/0.57 & 0
& 2.97 & 1,683 & 156/0.63 & 17
& 5.92 & 613 & 59/0.35 & 0
& 5.66 & 3,033 & 146/0.75 & 19
& -- & -- & -- & -- \\
w/o Tree Search
& 3.56 & 1,537 & 69/0.51 & 0
& 3.55 & 1,175 & 103/0.71 & 0
& 6.63 & 546 & 41/0.33 & 0
& \textbf{6.73} & 1,319 & 104/0.64 & 4
& 7.08 & 9,254 & 556/2.12 & 8 \\
Agent alone
& 2.35 & 2,889 & 122/0.91 & 5
& 3.04 & 2,181 & 187/1.01 & 23
& 4.84 & 2,032 & 77/0.54 & 17
& -- & -- & -- & --
& -- & -- & -- & -- \\
\bottomrule
\end{tabular}
\end{adjustbox}
\end{table*}

% Regenerate with: python scripts/make_ablation_figures.py figures
\begin{figure*}[t]
\centering
\includegraphics[width=\textwidth]{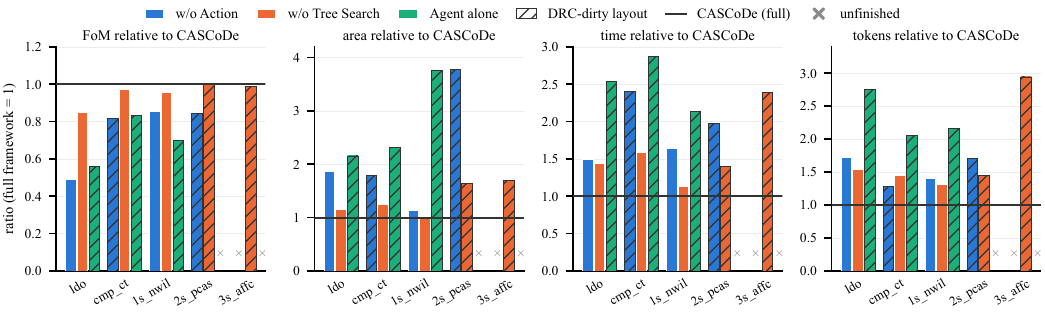}
\caption{Ablation variants relative to the full framework (horizontal
line at 1) on the five circuits of Table~\ref{tab:ablation}. Bars below 1
in FoM and above 1 in area, time, and tokens are losses; hatched bars are
DRC-dirty layouts and $\times$ marks an unfinished run.}
\label{fig:ablation}
\end{figure*}

% Required packages: booktabs
\begin{table*}[t]
\centering
\caption{Agent-model swap with prompt, tools, and budget unchanged: FoM,
routed area (\areaunit), time (min)/tokens (M), and DRC count; all layouts
pass LVS. *: fails the pass criterion.}
\label{tab:model_swap}
\footnotesize
\setlength{\tabcolsep}{2.6pt}
\renewcommand{\arraystretch}{1.2}
\begin{adjustbox}{max width=\textwidth}
\begin{tabular}{@{}l*{5}{rrrc}@{}}
\toprule
Agent model
& \multicolumn{4}{c}{\texttt{ldo}}
& \multicolumn{4}{c}{\texttt{cmp\_ct}}
& \multicolumn{4}{c}{\texttt{1s\_nwil}}
& \multicolumn{4}{c}{\texttt{2s\_pcas}}
& \multicolumn{4}{c}{\texttt{3s\_affc}} \\
\cmidrule(lr){2-5}\cmidrule(lr){6-9}\cmidrule(lr){10-13}
\cmidrule(lr){14-17}\cmidrule(lr){18-21}
& FoM & Area & Time/Tok. & DRC
& FoM & Area & Time/Tok. & DRC
& FoM & Area & Time/Tok. & DRC
& FoM & Area & Time/Tok. & DRC
& FoM & Area & Time/Tok. & DRC \\
\midrule
Opus 5.5 (default)
& 4.19 & 1,340 & 48/0.33 & 0
& 3.64 & 942 & 65/0.49 & 0
& 6.93 & 541 & 36/0.25 & 0
& 6.70 & 802 & 74/0.44 & 0
& 7.16 & 5,453 & 232/0.72 & 0 \\
Sonnet 5.5
& 4.31 & 1,667 & 72/0.44 & 0
& 3.91 & 1,761 & 77/0.62 & 0
& 7.01 & 992 & 51/0.34 & 0
& 6.91 & 1,572 & 134/0.79 & 0
& 6.51* & 9,901 & 301/1.24 & 0 \\
Astra
& 4.12 & 1,281 & 57/0.43 & 0
& 3.44 & 1,001 & 49/0.41 & 0
& 6.63 & 563 & 50/0.31 & 0
& 6.82 & 814 & 84/0.40 & 0
& 6.99 & 5,311 & 198/0.61 & 0 \\
\bottomrule
\end{tabular}
\end{adjustbox}
\end{table*}

\subsection{Area Reduction Analysis}
\label{sec:area}

\parhead{(1) Same-sizing comparison}Table~\ref{tab:same_sizing} reruns
every baseline on the final sizing CASCoDe delivers, so that only the
layout flow differs. The area advantage narrows but remains clear: per
class, CASCoDe goes from 3.5--4.0$\times$ smaller than APPlace in
Table~\ref{tab:layout_comparison} to 3.1--3.3$\times$, from
4.3--5.4$\times$ to 3.3--4.7$\times$ against MAGICAL, and from
2.0--3.0$\times$ to 1.7--2.3$\times$ against ALIGN, while the baselines
meet the criterion on only 45\%, 30\%, and 30\% of the circuits at this
sizing and on no three-stage amplifier.

\parhead{(2) Decomposition of the reduction}Table~\ref{tab:sizing_share}
quantifies the split: with APPlace as a fixed layout engine, re-sizing
accounts for 19\% of the reduction, arrangement (floorplan and
placement) for 62\%, and geometric fine-tuning for 19\%. Device density
agrees: 0.74 for CASCoDe against 0.39, 0.31, and 0.48. The reduction
therefore does not rest on re-sizing; it comes from co-optimizing
sizing with the arrangement decisions.

\subsection{Overhead and Mitigation}
The main cost of CASCoDe is time and tokens (Table~\ref{tab:cost}): a
run averages 61 trials, 94\,min, and 0.42\,M tokens against seconds to
minutes for the baselines. Most of the budget, however, refines an
already-winning incumbent: the first layout that beats every baseline in area and DRC
count while retaining 80\% of the pre-layout FoM appears after 8.5
trials, 11\,min, and 0.056\,M tokens on average, about 12\% of the run. In the run of
Fig.~\ref{fig:trajectory}, two floorplan actions and one re-sizing beat
every baseline by trial 10 (6.1\,min, 0.013\,M tokens); 29 further
refinement trials gain another 23\%.

\subsection{Ablation Study}
\label{sec:ablation}
Table~\ref{tab:ablation} isolates the two components of CASCoDe on
three benchmark circuits plus an LDO and a comparator from prior sizing
studies~\cite{toposizing,pvtsizing}. \emph{w/o Action} removes
the structured action set, leaving the raw tool interfaces; \emph{w/o
Tree Search} replaces the tree with a single chain that always extends
the latest trial; \emph{Agent alone} removes the execution environment,
leaving the tool command lines.

Fig.~\ref{fig:ablation} normalizes every variant to the full framework.
Against these variants the full framework yields the smallest area, the
least time, and the fewest tokens on all five circuits and the highest
FoM on four. Removing the action set inflates area by 1.1--3.8$\times$
and time by 1.5--2.4$\times$, and leaves DRC violations on
\texttt{cmp\_ct} and \texttt{2s\_pcas}; removing tree search inflates
them by up to 1.7$\times$ and 2.4$\times$, with the largest loss on the
three-stage \texttt{3s\_affc}. The agent alone finishes three cases,
none DRC-clean.

Reading Fig.~\ref{fig:ablation} panel by panel separates what each
component contributes. In the FoM panel, the single-chain variant stays
close to the full framework (0.85--1.00$\times$), whereas removing the
action set costs 15--51\% of the FoM (down to 0.49$\times$ on
\texttt{ldo}) and the agent alone 16--44\%: the structured actions are
what let the agent reach high-quality designs at all. The tree search
matters mainly for cost. Without it the agent cannot return to an
earlier parent, so on \texttt{3s\_affc} it spends 2.4$\times$ the time
and 2.9$\times$ the tokens and still ends 1.7$\times$ larger. The agent
alone is the most expensive variant on every circuit it finishes
(2.1--2.9$\times$ the time and 2.1--2.8$\times$ the tokens), and all of
its bars are hatched: without the execution environment it neither
reaches a compact layout nor a DRC-clean one.

\subsection{Sensitivity to the Agent Model}
\label{sec:model}
Table~\ref{tab:model_swap} swaps Opus~5.5, the latest high-end model,
for a mid-tier model of the same family (Sonnet~5.5) and another
vendor's flagship (Astra) on the same five circuits, with prompt, tools,
and budget unchanged. All three deliver DRC- and LVS-clean layouts on
every circuit: legality rests on the action space and the verification
loop, not on the model.

The models differ in how they spend the search budget. Sonnet is conservative, keeping FoM at or above
Opus on four circuits at 1.2--2.0$\times$ the area and 40\% more
trials, and on \texttt{3s\_affc} it neither passes nor saves area.
Astra is greedy, following one route with 12\% fewer trials to within
6\% of Opus's area while scoring a lower FoM than Opus on four of the
five circuits (by up to 5.5\%, on \texttt{cmp\_ct}). Both alternatives still produce smaller layouts than
every baseline of Table~\ref{tab:layout_comparison} on the three
benchmark circuits, although Sonnet's \texttt{3s\_affc} layout fails the
pass criterion.

\subsection{Limitations}
\label{sec:limits}
\parhead{Gap to expert layouts}The structured actions encode part of
the knowledge a layout designer applies, such as moving matched devices
as groups, abutting large capacitors to the active array, and folding
wide devices into rows. This clearly improves on the automated flows,
but the delivered layouts still fall short of carefully hand-crafted
ones. The agent can only combine the predefined actions with the
bridging tools it writes, and the objective covers routed area and
post-layout performance under a fixed set of testbenches. Practices
that DRC, LVS, and these metrics do not capture, such as gradient-aware
matching, shielding of sensitive nets, and reliability considerations,
are therefore not optimized explicitly.

\parhead{Cost--quality trade-off}CASCoDe buys layout quality with time
and tokens: a run averages 94\,min and 0.42\,M tokens, against seconds
to minutes for the baselines (Table~\ref{tab:cost}), and the cost grows
with circuit size, to 233\,min for the three-stage class. The first
layout that beats every baseline appears after about 12\% of this
budget and independent trials run in parallel, so a run can stop early
at much lower cost; the further refinement, such as the final 23\% in
the run of Fig.~\ref{fig:trajectory}, is paid for with additional
trials.

% Last page: do not stretch the columns to full height.
\raggedbottom
\section{Conclusion}
CASCoDe is a closed-loop LLM agent framework that runs feedback-guided tree search over a hierarchical action space from sizing to a verified GDS. One unchanged loop handles 20 SKY130 amplifiers, delivering DRC- and LVS-clean layouts on every circuit at 59--85\% less area than a conventional P\&R flow, most of it from arrangement decisions. Ablations confirm that both the action space and the tree search reduce area and search cost.
\clearpage
\bibliographystyle{IEEEtran}
\bibliography{references}

@inproceedings{magical,
 author={Biying Xu and Keren Zhu and Mingjie Liu and Yibo Lin and Shaolan Li and Xiyuan Tang and Nan Sun and David Z. Pan},
 title={{MAGICAL}: Toward Fully Automated Analog {IC} Layout Leveraging Human and Machine Intelligence},
 booktitle={IEEE/ACM International Conference on Computer-Aided Design (ICCAD)},
 year={2019},
 url={https://yibolin.com/publications/papers/ANALOG_ICCAD2019_Xu.pdf}
}

@article{align,
 author={Tonmoy Dhar and Kishor Kunal and Yaguang Li and Meghna Madhusudan and Jitesh Poojary and Arvind K. Sharma and Wenbin Xu and Steven M. Burns and Ramesh Harjani and Jiang Hu and Desmond A. Kirkpatrick and Parijat Mukherjee and Sachin S. Sapatnekar and Soner Yaldiz},
 title={{ALIGN}: A System for Automating Analog Layout},
 journal={IEEE Design \& Test}, volume={38}, number={2}, pages={8--18}, year={2021},
 doi={10.1109/MDAT.2020.3042177}
}

@misc{rowbased,
 author={Endalk Y. Gebru and {Ramprasath S.} and Ramesh Harjani and Sachin S. Sapatnekar},
 title={Row-Based Layout Synthesis for Analog Circuits Using Height-Quantized Primitives},
 howpublished={arXiv:2606.21767}, year={2026}, url={https://arxiv.org/abs/2606.21767}
}

@inproceedings{diffsp,
 author={Peng Xu and Mingzi Wang and Yapeng Li and Yuyang Ye and Tinghuan Chen and Tsung-Yi Ho and Bei Yu},
 title={{DiffSP}: Differentiable Sequence Pair-based Analog Placement},
 booktitle={ACM/IEEE Design Automation Conference (DAC)},
 year={2026},
 doi={10.1145/3770743.3804034}
}

@misc{postbo,
 author={Xiaohan Gao and Haoyi Zhang and Siyuan Ye and Mingjie Liu and David Z. Pan and Linxiao Shen and Runsheng Wang and Yibo Lin and Ru Huang},
 title={Post-Layout Simulation Driven Analog Circuit Sizing},
 howpublished={arXiv:2310.14049}, year={2023}, url={https://arxiv.org/abs/2310.14049}
}

@misc{panda,
 author={Haoyi Zhang and Weijian Fan and Xiaohan Gao and Bingyang Liu and Runsheng Wang and Yibo Lin},
 title={{PANDA}: An {LLM}-Enhanced Performance-Driven Analog Design Framework Bridging Design Intent and Layout Generation},
 howpublished={arXiv:2606.15052}, year={2026}, url={https://arxiv.org/abs/2606.15052},
 note={Preprint version cited}
}

@misc{analogmaster,
 author={Xian Rong Qin and Yong Zhang and Ying Hu and Tao Su and Bo-Wen Jia and Ning Xu},
 title={{AnalogMaster}: Large Language Model-based Automated Analog {IC} Design Framework from Image to Layout},
 howpublished={arXiv:2604.20916}, year={2026}, url={https://arxiv.org/abs/2604.20916}
}

@misc{icpi,
 author={Bingyang Liu and Ziming Wei and Xiaohan Gao and David Z. Pan},
 title={Simulation-Aware In-Context Policy Improvement for {LLM}-Aided Analog Layout Refinement},
 howpublished={arXiv:2608.13767}, year={2026}, url={https://arxiv.org/abs/2608.13767}
}

@inproceedings{mace,
 author={Wenlong Lyu and Fan Yang and Changhao Yan and Dian Zhou and Xuan Zeng},
 title={Batch {B}ayesian Optimization via Multi-objective Acquisition Ensemble for Automated Analog Circuit Design},
 booktitle={International Conference on Machine Learning (ICML)}, series={PMLR}, volume={80}, pages={3306--3314}, year={2018},
 url={https://proceedings.mlr.press/v80/lyu18a.html}
}

@inproceedings{autockt,
 author={Keertana Settaluri and Ameer Haj-Ali and Qijing Huang and Kourosh Hakhamaneshi and Borivoje Nikolic},
 title={{AutoCkt}: Deep Reinforcement Learning of Analog Circuit Designs},
 booktitle={Design, Automation \& Test in Europe Conference \& Exhibition (DATE)}, pages={490--495}, year={2020},
 doi={10.23919/DATE48585.2020.9116200}
}

@inproceedings{turbo,
 author={David Eriksson and Michael Pearce and Jacob R. Gardner and Ryan Turner and Matthias Poloczek},
 title={Scalable Global Optimization via Local Bayesian Optimization},
 booktitle={Advances in Neural Information Processing Systems}, volume={32}, pages={5497--5508}, year={2019},
 url={https://arxiv.org/abs/1910.01739}
}

@article{toposizing,
 author={Ziming Wei and Zichen Kong and Yuan Wang and David Z. Pan and Xiyuan Tang},
 title={{TopoSizing}: An {LLM}-Aided Framework of Topology-Based Understanding and Sizing for {AMS} Circuits},
 journal={IEEE Transactions on Computer-Aided Design of Integrated Circuits and Systems},
 volume={45}, number={11}, pages={5136--5148}, year={2026},
 doi={10.1109/TCAD.2026.3685525}
}

@inproceedings{pvtsizing,
 author={Zichen Kong and Xiyuan Tang and Wei Shi and Yiheng Du and Yibo Lin and Yuan Wang},
 title={{PVTSizing}: A {TuRBO-RL}-Based Batch-Sampling Optimization Framework for {PVT}-Robust Analog Circuit Synthesis},
 booktitle={Proceedings of the 61st ACM/IEEE Design Automation Conference (DAC)},
 articleno={47}, year={2024},
 doi={10.1145/3649329.3661850}
}

@inproceedings{applace,
 author={Bingyang Liu and Xiaohan Gao and Souradip Poddar and David Z. Pan},
 title={{APPlace}: Co-Optimizing Device Locations and Access Points for Routing-Aware Analog Placement},
 booktitle={Asia and South Pacific Design Automation Conference (ASP-DAC)}, year={2027},
 doi={10.1145/3840404.3848293},
 note={arXiv:2610.39515}
}

@article{layoutcopilot,
 author={Bingyang Liu and Haoyi Zhang and Xiaohan Gao and Zichen Kong and Xiyuan Tang and Yibo Lin and Runsheng Wang and Ru Huang},
 title={{LayoutCopilot}: An {LLM}-Powered Multiagent Collaborative Framework for Interactive Analog Layout Design},
 journal={IEEE Transactions on Computer-Aided Design of Integrated Circuits and Systems},
 volume={44}, number={8}, pages={3126--3139}, year={2025},
 doi={10.1109/TCAD.2025.3529805}
}

@inproceedings{loadbench,
 author={Markus Leibl and Filipe Azevedo and Stefan Uhlich and Andrea Bonetti and Arun Venkitaraman and Chia-Yu Hsieh and Phuoc Pham and Lorenzo Servadei and Helmut Graeb and Ricardo Martins},
 title={{LOADBench}: A Large Open Analog Dataset and Benchmark for Machine Learning in {IC} Design},
 booktitle={ACM/IEEE International Symposium on Machine Learning for CAD (MLCAD)},
 year={2026}, doi={10.1145/3831599.3840321}, url={https://arxiv.org/abs/2610.21591}
}

@inproceedings{charbon94,
 author={Edoardo Charbon and Enrico Malavasi and Davide Pandini and Alberto L. Sangiovanni-Vincentelli},
 title={Simultaneous Placement and Module Optimization of Analog {IC}'s},
 booktitle={ACM/IEEE Design Automation Conference (DAC)},
 pages={31--35}, year={1994}
}

@inproceedings{plantage,
 author={Martin Strasser and Michael Eick and Helmut Graeb and Ulf Schlichtmann and Frank M. Johannes},
 title={Deterministic Analog Circuit Placement Using Hierarchically Bounded Enumeration and Enhanced Shape Functions},
 booktitle={IEEE/ACM International Conference on Computer-Aided Design (ICCAD)},
 pages={306--313}, year={2008}
}

@inproceedings{alignprim,
 author={Meghna Madhusudan and Arvind K. Sharma and Yaguang Li and Jiang Hu and Sachin S. Sapatnekar and Ramesh Harjani},
 title={Analog Layout Generation Using Optimized Primitives},
 booktitle={Design, Automation and Test in Europe Conference (DATE)},
 pages={1234--1239}, year={2021}
}

@article{habal,
 author={Husni Habal and Helmut Graeb},
 title={Constraint-Based Layout-Driven Sizing of Analog Circuits},
 journal={IEEE Transactions on Computer-Aided Design of Integrated Circuits and Systems},
 volume={30}, number={8}, pages={1089--1102}, year={2011}
}

@article{aida,
 author={Nuno Louren{\c{c}}o and Ricardo Martins and Ant{\'o}nio Canelas and Ricardo P{\'o}voa and Nuno Horta},
 title={{AIDA}: Layout-Aware Analog Circuit-Level Sizing with In-Loop Layout Generation},
 journal={Integration, the VLSI Journal},
 volume={55}, pages={316--329}, year={2016}
}

@inproceedings{bagnet,
 author={Kourosh Hakhamaneshi and Nick Werblun and Pieter Abbeel and Vladimir Stojanovi{\'c}},
 title={{BagNet}: Berkeley Analog Generator with Layout Optimizer Boosted with Deep Neural Networks},
 booktitle={IEEE/ACM International Conference on Computer-Aided Design (ICCAD)},
 year={2019}
}

@inproceedings{closingloop,
 author={Mingjie Liu and Keren Zhu and Xiyuan Tang and Biying Xu and Wei Shi and Nan Sun and David Z. Pan},
 title={Closing the Design Loop: {Bayesian} Optimization Assisted Hierarchical Analog Layout Synthesis},
 booktitle={ACM/IEEE Design Automation Conference (DAC)},
 year={2020}
}

@inproceedings{ispd23,
 author={Ahmet Faruk Budak and Keren Zhu and Hao Chen and Souradip Poddar and Linran Zhao and Yaoyao Jia and David Z. Pan},
 title={Joint Optimization of Sizing and Layout for {AMS} Designs: Challenges and Opportunities},
 booktitle={ACM International Symposium on Physical Design (ISPD)},
 pages={84--92}, year={2023}
}

@inproceedings{analogcoder,
 author={Yao Lai and Sungyoung Lee and Guojin Chen and Souradip Poddar and Mengkang Hu and David Z. Pan and Ping Luo},
 title={{AnalogCoder}: Analog Circuit Design via Training-Free Code Generation},
 booktitle={AAAI Conference on Artificial Intelligence (AAAI)},
 year={2025}
}

@inproceedings{adollm,
 author={Yuxuan Yin and Yu Wang and Boxun Xu and Peng Li},
 title={{ADO-LLM}: Analog Design {Bayesian} Optimization with In-Context Learning of Large Language Models},
 booktitle={IEEE/ACM International Conference on Computer-Aided Design (ICCAD)},
 year={2024}
}

@misc{ledro,
 author={Dimple Vijay Kochar and Hanrui Wang and Anantha Chandrakasan and Xin Zhang},
 title={{LEDRO}: {LLM}-Enhanced Design Space Reduction and Optimization for Analog Circuits},
 howpublished={arXiv:2411.12930}, year={2024}, url={https://arxiv.org/abs/2411.12930}
}
\end{document}